\documentclass[3p,times,preprint]{elsarticle}
\usepackage{url}
\usepackage{lineno}

\modulolinenumbers[1] %

\usepackage{ecrc}
\usepackage{array}
\usepackage{adjustbox}
\usepackage{xcolor}

\volume{}

\firstpage{1}

\journalname{Preprint}

\runauth{Gonzalez-Ruiz et al.}

\jid{eai}

\jnltitlelogo{Preprint}

\usepackage{amssymb}

\usepackage[figuresright]{rotating}
\usepackage{enumitem}
\usepackage{siunitx}
\usepackage{booktabs}
\usepackage{multirow}
\usepackage{caption}
\usepackage{graphicx}
\usepackage{longtable}

\begin{document}

\begin{frontmatter}

\dochead{}

\title{Assessing Long-Term Electricity Market Design for Ambitious Decarbonization Targets using Multi-Agent Reinforcement Learning}

\author[label1,label3,label4]{Javier Gonzalez-Ruiz\corref{cor1}}
\author[label2,label3,label4]{Carlos Rodriguez-Pardo}
\author[label4,label5]{Iacopo Savelli}
\author[label1,label3,label4]{Alice Di Bella}
\author[label2,label3,label4]{Massimo Tavoni}

\address[label1]{Department of Electronics, Information and Bioengineering, Politecnico di Milano, Piazza Leonardo da Vinci 32, Milan, 20133, Italy.}

\address[label2]{Department of Management, Economics and Industrial Engineering, Politecnico di Milano, Piazza Leonardo da Vinci 32, Milan, 20133, Italy.}

\address[label3]{CMCC Foundation- Euro-Mediterranean Center on Climate Change, Via Marco Biagi 5, Lecce, 73100, Italy.}

\address[label4]{RFF-CMCC European Institute on Economics and the Environment, Via Bergognone 34, Milan, 20144, Italy.}

\address[label5]{Centre for Research on Geography, Resources, Environment, Energy and Networks (GREEN), Bocconi University, Via Roberto Sarfatti, Milan, 20136, Italy.}

\cortext[cor1]{javier.gonzalez@cmcc.it}

\begin{abstract}
Electricity systems are key to transforming today’s society into a carbon-free economy. Long-term electricity market mechanisms, including auctions, support schemes, and other policy instruments, are critical in shaping the electricity generation mix. In light of the need for more advanced tools to support policymakers and other stakeholders in designing, testing, and evaluating long-term markets, this work presents a multi-agent reinforcement learning model capable of capturing the key features of decarbonizing energy systems. Profit-maximizing generation companies make investment decisions in the wholesale electricity market, responding to system needs, competitive dynamics, and policy signals. The model employs independent proximal policy optimization, which was selected for suitability to the decentralized and competitive environment. Nevertheless, given the inherent challenges of independent learning in multi-agent settings, an extensive hyperparameter search ensures that decentralized training yields market outcomes consistent with competitive behavior. The model is applied to a stylized version of the Italian electricity system and tested under varying levels of competition, market designs, and policy scenarios. Results highlight the critical role of market design for decarbonizing the electricity sector and avoiding price volatility. The proposed framework allows assessing long-term electricity markets in which multiple policy and market mechanisms interact simultaneously, with market participants responding and adapting to decarbonization pathways.
\end{abstract}


\end{frontmatter}

\section{Introduction}
\label{Section - Introduction}

One of the key actors of a climate transition is electricity generation, given its relevance and its central role as an enabler of decarbonization across other sectors \cite{intergovernmental_panel_on_climate_change_working_2022}. In scenarios aligned with the Paris Agreement reviewed by the Intergovernmental Panel on Climate Change (IPCC), electricity generation is projected to reach net-negative carbon emissions before 2050, thus enabling the electrification of all end-use sectors. 

In this context, most scenarios informing policy-making are predominantly generated by central planning models, which have limited capacity to represent the wide set of policies needed to ensure the green transition. Most notably, given that wholesale electricity systems are predominantly structured as markets featuring active participation from private merchant actors alongside regulated institutions (including transmission companies), electricity market design is crucial for aligning the evolution of energy systems with climate objectives \cite{newbery_market_2018}. In parallel to their relevance for the energy transition, electricity systems and markets have come under increased scrutiny due to their vulnerabilities and questions regarding resilience in the face of large-scale crises, such as the disruptions triggered by the war in Ukraine \cite{batlle_power_2022}. 

Amid these concerns, a key policy challenge lies in the design of long-term electricity markets. These markets, including auctions, support schemes, and contracting mechanisms, aim to incentivize adequate investment in generation and flexibility resources. Their goals range from de-risking capital investment and ensuring resource adequacy, to shielding consumers from price volatility and ensuring affordability \cite{cramton_capacity_2013, newbery_missing_2016, oren_generation_2005}. Long-term markets complement short-term markets (e.g., day-ahead, intraday, and balancing markets), which focus on efficiently dispatching and utilizing existing assets \cite{us_department_of_energy_benefits_2006}. Regarding long-term markets, two dominant paradigms have emerged: Energy-only Markets (EoMs) and Capacity Remuneration Mechanisms (CRMs) \cite{cramton_capacity_2013}. However, the evolving policy landscape has prompted a surge in reform proposals. These range from enhanced long-term contracting for renewable energy sources in the form of Contract for Differences (CfD) \cite{zachmann_design_2023, newbery_efficient_2023, schlecht_financial_2024, european_parliament_improving_2024}, to the introduction of mandatory contracting schemes \cite{wolak_long-term_2022}, and the development of new financial instruments aimed at shielding consumers from price volatility \cite{batlle_power_2022}. A more far-reaching approach to transforming electricity markets is gaining traction alongside these incremental changes: hybrid electricity market models \cite{roques_adapting_2017, joskow_hierarchies_2022, corneli_prism-based_2020, keppler_why_2022}, frameworks in which governments assume a stronger role in guiding and selecting large-scale investments in a competition for the market. In contrast, decentralized markets continue to govern short-term operations via competition in the market. Although intriguing and attractive from a theoretical standpoint, such arrangements lack quantitative evaluations in the Literature and are still far from real-life implementations.

Considering this broad spectrum of potential market architectures, each with far-reaching implications beyond the electricity sector, modeling tools are essential for informed decision-making \cite{bublitz_survey_2019}, where Agent-based and partial equilibrium models have traditionally dominated this space. However, these approaches may face limitations in capturing the dynamics of transformative policy scenarios or exploring radically new institutional designs. More specifically, the explicit modeling of auctions and their system-wide interactions and complexities, a defining characteristic of modern electricity markets that will become increasingly significant as supporting and long-term mechanisms gain prominence, is often overlooked or only partially addressed in the existing literature. 

Multi-Agent Reinforcement Learning (MARL) offers a promising modeling alternative that addresses these shortcomings, providing a highly flexible modelling framework that can complement existing approaches by simulating adaptive behaviours in complex, multi-agent systems. Nevertheless, the application of MARL to electricity market modeling is still in its early stages and requires further development to become a robust tool for policy and technical assessment. This work concentrates on enabling MARL models as a comprehensive tool for long-term electricity market assessments, with contributions expanded as follows:

\begin{itemize}[noitemsep, left=8pt]
\item We develop an open-source multi-agent environment that models the long-term electricity market, extending current implementations that concentrate on short-term markets. In its current version, the model allows for investing in generation assets through stylized capacity and contract for difference markets, aside from merchant investments. Moreover, integrating other incentive instruments and accommodating different market designs can be easily extended. 
\item This work includes a detailed process for a hyperparameter search in the context of proximal policy optimization applied to the electricity markets. Efforts in this area showcase the potential and difficulties of applying MARL to competitive environments. 
\item By integrating a long-term market environment with a MARL training pipeline, the first application to the best of our knowledge, this work provides a versatile framework for evaluating ambitious decarbonization strategies in electricity systems. The approach offers near-unlimited flexibility in electricity market design. Four key advantages distinguish this modeling framework from existing methods in the literature:
\begin{itemize}
\item{First, the model explicitly incorporates auction mechanisms, which are central to many long-term market designs and are commonly used to facilitate and de-risk investment in energy infrastructure}. 
\item{Second, it supports comparing multiple market instances and policy layers, such as carbon taxes, within a unified framework, enabling a systematic evaluation of the effectiveness and potential redundancy of various policy instruments}. 
\item{Third, agents in the model manage portfolios that include both new investments and existing assets, allowing the analysis of possible divergent incentives and behaviors of incumbent versus entrant actors}. 
\item{Fourth, the model captures the impact of market competition on outcomes, a critical feature in wholesale electricity systems that often exhibit comparatively high market concentration}. 
\end{itemize}
\item The code for this framework is open-source, available via a public repository\footnote{\url{https://github.com/jjgonzalez2491/MARLEY_V1}}. 
\end{itemize}

The rest of this article is organized as follows. Section \ref{Section - Related work} presents the related work on modeling practices for electricity markets. Section \ref{Section - Long-term Electricity Market Environment} introduces the complete MARL implementation. Next, Section \ref{Section - Training and Validation} describes the training setup and the process for hyperparameter selection. After, Section \ref{Section - Market results} presents market results obtained with the MARL model. Finally, Section \ref{Section - Conclusions} concludes. 

\section{Related Work}
\label{Section - Related work}

This work addresses a research gap by applying Multi-Agent Reinforcement Learning to long-term electricity markets, where strategic decisions regarding investment in utility-scale generation assets are crucial. To establish the foundation for this contribution, this section presents related literature on two main fronts: first, highlighting current practices in long-term electricity market analysis, and second, examining the flexibility, opportunities, and limitations identified in applying MARL to a similar field. As such, the first two subsections introduce Agent-Based and Partial-Equilibrium models for long-term electricity markets, the two primary approaches that have addressed pressing concerns related to decarbonization efforts and have examined investment strategies in detail. From this baseline, applications of Multi-Agent Reinforcement Learning to short-term electricity markets are discussed, drawing connections to long-term market modeling where applicable.

\subsection{Agent-Based models}

In Agent-Based models, electricity markets are constructed and represented via the detailed depiction of decisions and interactions between market players, actors, and policy-makers. In this category, EmLab-Generation, an Agent-Based model designed for long-term electricity market analysis \cite{chappin_simulating_2017}, has been used to analyze the impact of capacity markets in a system with a growing share of renewable energy sources. Specifically, studies show the effectiveness of the scheme in comparison to strategic reserves for ensuring reliability via the promotion of low-cost peak generation units \cite{bhagwat_effectiveness_2017}. Similarly, authors in \cite{bhagwat_analysis_2017} demonstrate that capacity markets that allocate long-term commitments to investments, instead of annual contracts, are preferable when ensuring security of supply. In \cite{khan_how_2018}, EmLab is enhanced with short-term modules, showcasing the potential of demand response and energy storage systems to outright replace or compete against capacity markets. Finally, given the possible phase-out of support schemes for renewable technologies in Germany and the Netherlands, scenarios demonstrate significant reductions in new capacity additions of low-carbon technologies and price increases \cite{marc_melliger_phasing_2022}. 

Expanding on the previous framework, Brain-Energy has been developed to increase agent heterogeneity and include institutional agents, such as governments and regulators, enabling endogenous decision-making regarding key policies in the energy sector \cite{barazza_impact_2020, barazza_co-evolution_2020}. Analysis of the UK, Germany, and Italy transition pathways using Brain-Energy shows that historic-path dependence in investment choices can displace low-carbon investments in scenarios with weak regulatory frameworks \cite{barazza_key_2021}. Furthermore, in \cite{barazza_co-evolution_2020}, authors argue that, in scenarios with heterogeneous agents and higher capital requirements, more aggressive policy action to promote decarbonization efforts is necessary to achieve environmental targets. 

Authors in \cite{anwar_modeling_2022} have developed the EMIS-AS model by implementing a richer representation of electricity systems and market sessions. EMIS-AS learning capabilities, implemented via Kalman Filters, are particularly relevant and aim to forecast key parameters for the investment profitability assessments. Using the previous framework, several market designs are tested when pursuing clean energy targets ranging between 45\% and 100\% by 2030 \cite{frew_interaction_2023}. Among other insights, authors find that energy-only and capacity markets can achieve clean energy targets while maintaining operational constraints. Moreover, carbon pricing is the most effective mechanism for reaching the first wave of renewable penetration, while stacking several mechanisms to promote the energy transition in scenarios with more aggressive targets demonstrates only marginal benefits. Expanding these ideas, the EMIS-AS model enabled a detailed comparison between capacity markets and operating reserve demand curves \cite{anwar_can_2024}. 

This modeling category has also included risk metrics in the agent decision-making formulation. In \cite{anwar_modeling_2022}, risk-averse agents compare projects and technologies through a utility function, implicitly assigning higher discount rates to comparatively larger investments. More broadly, authors in \cite{yang_investment_2023} implement and compare three risk metrics in an agent-based model: Value-at-Risk, Mean-Variance, and Adjusted discount rate. Finally, it is worth mentioning the family of short-term Agent-Based models, such as AMIRIS, that complement the long-term perspective with decisions regarding the dispatch of power plants, in addition to interactions between a broader set of agents and actors in short-term markets \cite{deissenroth_assessing_2017,schimeczek_amiris_2023}. 

\subsection{Partial Equilibrium and Bi-level optimization models}

Partial Equilibrium and/or Bi-level optimization for electricity markets refer to model formulations where hierarchical optimization problems represent a market and are generally resolved in a Nash equilibrium. In \cite{gabriel_complementarity_2014}, a comprehensive overview of the different types of formulations and their application to electricity markets is presented.

In this modeling category, particular attention has been given to long-term market design, and most relevantly, the impact of Capacity Markets in the presence of risk-averse agents. Initially, \cite{hoschle_electricity_2017} introduced an equilibrium model capable of representing both Capacity Markets and Strategic Reserves. Results showcased comparative benefits in the former policy option, considering the additional incentives it provides for actors to participate in the rest of the market. Extending the model, \cite{hoschle_admm-based_2018} presented a market formulation that includes risk-adverse agents and a solution algorithm based on the Alternating Direction Method of Multipliers. Similar to the previous case, when compared to an Energy-only framework, the capacity market improved the system costs while maintaining the necessary reserve margins. On this note, authors in \cite{mays_asymmetric_2019}, using a two-stage stochastic program for capacity expansion, argue that capacity mechanisms have an asymmetric effect on the risk profiles of generation technologies, which promotes the integration of low-capital, high-variable-cost technologies. Regarding the impact of demand response on capacity market design, analysis carried out in \cite{kaminski_impact_2021} using a stochastic non-cooperative capacity planning model with risk-averse investors demonstrated that these schemes partially mitigate the effect of risk aversion on social welfare. 

Aside from Capacity Market design, authors in \cite{billimoria_insurance_2022} design a mechanism capable of aligning consumer preferences with potential investors for strategic reserves by implementing two bi-level optimization problems, the first representing an Energy-only market, and a second in which retailers present their willingness to pay for extra levels of reliability to be provided by an insurer of last resort. Regarding the cost of capital on systems with high shares of wind and solar technologies, the two-stage stochastic equilibrium model developed in \cite{mays_financial_2023} complements standard market models by including hedge providers, thus being able to quantify the impact of market and price volatility in the cost of capital for particular technologies.

Lastly, equilibrium models have also been utilized to understand the implications of incomplete risk-trading schemes in electricity markets. By considering the system conditions leading to the contingencies faced by ERCOT during February 2021, in \cite{mays_private_2022}, the authors promote, among other measures, a shift towards mandatory contracting obligation on retailers. To directly understand the impact of missing markets in the presence of risk-averse preferences, in \cite{dimanchev_consequences_2024} three models for the power system are compared: a risk-neutral expansion problem, a risk-averse and missing risk market scenario, in which both supply and demand formulation are risk-averse agents but their optimization problems are solved independently, and a risk-averse expansion problem. By following a similar framework, the work in \cite{dimanchev_choosing_2024} aims to support the design of decarbonization policies when considering risk-averse agents in incomplete electricity markets. 

\subsection{Multi-Agent Reinforcement Learning and Electricity Markets}

Reinforcement Learning (RL) is a particular subset of Machine Learning techniques in which agents are trained to perform specific tasks by performing actions in their environment, having received the corresponding feedback from these interactions \cite{sutton_reinforcement_2020, winder_reinforcement_2021}. Recent advancements have enabled agents to be parametrized using Neural Networks, extending the reach of RL to areas where a flexible, adaptable, and scalable nonlinear optimizer is advantageous \cite{winder_reinforcement_2021, zuccotto_reinforcement_2024}.  

A natural extension of RL that models the interaction of several agents within the same environment, trained to perform tasks with shared or competing objectives, is denoted Multi-Agent Reinforcement Learning (MARL) \cite{albrecht_multi-agent_2024}. MARL has been extensively applied to various subjects, including the energy sector and policy-making \cite{sven_gronauer_multi-agent_2021}. Particularly relevant to this work are applications where, beginning with naive and unstructured environments characterized by agent competition, neural networks have outperformed both humans and traditional algorithms in solving highly complex tasks \cite{openai_dota_2019,vinyals_grandmaster_2019, yu_surprising_2022}. In the electricity sector, RL and MARL have been applied across the electricity value chain \cite{yang_reinforcement_2020, kell_machine_2022, zhu_reinforcement_2023}. Among these applications, the most prominent focus has been on analyzing bidding and participation strategies in short-term electricity markets, moving beyond the conventional paradigm in which agents submit bids approximating their marginal production costs. In this context, MARL represents a technique to improve current Agent-based models by filling the gap with partial-equilibrium setups. Specifically, MARL takes advantage of the flexibility that has characterized agent-based models, but extends it to setups where partial-equilibrium models have been used to understand efficient and competitive market interactions. 

Starting from \cite{ye_multi-period_2019}, authors apply a Deep Policy Gradient algorithm to model price-bidding strategies of single-plant Generation Companies (GENCO) in a short-term market. Tests showcased that the proposed RL algorithm reached a Nash Equilibrium in not-congested cases compared to a simplified partial-equilibrium model. Extending the previous work, in \cite{ye_deep_2020}, a Deep Deterministic Policy Gradient (DDPG) algorithm is applied to a market and network configuration resembling the previous implementation. In this case, the system improves computational and market performance compared with other RL algorithms, namely Q-learning and Deep-Q network. Compared with the Mathematical Programs with Equilibrium Constraints solution, it obtained higher profits for the single-plant GENCOs. Similarly, the work in \cite{liang_agent-based_2020} uses a DDPG to analyze a system in which GENCO presents quadratic cost functions, parametrized by their bidding strategy. Most relevantly, results highlight the importance of hyperparameter selection, as increasing the discount factor during training exhibited increments in the profit earned by agents from the market, thus deviating from the equilibrium obtained in the analytic solution. Shifting from the independent learning presented thus far, where each agent is trained concurrently without a centralized architecture, in \cite{du_approximating_2021}, the Multi-Agent DDPG (MADDPG) is implemented to reach the Nash Equilibrium in a day-ahead market, concentrating again on price-bidding strategies. Authors find that the centralized training and decentralized execution framework in the algorithm improved computational efficiency compared to the independent learning methods. Continuing with DDPG, in \cite{graf_computational_2023}, the authors conduct an extensive hyperparameter search, enabling the independent learning scheme to reach a Bertrand Equilibrium in a stylized system. Furthermore, a Mean-Field algorithm based on DDPG is applied to analyze trading strategies in peer-to-peer (P2P) energy markets \cite{qiu_mean-field_2023}, a setup where prosumers have limited access to information for their decision making. Results demonstrate agents can learn efficient strategies to bid prices and quantities in the double-auctions from the P2P market, while the Mean-Field algorithms outperform other MARL options. Deviating from models dedicated to price-bidding strategies, a Soft-Actor Critic algorithm is proposed to bid prices and quantities jointly in a short-term market \cite{xu_joint_2022}. Although the framework is not applied to a multi-agent context, the authors find the computational and economic benefits of the coupled evaluation in bidding for short-term markets. 

Shifting from DDPG, authors in \cite{harder_fit_2023} develop the model ASSUME, an agent-based approach focused on short-term markets and structured upon the multi-agent Twin-Delayed DDPG. By following a centralized training and decentralized execution paradigm, authors can extend their framework to hundreds of agents, thus showcasing prices that resemble real market prices from the German electricity market. The model includes an additional incentive that penalizes inefficient actions undertaken by agents, complementing the pure economic profits as the training reward utilized in most of the work presented thus far. In \cite{harder_how_2024}, the previous model's performance is compared with its bi-level optimization problem counterpart, showcasing limits in the RL approach while achieving the theoretical equilibrium, but also demonstrating the potential in the modeling framework in terms of flexibility and scalability to more complex setups. Similarly, by applying explainable artificial intelligence techniques, authors analyze the bidding strategies obtained by the ASSUME model \cite{miskiw_explainable_2024}. In particular, agents are willing to increase their strategic bids in cases where they are potential price setters. At the same time, this incentive is reduced when it is more likely that their bids will be lower than the marginal market price. 

In parallel to the work focused on short-term market analysis, other works have placed efforts extending MARL towards related topics. Worth noting are the SustainGym environments \cite{yeh_sustaingym_nodate}, which have been designed to enable a standardized comparison of RL and MARL algorithms in systems from the energy and environmental sectors. Complementary, in a stylized model, authors in \cite{renshaw-whitman_non-stationarity_2024} propose the inclusion of an active regulator/central planner as part of the learning agents in a MARL market simulation, enabling policy design that endogenously responds to strategic behavior from market actors, and vice-versa.

\section{Multi-Agent Long-term Electricity Market Reinforcement Learning Environment}
\label{Section - Multi-Agent Long-term Electricity Market Reinforcement Learning Environment}

This section describes the implementation of the Long-term Electricity Market in the Multi-Agent Reinforcement Learning framework. Section \ref{Section - Long-term Electricity Market Environment} introduces the long-term electricity market environment. Next, Section \ref{Section - Multi-Agent Reinforcement Learning applied to the Market Environment} describes the MARL scheme applied to the long-term electricity market environment. As explained throughout this section, maintaining a rich and complex representation of a long-term electricity market that enables policy assessments is a key goal of this work. As a result, design elements and features in the environment have become intrinsically and inevitably connected with the algorithm selection. Thus, the generalization of this environment to other MARL algorithms remains a compelling area for future research.

\subsection{Long-term Electricity Market Environment}
\label{Section - Long-term Electricity Market Environment}

\begin{figure}[h!]
    \centering
    \includegraphics[width=0.8\linewidth]{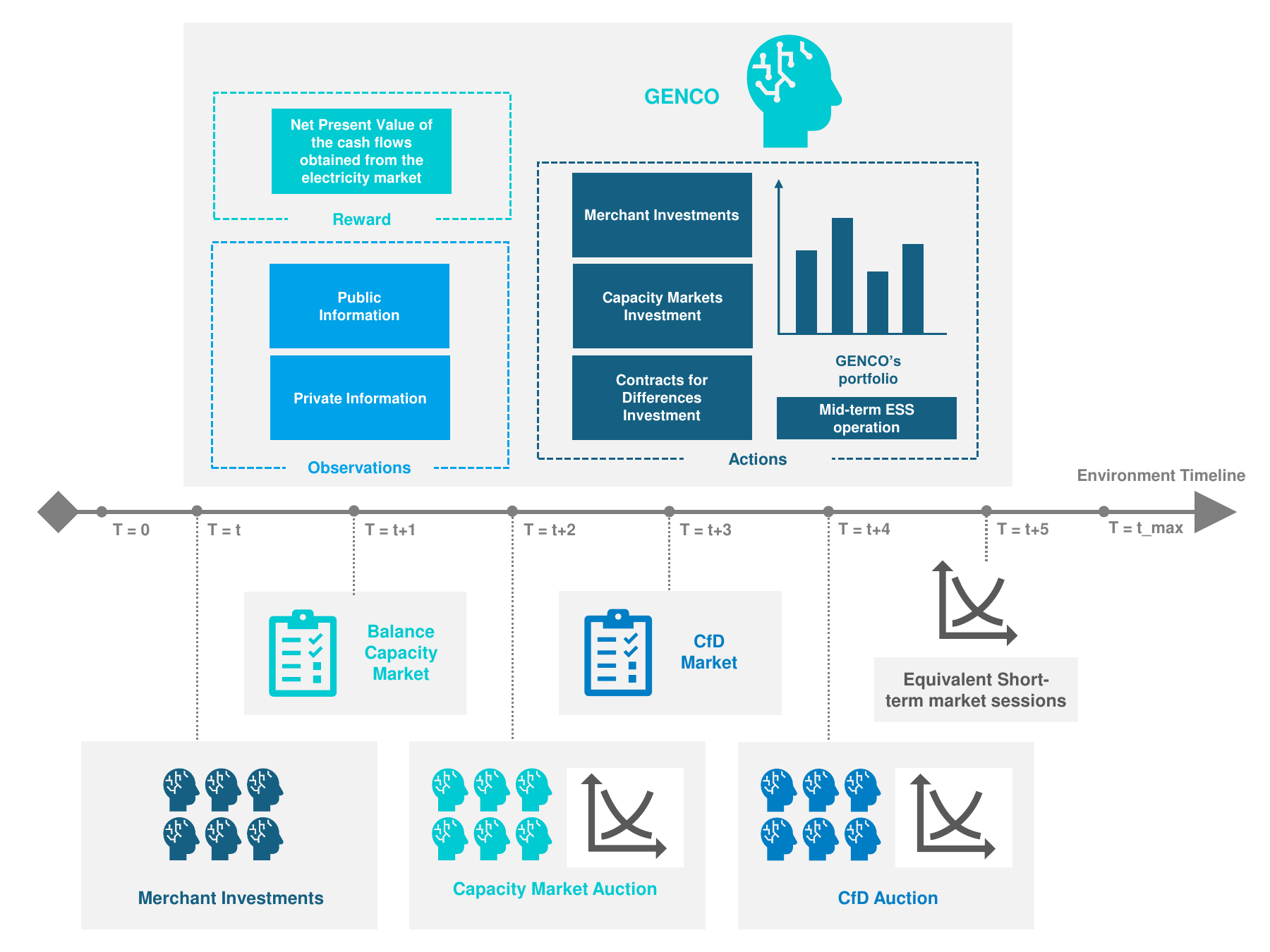} %
    \caption{Structure of the long-term electricity market in the reinforcement learning model. The diagram highlights the interactions between GENCO agents and the market environment. From the market, agents receive observations \textit{(public information, published by entities such as Market Operators and TSOs, and private information, regarding the performance of their portfolio)}, to take actions \textit{(investment decisions)} in the system, with the aim to maximize their profits during the simulation. The figure illustrates one year of market operation, where agents participate in the short-term market at each environment step while making investment decisions annually. Investment decisions occur sequentially through mutually exclusive entry mechanisms \textit{(merchant, Capacity Market, CfD market)}, with the latter two depending on system balance calculations, as detailed in Section \ref{Section - Long-term Electricity Market Environment}.}
    \label{fig: General Overview}
\end{figure}

The Long-term Electricity Market Environment, schematized in Figure \ref{fig: General Overview}, is described via its two main structural components, starting from the electricity market design and moving towards translating such market to the MARL framework. Following this structure, \ref{Appendix - Long-term Electricity Market Environment} goes into further detail on the model description. 

\subsubsection{Long-term Electricity Market}

The model presented in this work targets mid- and long-term (several years or, at most, a couple of decades) decarbonization analysis in wholesale electricity markets, focusing on utility-scale investment decisions and market mechanisms designed to promote them. To this end, the model represents GENCOs, as shown in the upper part of Figure \ref{fig: General Overview}. GENCOs are assumed to be profit-maximizing entities while subject to system conditions, policy signals, and competition from other market players. GENCOs invest in generation and Energy Storage Systems (ESS) assets to participate in the electricity market and expand their portfolio. The characteristics of generation and storage technologies are detailed in \ref{Appendix - Generation technologies} and \ref{Appendix - Energy Storage Systems}. The investment framework assumes unrestricted access to financing with no equity constraints, capital expenditures (CAPEX) incurred during the construction phase, and pre-tax cash flows. 

To represent day-ahead markets, the model uses equivalent periods through representative days. Representative days are defined as sequential 24-hour windows, with an hourly resolution, intended to capture daily patterns in electricity systems. To model renewable resource availability, time series projections from \cite{antonini_weather-_2024,antonini_weather-_2024-1} describe their hourly capacity factor. In the case of electricity demand, projections for the Italian power system are used to describe the yearly profile \cite{di_bella_mitigation_2025}. Once all necessary time series are obtained, the TSAM Python library \cite{hoffmann_typical_2021} transforms them into representative periods, which are then applied to the equivalent short-term market sessions. For each hour in the equivalent short-term sessions, GENCOs participate in the market by presenting bids for their portfolio (energy quantity and offered price). In line with most long-term partial-equilibrium models, the environment assumes that GENCOs present bids for the short-term market that correspond to their marginal production costs. The bids from all GENCOs and resources in the system serve as input for double-sided marginal price auctions, dispatching resources based on their merit order and setting a unique system price. 

Regarding the operation of the electricity system, the model adopts a copper plate assumption, similar to the day-ahead market clearing in, for example, France and Germany. Although limited, this approach directly evaluates long-term electricity market design as a tool or potential obstacle towards ambitious decarbonization objectives. Future work will concentrate on improving network modeling within the MARL context, with possible pathways for enhancement detailed in \ref{Appendix - Transmission Infraestructure}.

As previously stated, GENCOs are also able to invest in ESS assets. However, ESS investments are restricted to technologies with relatively short-duration storage capacities (3–4 hours), such as Lithium-ion batteries. In contrast, longer-duration storage technologies remain under incumbent agents' control for operational decisions. In each equivalent short-term market session, GENCOs decide the desired state of charge level of the long-duration ESS for the next period, thus representing inter-period flexibility in the technology. Within the day-ahead markets, both short and long-duration ESSs are operated for efficient system operation, as described in \ref{Appendix - Energy Storage Systems}. 

Given this selection of generation and storage assets, reduced in comparison to existing planning models but representative of the main trends in the sector, investment can occur via the three available and mutually exclusive channels and markets:

\begin{itemize}[noitemsep, left=8pt]
\item \textbf{Merchant Investments:} Any GENCO can freely decide to place projects of different technologies under construction, with no centralized planning process by the regulator. Once a merchant investment enters operation, the project earns revenues directly from the day-ahead market. 
From a GENCO's perspective, its plant has no protections against sustained low prices during excess supply situations, which might prove insufficient to justify the investment. However, when scarcity conditions occur, the plant will harness the full-fledged scarcity rents. 
\item \textbf{Contract for Difference Market:} In this market, a regulator \textit{(or central planner)} establishes a penetration target for RES in the system, defined as the share between renewable production and total demand. When GENCOs have built insufficient RES projects to achieve the desired penetration target, an auction takes place to cover the missing renewable share. 
Winning projects in the auction are awarded a stylized version of a two-way Contract for Difference \cite{european_parliament_improving_2024}. From a GENCO perspective, the two-way CfD ensures a fixed price for the total output of the participant plant. Equivalently, the GENCO hedges its project against low-price conditions in the system. In exchange, the GENCO must build the project presented in the auction and withhold the financial obligation associated with the CfD during the contract's lifetime (usually 15 to 25 years). 
\item \textbf{Capacity Market:} In this market, the regulator ensures resource adequacy by guaranteeing sufficient resources to meet peak demand conditions. The Capacity Market is inspired by the Reliability Option framework \cite{cramton_colombia_2007, cramton_capacity_2013, mastropietro_reliability_2024}. From a GENCO's perspective, participation in the Capacity Market provides a premium linked to the project’s contribution to system adequacy, as measured by its capacity credits and the auction allocations. In return, the GENCO must develop the generation project and protect consumers from high-price events through the Reliability Option mechanism. 
\end{itemize}

With this design, the environment concentrates on well-established long-term market arrangements, serving as a baseline that facilitates validation, model comparison, and policy assessments. Exploring other market arrangements, such as those from hybrid design schemes, constitutes an interesting direction for future studies.
 
Apart from market design, the model includes two additional policy instruments that could help to shape and define scenarios. The first is a Carbon Tax, directly linked to the \(CO_2\) emissions produced by generation technologies, which passes through to consumers via the bids in the short-term market. The second instrument is an exogenous limit for investing in specific technologies. This limit can be enforced at any point during the simulation and applied discriminately to market players and investment channels. 

\subsubsection{Electricity Market as a Multi-Agent Reinforcement Learning Environment}

This section integrates the Long-term Electricity Market into the Gymnasium standard \cite{towers_gymnasium_2023}, and more specifically, the version of Multi-Agent environments developed by the RLLIB team \cite{liang_rllib_2018}. Section \ref{Appendix - MARL Environment} presents further information for the RL Environment. 

\paragraph{Environment Structure}

In the Gymnasium standard, environments for Reinforcement Learning use steps to simulate the transitions in the underlying Markov Decision Process. In each step, agents observe the system's state, take action, and receive the corresponding reward. In the context of the market model, these environment steps are directly linked to Equivalent Short-Term Market sessions and the corresponding Representative Periods. Each step is designed to represent a fixed period of operation by condensing interactions into a single 24-hour representation. 

During each environment step, the GENCOs' portfolios participate in the Equivalent Short-Term Market, generating market outcomes that serve as the basis for the environment's observations and rewards. However, aside from operating mid-term ESS, agents do not actively make decisions affecting the short-term market: generation assets bid their availability and marginal costs to the day-ahead market, without strategic bidding enabled, and the intra-day charge/discharge cycles of ESS are defined for efficient system operation. Instead, investment decisions are enabled every year, allowing agents to take actions to expand their portfolios. Investments in the different markets occur every two environment steps and in sequence across the year, as shown in the lower panel of Figure \ref{fig: General Overview}. This implementation enables yearly investments in all the markets if system conditions demand it.  

\paragraph{Reward}

In the model, agents are assumed to maximize the net present value of the cash flows obtained from the electricity market. This involves calculating and aggregating revenues and costs from all assets and markets relevant to the agents' portfolio across the simulation. Yet, to be consistent with real company operations, the reward function is designed to be passed to agents step-by-step, as in equation (\ref{eq:total_reward}). The reward is divided into two parts: profits and investment costs. On the one hand, the profits \(\left(P_{\text{M}},P_{\text{CM}},P_{\text{CfD}}\right) \) derived from all markets in the system; on the other hand, the investment costs \(\left(IC_{\text{M}},IC_{\text{CM}}, IC_{\text{CfD}}\right)\) of new portfolio additions, similarly disaggregated across markets. Apart from the previous terms, and to improve clarity in the financial modeling across the formulation, the discounting of cash flows is carried out inside the environment while considering the discount rate as an exogenous parameter. The detailed disaggregation of profits and costs for each market is presented in \ref{Appendix - Agent Reward}.

\begin{equation}
r_{\text{t}} = \left( p_{m} + P_{\text{CM}} + P_{\text{CfD}} - \left( IC_{\text{M}} + IC_{\text{CM}} + IC_{\text{CfD}} \right) \right) \left(\frac{1}{(1 + r)^t}\right) 
\label{eq:total_reward}
\end{equation}

\paragraph{Action Space}

In the current implementation, the actions available to GENCO for interacting with their environment are, in principle, a combination of Discrete \textit{(e.g., investment quantities)} and Continuous \textit{(e.g. bidding prices for long-term auctions)} actions. Yet, considering the advantages provided by Action Masking to represent both internal and external constraints for the agents and the system \cite{huang_closer_2022}, the model adopts a fully multi-discrete action space, increasing the number of discretization steps for continuous variables to enhance the realism of agent strategies, following the recommendations of \cite{delalleau_discrete_2019}.

In this context, Table \ref{Appendix - Table: List of Actions} provides a comprehensive overview of the actions available to GENCOs. In particular, Action Masking is used to control which agents have access to which decisions \textit{(e.g., to limit investments in specific technologies)} and when a given action can take place \textit{(e.g., considering that investments are not enabled in every environment step)}. Although flexible for constraint representation, using Multi-discrete actions creates several limitations in the implementation. First, investment limits are regulated solely by technology-specific maximum rates (Tables \ref{Table: Generation Technology characteristics} and \ref{Table: Energy Storage Technology characteristics}). Second, agents have prior knowledge of auction price caps. Last, the number of available actions (Table \ref{Appendix - Table: List of Actions}) scales with the number of technologies, justifying the choice of a limited yet representative set of options aligned with energy transition trends.

\paragraph{Observation Space}

The design of the observation set available to GENCOs in the model follows three key principles. First, the selected variables should align with real-world market conditions, where agents can access publicly shared system information while specific details remain private and inaccessible to competitors. Second, no internal price forecasting tools are included in the observation set, ensuring that the model retains full autonomy in decision-making. Finally, the selection process adheres to the principle that any information available in the market is also provided to the agents.

Considering this, Table \ref{Appendix - Table: List of Observations} lists the observations made available to agents across environment steps, subdivided according to their thematic category. The set of observations includes indicative values for the time series modeling demand and variable resource availability, the composition of the energy mix via the assets in operation, individual and system-wide reservoir levels, market information \textit{(such as prices and balances from the Capacity and CfD markets)}, the aggregated reward per technology in these markets, policy-relevant information, and the time associated to the environment step. The framework designed for the GENCO's observation does not depend on the number of agents in the market, as it harnesses aggregated system information, given that the number of agents in each simulation is constant. 

\subsection{Multi-Agent Reinforcement Learning and Competitive Environments}
\label{Section - Multi-Agent Reinforcement Learning applied to the Market Environment}

To start, the preliminaries for Single-Agent and Multi-Agent RL are provided. Next, details of Single-Agent and Multi-Agent PPO algorithms are presented. From these premises, this section concludes with the arguments used for the algorithm selection. 

\subsubsection{Single-Agent Reinforcement Learning}
\label{Subsection - Single-Agent and Multi-Agent Reinforcement Learning}

In RL, the agent's interactions with the environment are modeled using a Markov Decision Process (MDP). MDPs are defined by the tuple \( \mathcal{M} = (\mathcal{S}, \mathcal{A}, \mathcal{P}, r, \gamma) \), where $\mathcal{S}$ corresponds to set defining the state space, $\mathcal{A}$ represents the action space, $\mathcal{P}: \mathcal{S} \times \mathcal{A} \to \mathcal{S}$ is the transition probability between states as a function of the agent's actions, and $r: \mathcal{S} \times \mathcal{A} \to \mathbb{R}$ is the reward function, which depends on the system state and the agent's actions. For simplicity, the notation assumes a fully observable setting, thus allowing the agent to directly translate states to observations $\mathcal{O} \to \mathcal{S}$ \cite{sutton_reinforcement_2020, winder_reinforcement_2021}.  
\sloppy

Trajectories in the MDP can be formalized by the succession of states, actions, and rewards, \( \tau = (s_0, a_0, r_0, \dots s_t, a_t, r_t, \dots, s_T, a_T, r_T) \), where the subscript \(t\) represents time steps,  \([0, T]\) denote the initial and terminal steps in the system, and states, actions, and rewards are sampled from the corresponding sets. The system is a finite-horizon discounted Markov Decision Process for cases where \(T\) is finite, and rewards are discounted by a $\gamma \in (0,1]$. Generally, RL algorithms aim to maximize the expectation of rewards, shown in expression (\ref{eq: objective function}) by learning a policy \(\pi_{\theta}\), characterized by trainable parameters \(\theta\). In Deep RL, policies are represented with Neural Networks, where \(\theta\) refers to the set of describing parameters, varying according to the selected architecture.

\begin{equation}
\label{eq: objective function}
J(\theta) =  \mathbb{E}\left[ \sum_{t=0}^{T} \gamma^t r_t \right]. 
\end{equation}

For optimizing the objective function, auxiliary expressions can be defined. Starting from the concepts of State-Value, in expression (\ref{eq: State-Value function}), and Action-Value functions, in equation (\ref{eq: Action-Value function}), defined as the expected future reward of being in a specific state and following policy \( \pi_{\theta} \), in the former, and the expected future reward of selecting a particular action in a specific state, in the latter. The Advantage function \(A_{t}\), quantifying the reward improvement of taking action \(\ a_{t}\) in state \(\ s_{t}\), in comparison to the expected reward in state \(\ s_{t}\), can be obtained as the difference between Action and State Value functions. In Actor-Critic RL algorithms, the Action-Value function is associated with the policy \(\pi_{\theta}\), while the State-Value function serves as a baseline for performance evaluation.

\begin{equation}
\label{eq: State-Value function}
V_{\pi_{\theta}}(s_t) = \mathbb{E}_{\pi_{\theta}} \left[ \sum_{t=k}^{T} \gamma^{k-t} r_k \mid s_t \right]
\end{equation}

\begin{equation}
\label{eq: Action-Value function}
Q_{\pi_{\theta}}(s_t, a_t) = \mathbb{E}_{\pi_{\theta}} \left[ \sum_{t=k}^{T} \gamma^{k-t} r_k \mid s_t, a_t \right]
\end{equation}

Two key assumptions commonly used in RL algorithms to ensure convergence are particularly relevant compared to the multi-agent framework. The first is the Markov Property, describing MDPs in which the future state depends only on the current state and actions. Similarly, the second is the stationarity of transition dynamics, a condition achieved when \(\ \mathcal{P}\) remains constant during training and execution, even if the transitions are described stochastically  \cite{sutton_reinforcement_2020, winder_reinforcement_2021, bick_towards_2021}. Finally, it is worth mentioning that the framework outlined above operates under risk-neutral assumptions, where the objective is to maximize expected rewards. Theoretical and algorithmic advances have extended reinforcement learning to accommodate risk-averse decision-making \cite{bisi_risk-averse_2022,bonetti_risk-averse_2023}. Nevertheless, the formal evaluation of explicit risk-averse algorithms in the current framework is left for future work.

\subsubsection{Multi-Agent Reinforcement Learning}

The framework for Multi-Agent Reinforcement Learning (MARL) can be introduced as a Partially Observable Stochastic Game (POSG) \cite{albrecht_multi-agent_2024}. In particular, the POSG considers a set of $N$ agents, $I = \{1,\dots, i, \dots, N\}$, each with its set of actions \(\mathcal{A}_i\), where \(\mathcal{A} =\mathcal{A}_1,\dots, \mathcal{A}_N\), interacting within an environment with a finite number of states \(\mathcal{S}\). In the system, transition probabilities are defined as $\mathcal{P}: \mathcal{S} \times \mathcal{A}\times\mathcal{S} \to \mathbb{R}$, a function of states and Actions, an agents harness a reward based on a function \(\mathcal{R}_i: \mathcal{S}\times\mathcal{A}\times\mathcal{S}\to \mathcal{R}\). Finally, agents have access to system states via an Observation function \(\mathcal{O}_i: \mathcal{A} \times \mathcal{S} \times \mathcal{O}_i\).

\sloppy
Analogous to MDP, agent's trajectories can be formalized by the succession of observations, actions, and rewards, \( \tau_i = (o_{i,0}, a_{i,0}, r_{i,0}, \dots o_{i,t}, a_{i,t}, r_{i,t}, \dots, o_{i,T}, a_{i,T}, r_{i,T}) \), where the subscript \(t\) represents time steps, \([0, T]\) denote the initial and terminal steps in the system, $\gamma$ is the discount factor, and observations, actions, and rewards are sampled from the corresponding sets and functions of each agent.

Unlike single-agent MDPs, a POSG requires the definition of a solution concept to guide RL algorithms to find the set of joint policies \(\mathcal{\pi} =\pi_1,\dots, \pi_N\) that aim to achieve the desired objective \cite{albrecht_multi-agent_2024}. The long-term electricity market environment is considered competitive, as agents intend to maximize their profit, with no explicit mechanism for cooperation or communication, apart from standard market interactions, and a solution concept that should approach a Nash Equilibrium. 

Aside from the environment category, MARL is subject to additional challenges compared with the Single-Agent framework \cite{albrecht_multi-agent_2024}. The POSG is non-stationary and non-Markovian, as the transition probability function depends on all agents' actions sampled from policies subject to change during training. Moreover, MARL suffers from credit assignment issues, given that in the reward estimation functions, it is difficult to distinguish between a reward variation caused by the agent's actions and a change produced by other factors at play. Additionally, MARL algorithms may converge to suboptimal equilibria in environments governed by equilibrium solution concepts. Finally, MARL implementations and applications may suffer from scalability issues, given the accelerated growth in the number of states, actions, and observations in POSG as a function of the number of agents in the system. Nevertheless, MARL remains a rapidly evolving field, with ongoing advancements in both theoretical foundations and practical applications \cite{sven_gronauer_multi-agent_2021, kell_machine_2022, zhu_reinforcement_2023, albrecht_multi-agent_2024, zuccotto_reinforcement_2024}. 

\subsubsection{Proximal Policy Optimization and Independent Multi-Agent Learning}
\label{Subsection - Proximal Policy Optimization and Independent Multi-Agent Learning}

This section introduces PPO, following the work in \cite{schulman_proximal_2017,bick_towards_2021}. Policy Gradient Algorithm aims to obtain a stochastic policy \( \pi_{\theta}(a_t | s_t) \) that maximizes the expected reward of the RL agent, as described in expression (\ref{eq: objective function}). To produce such a policy, algorithms of the REINFORCE type, a category within PGM, apply stochastic gradient ascent (SGA) to an expression that seeks to find actions that maximize the expected reward in the given state, as measured by the comparison between the Action-state function \textit{(the Actor)} representing the agent's policy, and the Value-state function \textit{(the Critic)} \cite{sutton_reinforcement_2020, bick_towards_2021, winder_reinforcement_2021}.  

PPO improves upon standard PGM and REINFORCE algorithms by redesigning the objective function to avoid significant updates in the Actor network \cite{schulman_proximal_2017}. The modifications mitigate the risk of converging to a local maximum during training and enable the re-use of the trajectories collected with a given Actor-Critic combination in multiple iterations \textit{(epochs)} of SGA. In particular, PPO proposes a three-part objective function, shown in Equation (\ref{eq: objective PPO}) and further disaggregated in expressions (\ref{eq: clip PPO})-(\ref{eq: Vtarget PPO}), where: 

\begin{equation}
\label{eq: objective PPO}
L(\theta, w) = L^{\text{CLIP}}(\theta) + h L^{\text{entropy}}(\theta) - v L^{\text{VF}}(w)
\end{equation}

\begin{equation}
\label{eq: clip PPO}
L^{\text{CLIP}}(\theta) = \hat{\mathbb{E}_t}\left[ \min \left( p_{t}(\theta) \hat{A_t}, \text{clip} \left(p_{t}(\theta), 1 - \epsilon, 1 + \epsilon\right) \hat{A_t} \right) \right]
\end{equation}

\begin{equation}
\label{eq: Advantage PPO}
\hat{A_t} =V_{t}^{target} - V_{w_{old}}(s_t)
\end{equation}

\begin{equation}
\label{eq: Vf PPO}
L^{\text{VF}}(w) = \hat{\mathbb{E}_t} \left[ \left( V_{w}(s_t) - V_{t}^{target} \right)^2 \right]
\end{equation}

\begin{equation}
\label{eq: Vtarget PPO}
V_{t}^{target} = r_t + \gamma r_{t+1} + \gamma^2 r_{t+2} + \dots + \gamma^{n-1} r_{t+n-1} +\gamma^{n} V_{w_{old}}(s_{t+n})
\end{equation}

\begin{itemize}[noitemsep, left=8pt]
\item The terms \( L^{\text{CLIP}}(\theta)\) and \(L^{\text{entropy}}(\theta) \) guide the updates in the Actor network \( \pi_{\theta}(a_t | s_t) \), while \(L^{\text{VF}}(\theta)\) drive updates in the Critic network \( V_{w}(s_t) \); 
\item \(\theta\) and \(w\) are the parameters of the Actor and Critic networks. Moreover, parameters from the previous algorithm iteration are referred to as \(\theta_{old}\) and \(w_{old}\);
\item The main objective function of PPO, \( L^{\text{CLIP}}(\theta)\), uses an Advantage estimation to shift the Actor Policy toward actions that maximize the expected reward while controlling for the maximum size in the updates using the combination of the \( \text{clip} \) and \(\min \) functions;
\item \( \epsilon \) is a hyperparameter that, in conjunction with the clipping function, limits the ratio between new and old policies to remain in the range \([1 - \epsilon, 1 + \epsilon]\);
\item \( L^{\text{entropy}}(\theta) \) is an entropy term that procures the exploration of new strategies by inducing randomness in action selection in the Actor network. For cohesiveness, the entropy term is not expanded, but \cite{bick_towards_2021} includes expressions for both continuous and discrete action spaces;
\item \( p_{t}(\theta) \) is defined as the variation in actor policies between the algorithm updates, measured by the fraction \( \frac{\pi_{\theta}(a_t | s_t)}{\pi_{\theta_{\text{old}}}(a_t | s_t)} \); 
\item \(A_{t}\) corresponds to the Advantage function, estimated via the difference between an estimate of the Action-state function, \(V_{t}^{target}\), denoted as target state value, and the previous version of the Value-state function \(V_{w_{old}}(s_t)\);
\item \(L^{\text{VF}}(\theta)\) drives the updates in the Critic network via minimizing the squared error between the predicted State-value function \(V_{w}(s_t) \), and the estimated target state value \(V_{t}^{target}\);
\item \(V_{t}^{target}\) is used as a proxy of the Action-state value function, and is calculated using the accumulated rewards in a given trajectory with length \((t+n)\); and
\item \( h \) and \( v \) are coefficients that balance exploration, via the entropy term, and the weight of the value function loss with respect to the total loss. 
\end{itemize}

Given the PPO objective function, the training cycle starts with initializing the Actor and Critic networks. Using the current version of the Actor Policy, trajectories are collected from the environment until the batch, controlled by the hyperparameter with the same name, is filled. Given these trajectories, auxiliary state-value functions are estimated. Moreover, multiple epochs of SGA are performed using the same batch of collected trajectories, either through randomly selected subsets of trajectories \textit{(mini-batches)} or full-batches of trajectories. Results from the SGA are then used to update the Actor and Critic networks. Last, the process is repeated until convergence, or until the training budget is reached. 

To implement IPPO, the single-agent version of the algorithm is extended to all agents in the environment. Starting from the definition of the Actor, \( \pi_{i,\theta_{i}}(a_{i,t} | o_{i,t}) \), and Critic networks, \( V_{i,w_i}(o_{i,t}) \), for agent \(i \). In these definitions, parameters \( \theta_{i}\) and \(w_i\) are, in principle, independent and different between agents. Both functions are no longer associated with the system state but depend upon the observations \(o_{i,t}\) harvested by agents from the environment. 

Given this starting point, the PPO objective function, showcased in expression (\ref{eq: objective MAPPO}) and detailed in (\ref{eq: clip MAPPO})-(\ref{eq: Vf target MAPPO}), is extended across agents in the environment. In general, the description of the main variables from the single-agent case holds while extending the notation to account for the multiplicity of agents. Similarly, IPPO follows the same principles as the single-agent case. Critic and Actor networks are initialized for each agent. Trajectories are collected using the most up-to-date policies until the batch is complete, and auxiliary reward functions are estimated. The collected batch undergoes multiple SGD updates, independent for each agent, based on the PPO objective function. Networks are updated, and the process repeats iteratively until convergence or until the training budget is reached. 

\begin{equation}
\label{eq: objective MAPPO}
L_{i}(\theta_{i}, w_{i}) = L_{i}^{\text{CLIP}}(\theta_{i}) + h_{i} L_{i}^{\text{entropy}}(\theta_{i}) - v_{i} L_{i}^{\text{VF}}(w_{i})
\end{equation}

\begin{equation}
\label{eq: clip MAPPO}
L_{i}^{\text{CLIP}}(\theta_{i}) = \hat{\mathbb{E}_{i,t}}\left[ \min \left( p_{i,t}(\theta_{i}) \hat{A_{i,t}}, \text{clip} \left(p_{i,t}(\theta_{i}), 1 - \epsilon_{i}, 1 + \epsilon_{i}\right) \hat{A_{i,t}} \right) \right]
\end{equation}

\begin{equation}
\label{eq: Advantage MAPPO}
\hat{A_{i,t}} =V_{i,t}^{target} - V_{i,w_{i,old}}(o_{i,t})
\end{equation}

\begin{equation}
\label{eq: Vf MAPPO}
 L_{i}^{\text{VF}}(w_{i}) = \hat{\mathbb{E}_t} \left[ \left( V_{i,w_i}(o_{i,t}) - V_{i,t}^{target} \right)^2 \right]
\end{equation}

\begin{equation}
\label{eq: Vf target MAPPO}
V_{i,t}^{target} = r_{i,t} + \gamma_i r_{i,t+1} + \gamma_i^2 r_{i,t+2} + \dots + \gamma_i^{n-1} r_{i,t+n-1} +\gamma_i^{n} V_{i,w_{i,old}}(o_{i,t+n})
\end{equation}

Considering this baseline, it is worth discussing the implications of Independent Multi-Agent PPO in light of the MARL challenges highlighted in previous Sections. Beginning with the Advantage and Value function target definitions, where expressions (\ref{eq: Advantage MAPPO}) and (\ref{eq: Vf target MAPPO}) show that the accumulated rewards in the trajectories exclusively guide the evaluation of individual agent action performance. As a result, the algorithm will inherently suffer from Credit Assignment issues, as it is impossible to discern the actual effect of individual actions on the agent's performance beyond the information provided by the observations. Similarly, the estimates produced by value and policy networks are affected by exogenous variability in the environment, from the single-agent perspective, induced by other agents' actions. As demonstrated in \cite{lowe_multi-agent_2017}, this issue can prevent meaningful learning altogether, even in relatively simple environments. Furthermore, IPPO does not explicitly constrain the learning process to reach an equilibrium. Because of that, the algorithm neither ensures that an equilibrium can be reached, nor guarantees that the obtained solution corresponds to a Nash Equilibrium \cite{albrecht_multi-agent_2024}. Moreover, if an equilibrium is achieved, it would result from agents independently adjusting their strategies until significant variations in the Actor and Critic networks are no longer encouraged. In this sense, the algorithm cannot discriminate and/or select between multiple equilibria. 

\subsubsection{Algorithm Selection for the Multi-Agent Long-term Electricity Market Environment}
\label{Subsection - Algorithm Selection for the Multi-Agent Long-term Electricity Market Environment}

Despite the possible concerns highlighted in the last section, independent learning and PPO (IPPO) have been selected as the model's core \textit{solver} for this work. This section provides the foundations for this selection, first addressing the key issues of algorithm selection, and later delving into training paradigms for multi-agent configurations.

To begin, it is helpful to summarize the market model's key characteristics and desired features. A primary contribution of the MARL approach is its ability to represent auctions as an explicit entry mechanism for market participants. In this context, enabling agents to develop stochastic policies could be advantageous, as it allows for a richer set of learned behaviors and strategies for market interaction. From an implementation standpoint, the market environment is fast and computationally efficient, making it well-suited for parallelized trajectory collection. Moreover, no explicit model is known beforehand to explain market interactions.

This set of characteristics leads to the selection of the family of model-free policy optimization algorithms, in line with the conclusions in \cite{harder_fit_2023}. Yet, in this case, the use of multi-discrete actions and the possibility of parallelization given the fast-running environment lean towards PPO \cite{schulman_proximal_2017} and IMPALA \cite{espeholt_impala_2018} algorithms, with the latter being better suited for large-scale RL applications. In contrast, algorithms derived from DDPG, such as MADDPG, are less appropriate for the current context, given their exclusive support for continuous actions \cite{winder_reinforcement_2021}. This reasoning is consistent with recommendations from Ray and RLlib \cite{liang_rllib_2018}, the framework currently used for the implementation.

Regarding training paradigms, four main trends have emerged for multi-agent environments \cite{albrecht_multi-agent_2024}. First, independent learning treats each agent as a separate learner using the same algorithm, resulting in fully decentralized training and action execution. Second, centralized training with decentralized execution uses shared value functions (e.g., system-wide action/state value functions) to address non-stationarity and credit assignment \cite{lowe_multi-agent_2017}. Third, mean-field methods enable agents to interact with a limited subset of peers, modeling local interactions while preserving decentralized training and execution \cite{yang_mean_2018}. Finally, population-based approaches train a single policy across multiple agents, leveraging diverse training conditions to enhance robustness \cite{vinyals_grandmaster_2019}.

For the model introduced in Section \ref{Section - Long-term Electricity Market Environment}, independent learning emerges as the most suitable learning architecture. To start, GENCOs in the wholesale electricity market are competitors, where explicit information and strategy sharing are discouraged and/or strictly prohibited. Moreover, independent learning aligns well with a key characteristic of electricity markets: repeated market interactions. This repetition allows agents to refine their strategies continuously and, in some cases, even develop tacit collaboration with competitors. Additionally, certain actions, such as bids in specific auction types, remain private and are never disclosed to other players. Furthermore, independent learning facilitates experimentation with heterogeneous and misaligned agent objectives.

On the contrary, alternative paradigms could face substantial limitations. Training a fully shared value function across dozens of agents, each with multidimensional discrete action spaces, poses severe scalability challenges that may hinder effective learning. Furthermore, shared-value function methods commonly assume a shared-reward function \cite{lowe_multi-agent_2017}, a characteristic of cooperative environments that cannot be replicated for long-term electricity markets. Additionally, while effective in peer-to-peer energy trading scenarios \cite{qiu_mean-field_2023}, mean-field approaches are less relevant in wholesale markets where explicit coalition formation is excluded by design. Similarly, population-based methods, which augment the independent learning paradigm across multiple instances, remain out of reach of most academic research, given their extreme computational requirements \cite{vinyals_grandmaster_2019}.

These arguments strongly support independent learning as the most appropriate choice under limited computational resources. While population or league-based training may offer superior robustness when resources are abundant, independent learning provides a realistic, scalable, and conceptually faithful representation of competition in wholesale electricity markets.

Complementing the discussion related to algorithm selection and the training paradigm, multi-agent PPO has shown strong empirical performance in various competitive and cooperative scenarios \cite{de_witt_is_2020, yu_surprising_2022}. Specifically, empirical studies have argued that the PPO objective function, designed to shield policy gradient methods from large and destructive policy updates, has also proven effective in mitigating the non-stationary conditions of multi-agent settings \cite{yu_surprising_2022}. More specifically, these empirical results have shown that the PPO hyperparameters can be adjusted to trade off sample efficiency for training stability beyond what is customarily done in single-agent environments, which, in turn, appears to be the main reason behind successful multi-agent applications.

Overall, the previous reasons motivate the selection of IPPO for this work. Nonetheless, acknowledging the challenges associated with IPPO in competitive environments, further analyses are carried out before delving into market outcomes. Section \ref{Section - Training and Validation} first evaluates the algorithm's performance, discusses hyperparameter selection, and compares the equilibria emerging from different parameter choices. Furthermore, extensive testing is conducted in Section \ref{Section - Market results}, highlighting expected and unexpected behaviors in the solutions from an electricity market perspective, serving as complementary validation of the modeling approach.

\section{Implementation details, Training, and Hyperparameter Selection}
\label{Section - Training and Validation}

This Section details the implementation, describes the training, and presents the hyperparameter Selection applicable to the current context.

\subsection{Implementation details}

The implementation is based on the RLLIB libraries, and particularly, the Python environment uses Python 3.9, Ray 2.4.3, PyTorch 2.1.2, and CUDA 11.8. For training, single computing nodes in a supercomputing system were used. Each node comprises two Intel Xeon Platinum 8360Y (36 cores), two NVIDIA A100 GPUs, and 800 GB of RAM. During training, 69 parallel environments were used for sampling, only one GPU was employed for SGA, and 50 GB of RAM were allocated. Job scheduling was carried out using the LSF system, with reproducibility scripts shared as part of the repository.

\subsection{Training}
\label{SubSection - Training}

Extensive tests are carried out to analyze the training behavior and serve as an input for the hyperparameter Selection discussed in the next Section. For these tests, two environment configurations are used: an EoM, denoted as environment A, and a system with all investment channels enabled, denoted as environment B. In both, 16 agents compete in the market; 8 incumbents with all technologies available for investments, seven entrants, each with one technology available, and a last agent dedicated to operating its mid-term hydro-reservoirs. The Maximum investment per technology is set to 4GW, a relatively high value considering the system conditions, thus increasing competitive pressure in the market. Apart from these differences, both environments abide by the description and input parameters presented in \ref{Appendix - Long-term Electricity Market Environment}.  

With these environments, 58 tests are carried out starting from the recommendations of \cite{yu_surprising_2022}, showing performance improvements using conservative hyperparameter setups for multi-agent training. From this starting point, the runs aim to evaluate the contribution of the most relevant parameters in the algorithm: Clipping parameter \( \epsilon\), Entropy Coefficient \(h\), Batch Size, and Actor and Critic Network Architectures. Regarding the latter, testing is conducted using Multi-Layer Perceptrons (MLPs) and Long Short-Term Memory (LSTM) configurations. Furthermore, a training budget of 25 hours of wall time is set for all tests. Last, no hyperparameter scheduling is applied to ensure the training process represents algorithm behavior. Following these guidelines, Table \ref{Table: Notation Summary for Tests Conducted During Hyperparameter Search.} summarizes the training tests, while Tables \ref{Appendix - Table: List of hyperparameters used during search.} and \ref{Appendix - Table: Full Notation for Tests Conducted During Hyperparameter Search.} provide complementary information. 

\begin{table}[hbt!]
\small
\centering
\resizebox{\textwidth}{!}{%
\begin{tabular}{|c|c|c|c|c|c|}
\hline
\textbf{Code} &
  \textbf{Network type} &
  \textbf{Test description} &
  \textbf{Code} &
  \textbf{Network type} &
  \textbf{Test description} \\ \hline
\begin{tabular}[c]{@{}c@{}} \textbf{\textit{M.1, M.2,}} \\ \textbf{\textit{M.3, M.4}}\end{tabular} &
  MLP &
  Increasing Clipping Factor - \( \epsilon\)&
  \textbf{\textit{L.3, L.4}} &
  LSTM &
  \begin{tabular}[c]{@{}c@{}}MLP in tail\\ Medium LSTM    \\ Increasing Batch Size\end{tabular} \\ \hline
\begin{tabular}[c]{@{}c@{}}\textbf{\textit{M.5, M.6, M.2,}} \\ \textbf{\textit{M.7, M.8, M.9}}\end{tabular} &
  MLP &
  Increasing Batch Size &
  \textbf{\textit{L.5, L.6}} &
  LSTM &
  \begin{tabular}[c]{@{}c@{}}MLP in tail\\ Long LSTM \\ Increasing Batch Size\end{tabular} \\ \hline
\textbf{\textit{M.10, M.2, M.11}} &
  MLP &
  Increasing Entropy Coefficient - \(h\)&
  \textbf{\textit{L.7, L.8}} &
  LSTM &
  \begin{tabular}[c]{@{}c@{}}MLP in head\\ Short LSTM  \\ Increasing Batch Size\end{tabular} \\ \hline
\begin{tabular}[c]{@{}c@{}}\textbf{\textit{M.12, M.2}}, \\ \textbf{\textit{M.13, M.14}}\end{tabular} &
  MLP &
  \begin{tabular}[c]{@{}c@{}}Varying Network Architecture by increasing\\the size in hidden layers\end{tabular} &
  \textbf{\textit{L.9, L.10}} &
  LSTM &
  \begin{tabular}[c]{@{}c@{}}MLP in head\\ Medium LSTM \\ Increasing Batch Size\end{tabular} \\ \hline
\textbf{\textit{L.1, L.2}} &
  LSTM &
  \begin{tabular}[c]{@{}c@{}}MLP in tail\\ Short LSTM  \\ Increasing Batch Size\end{tabular} &
  \textbf{\textit{L.11, L.12}} &
  LSTM &
  \begin{tabular}[c]{@{}c@{}}MLP in head\\ Long LSTM \\ Increasing Batch Size\end{tabular} \\ \hline
\end{tabular}%
}
\caption{Summary of tests conducted during the hyperparameter search. According to the order presented in the testing category, the parameter of interest is increased \textit{(e.g. for Batch Size tests, M.6 configuration has a higher Batch Size than M.5) }. For individual hyperparameter tests, the corresponding code combines the environment and the set of hyperparameters, where A indicates the Energy-only-Market and B the Capacity Market environments \textit{(e.g. AM.11 indicates the test in the EoM environment)}.}
\label{Table: Notation Summary for Tests Conducted During Hyperparameter Search.}
\end{table}

\begin{figure}[hbt!]
    \centering
    \includegraphics[width=0.9\linewidth]{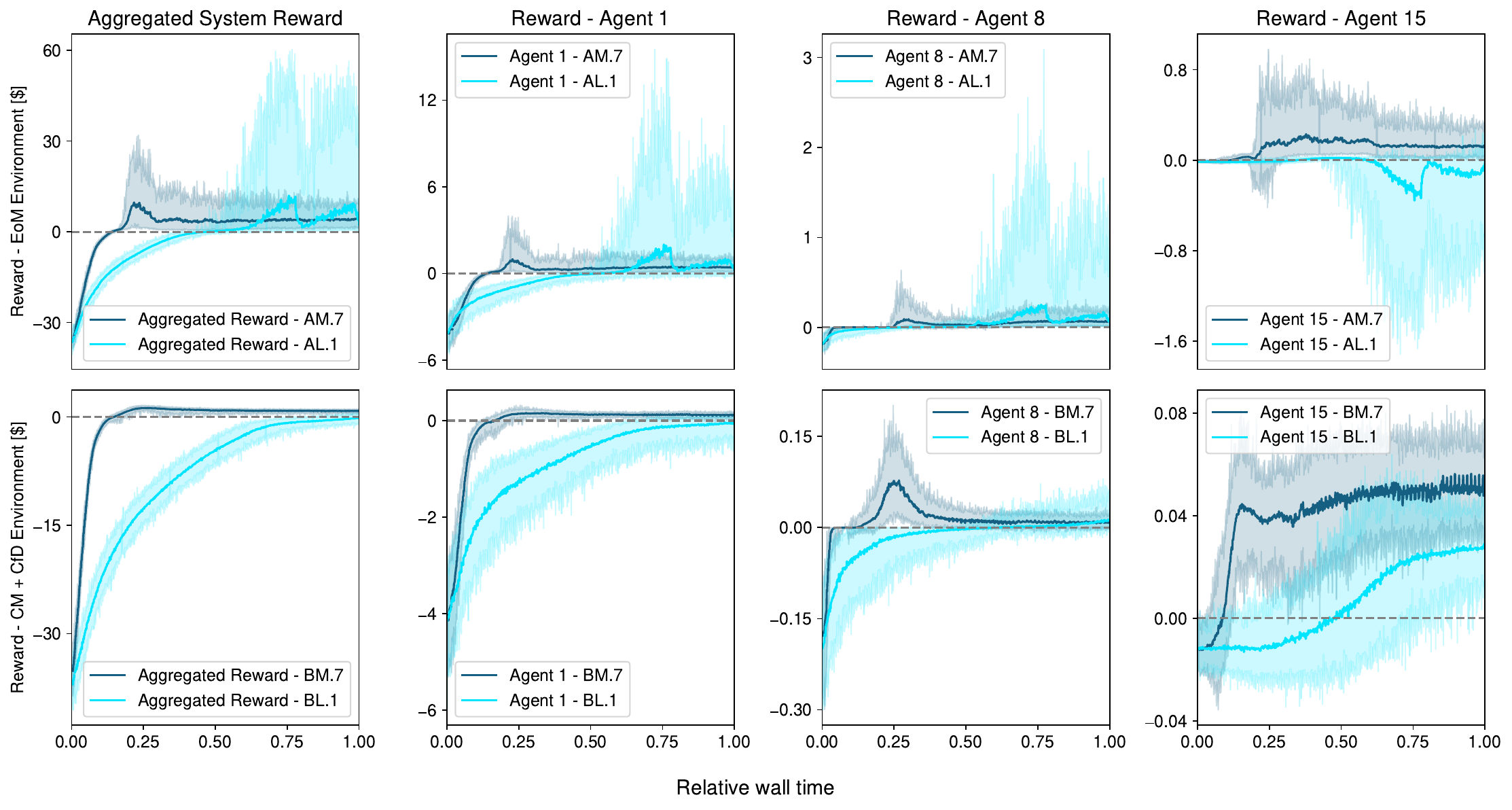} %
    \caption{Evolution of aggregated and individual reward during training for Hyperparameter configurations M7 (\textit{MLP network)} and L1 (\textit{LSTM network)}. The upper graphs present results for the EoM environment, while the lower ones concentrate on the Capacity plus CfD market. The left panels show the aggregate reward for the system. Panels in the second to the fourth columns present the reward evolution during training for Agent 1 \textit{(Incumbent)}, Agent 8 \textit{(Entrant - Solar PV)}, and Agent 15 \textit{(mid-term storage operation)}, respectively. Results are normalized according to the relative wall time used for training. Solid curves indicate average values obtained during sampling, while shaded areas represent minimum and maximum values.}
\label{fig: Summarized Training Behavior}
\end{figure}

Showcasing the training behavior, Figure \ref{fig: Summarized Training Behavior} presents the evolution of aggregated reward as a function of sampled environment steps in four selected tests. Several consistent trends can be identified during training. First, training begins with large negative rewards, driven by excessive merchant investments beyond system requirements. This oversupply results in persistently low prices, limiting the profits agents can extract from the market. From this initial condition, agents gradually reduce their investments until aggregate profits turn positive. At that point, they typically continue reducing investments, increasing the profitability of their installed assets, until scarcity conditions emerge. In the last stages of training, agents refine their strategies, filtering profitable from unprofitable technologies for investment. Figures \ref{Appendix - fig: Training Behavior - EoM} to \ref{Appendix - fig: Agent training - CM + CfD} further describe the training behaviors of test cases. 

\subsection{Hyperparameter Selection}
\label{SubSection - Hyperparameter Selection}

A Hyperparameter Selection using a complementary set of metrics to assess performance is carried out using the testing runs presented previously. The first of the metrics is the aggregated system reward, interpreted as the net present value of the profit accumulated by all agents in the simulation and given by the sum of expression \ref{eq: objective function} for all agents. It has been observed that a system with well-trained agents would exhibit a relatively low, but positive, aggregated reward. This is because, while individual agents try to maximize their profits, they should be limited by market interactions with their competitors. Moreover, periods with excessive scarcity events, those behind large profit spikes in the simulation, would also open the opportunity for additional investments in the system. Finally, none of the agents are obliged to participate in market sessions. Consequently, if profits harnessed from the system are consistently negative, agents can stop investment altogether, thus mitigating their losses. 

Nonetheless, a relatively low aggregated reward could stem from various factors and conditions, such as inefficient/random agent behavior or uneven distribution of profits across agents. As a result, relying solely on the aggregated reward could be insufficient to assess the hyperparameter configurations properly. Therefore, three additional metrics are proposed to facilitate comparison and evaluation across runs: 

\begin{itemize}
    \item \textbf{Penalty:} Following the auxiliary reward function implemented in \cite{harder_fit_2023}, a metric, initially conceptualized as a penalty to be used during training, was devised to quantitatively measure the distance between the agent's decisions and the market equilibrium. The penalty aims to represent the net present value that can be obtained by projects participating in the market in a given system condition. This profitability is based on agents' actions and virtual plants and is calculated per agent, market, and technology.
    \item \textbf{HHI index}: In its current version, the environment models relatively homogeneous GENCOs, differentiated only by their existing assets and enabled investments. Thus, it is reasonable to expect that, if training is successful across market participants, most of them should be able to place investments in the system. As a proxy to measure uniformity across agents, the Herfindahl–Hirschman Index (HHI) \cite{us_department_of_justic_merger_2023} is calculated using the installed capacity at the end of the simulation.
    \item \textbf{League Ranking:} Inspired by the League-based training used in \cite{vinyals_grandmaster_2019}, a league-based evaluation of the hyperparameter configurations is implemented. In the league, market simulations are launched using, for each agent, a random selection of networks \textit{(e.g. Agent 1 uses networks from run M.6, Agent 2 from run L.1, and so on)}. After one episode is completed, the agent's accumulated rewards are recorded. By repeatedly randomizing network selection and running multiple episodes, it is possible to compare the performance of hyperparameter configurations by directly using the agents' reward.
\end{itemize}

\begin{figure}[hbt!]
    \centering
    \includegraphics[width=0.9\linewidth]{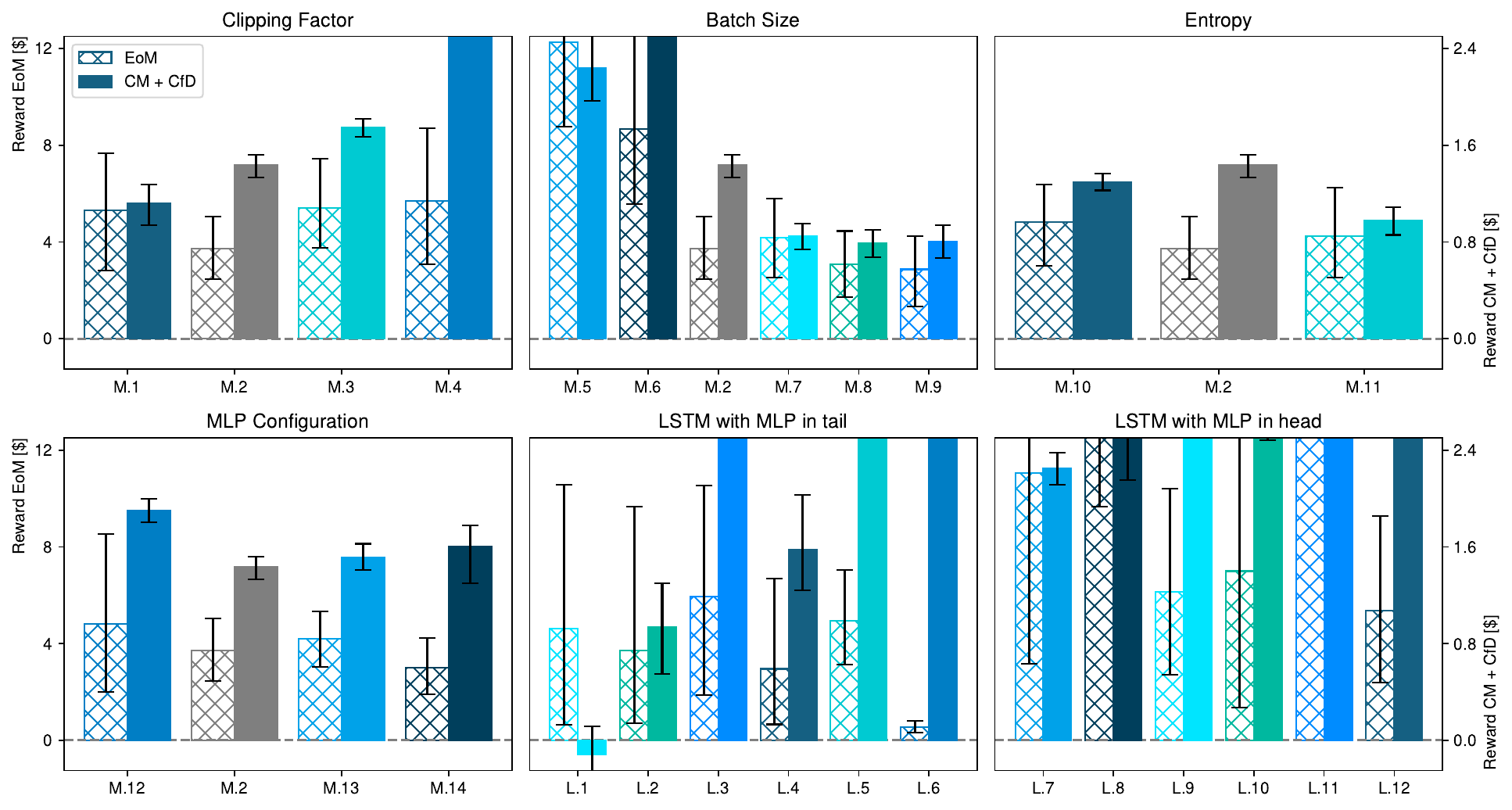} %
    \caption{Average Aggregated Reward for different hyperparameter configurations in the EoM and CM + CfD environments.  Results are obtained using 100 episodes in the environment with the most updated agents' versions. Hatched bars indicate the outcome for the EoM environment, while solid bars represent the CM + CfD environment.  Error bars showcased the 10th and 90th percentiles from the 100 episodes. The Y axes in the Figure are adjusted to facilitate comparison among the most relevant hyperparameter configurations.}
\label{fig: Aggregated Reward during training.}
\end{figure}

To compare results among runs, market simulations are carried out using the most updated agents for each particular hyperparameter configuration to evaluate system performance, given the lack of stopping criteria in the algorithm. Starting from the aggregated reward, as displayed in Figure \ref{fig: Aggregated Reward during training.}, results show that hyperparameter selection and network configuration significantly impact the learning process, system interactions, and outcomes. Among the hyperparameter selection, it can be noted that increasing the clipping factor while reducing the batch size leads to higher aggregated rewards. In contrast, modifying entropy and MLP configurations results in negligible changes. On the other hand, LSTM-based configurations exhibit greater fluctuations in aggregated reward, likely due to a more unstable learning process, as illustrated in Figures \ref{Appendix - fig: Training Behavior - EoM} to \ref{Appendix - fig: Agent training - CM + CfD}.

\begin{figure}[hbt!]
    \centering
    \includegraphics[width=1\linewidth]{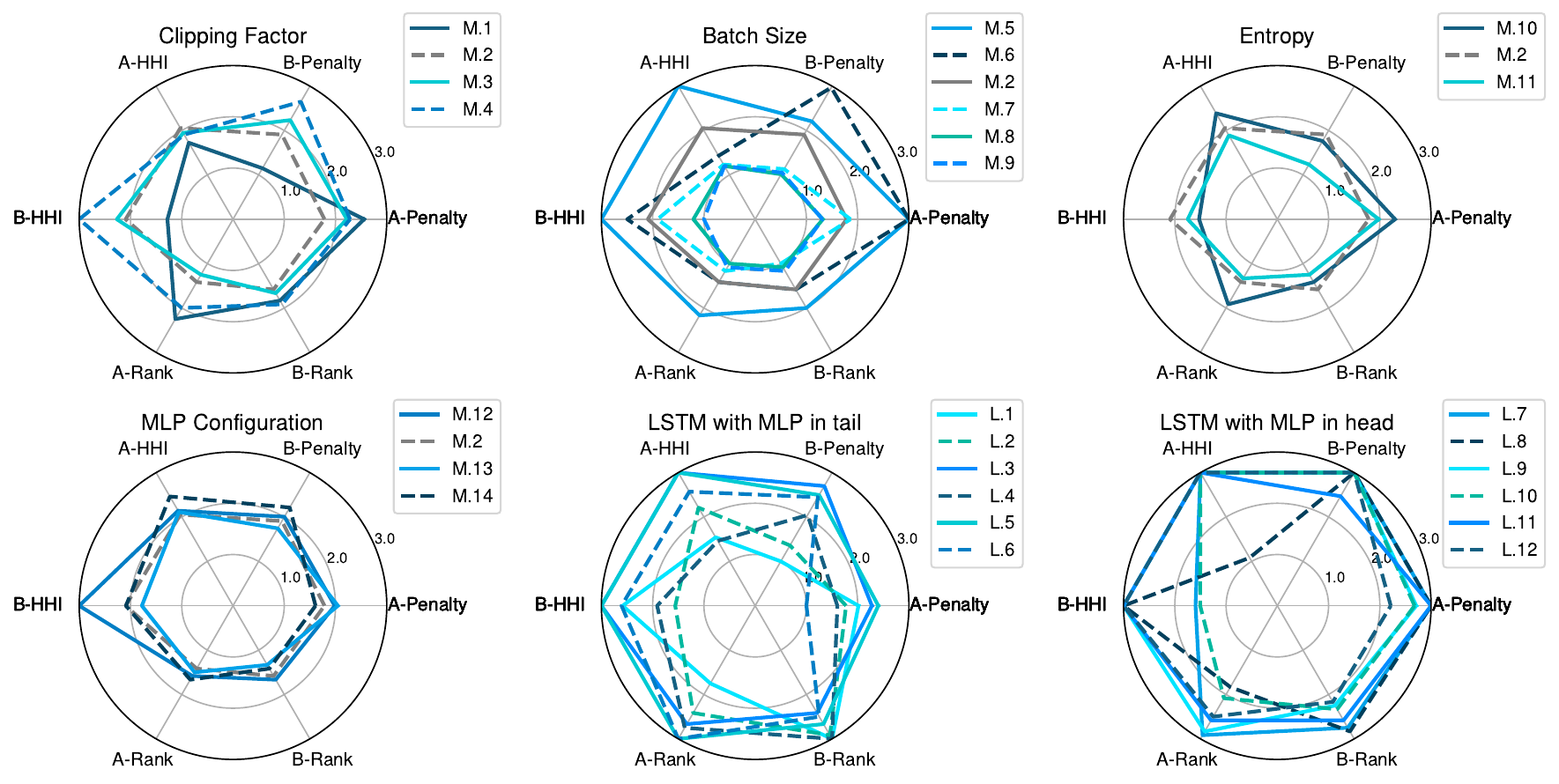} %
    \caption{Average Penalty, HHI index, and League Ranking for hyperparameter configurations in the EoM \textit{(A)} and CM + CfD \textit{(B)} environments. Penalty and HHI index Results are obtained using 100 episodes in the environment with the most updated agents' versions. League Ranking is obtained from the competition set between all hyperparameter configurations per environment, where agents rank according to their overall performance between 1 \textit{(best)} and 26 \textit{(worst)}. To facilitate visualization in the polar plot, all metrics undergo zero-centered median normalization, are clipped between 0 and 2, and are shifted by one unit.}
\label{fig: Additional metrics.}
\end{figure}

Figure \ref{fig: Additional metrics.} presents the application of the additional evaluation metrics to the hyperparameter runs. In the polar plot, runs/curves approaching the unit circle are the best performers. These results confirm the positive impact of increasing the Batch Size in the PPO algorithm and corroborate the erratic behavior of all LSTM configurations tested. Regarding other parameters evaluated, the tests are less conclusive. Yet, a slight advantage, especially in the League Ranking, is observed for hyperparameter configurations with intermediate Clipping Factors and higher Entropy terms. Further details regarding the metrics are provided in Figure \ref{Appendix - fig: Penalty and HHI for hyperparameter configurations in the EoM and CM + CfD environments.}, for the Penalty and HHI index, and in Table \ref{Appendix - Table: League competition results.}, for the League Ranking. 

\begin{figure}[hbt!]
    \centering
    \includegraphics[width=0.8\linewidth]{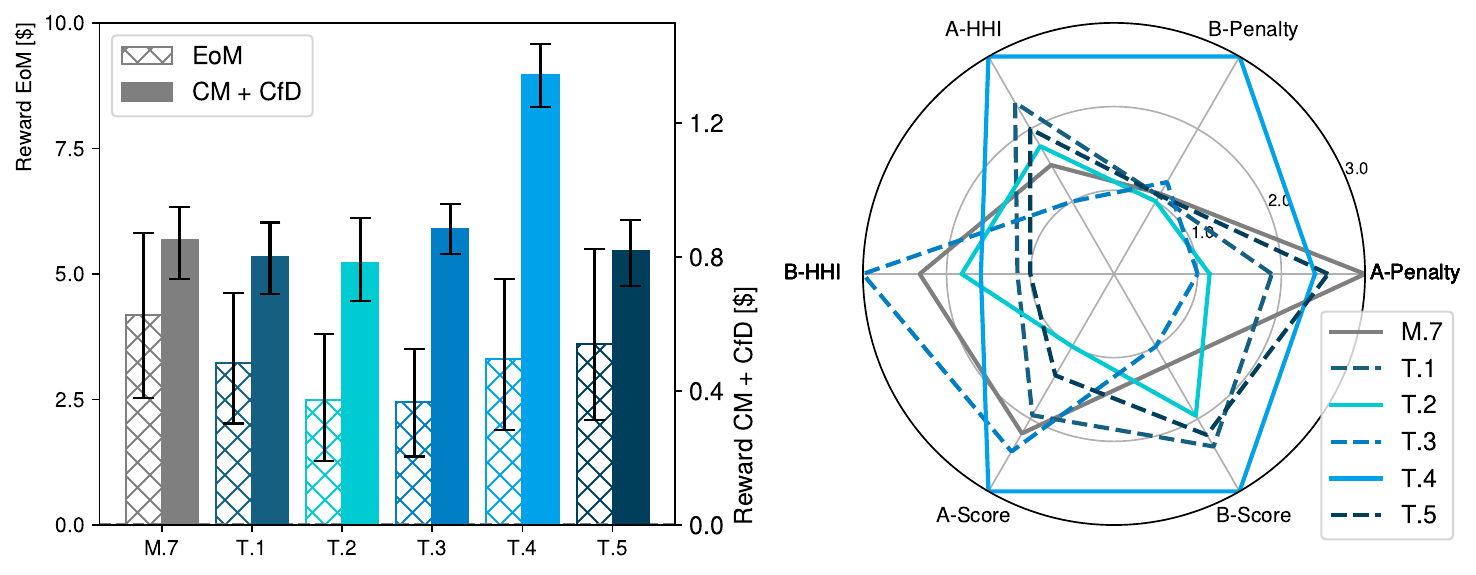} %
    \caption{Average Penalty, HHI index, and League Ranking for hyperparameter configurations from the ablation study in the EoM \textit{(A)} and CM + CfD \textit{(B)} environments. Penalty and HHI index Results are obtained using 100 episodes in the environment with the most updated agents' versions. League Score is obtained from the competition set between the hyperparameter configurations from the ablation study per environment, where agents are scored according to their overall performance between 0 \textit{(best)} and 1 \textit{(worst)}. To facilitate visualization in the polar plot, all metrics undergo zero-centered mean normalization, are clipped between 0 and 2, and are shifted by one unit.}
\label{fig: Additional metrics ablation.}
\end{figure}

Based on previous assessments, an additional ablation study uses the hyperparameter configurations presented in Table \ref{Table: Tests Conducted During Hyperparameter Search - Ablation.}. Figure \ref{fig: Additional metrics ablation.} presents a consolidated view of aggregated rewards and complementary evaluation metrics, where comparisons are made exclusively among these relatively well-performing configurations. The results highlight the importance of the Batch Size in the algorithm, as it is the only hyperparameter that differs substantially across configurations. In contrast, other configurations exhibit similar performance, making it challenging to determine an optimal choice based on the selected evaluation criteria. Ultimately, configuration T.1 is chosen for all subsequent tests in this study. This decision is based on its relatively large network, which enhances the expressiveness of agent actions, its higher entropy, which led to a slight performance improvement across all tests by improving exploration, and its intermediate clipping factor. Nonetheless, no significant changes would be expected if any configurations from Table \ref{Table: Tests Conducted During Hyperparameter Search - Ablation.}, except for T.4, were to be used instead. Importantly, the selected Batch Size enables simulations of up to 40 agents in the tested hardware.

\begin{table}[hbt!]
\centering
\resizebox{0.9\columnwidth}{!}{
\begin{tabular}{|l|c|c|c|c|}
\hline
\textbf{Code/Parameter} &
  \multicolumn{1}{c|}{\textbf{Clipping factor}} &
  \multicolumn{1}{c|}{\textbf{Batch Size}} &
  \multicolumn{1}{c|}{\textbf{Entropy}} &
  \multicolumn{1}{c|}{\textbf{MLP configuration}} \\ \hline
\textit{\textbf{M.7}}  & 0.05 & 35328 & 0.000001 & [256-256] \\ \hline
\textit{\textbf{T.1}} & 0.1  & 35328 & 0.01     & [512-512] \\ \hline
\textit{\textbf{T.2}} & 0.1  & 35328 & 0.01     & [256-256] \\ \hline
\textit{\textbf{T.3}} & 0.1  & 35328 & 0.000001 & [512-512] \\ \hline
\textit{\textbf{T.4}} & 0.1  & 17664 & 0.01     & [512-512] \\ \hline
\textit{\textbf{T.5}} & 0.05 & 35328 & 0.01     & [512-512] \\ \hline
\end{tabular}%
}
\caption{Tests and hyperparameters used in the ablation study. For individual tests, hyperparameters not mentioned in the corresponding field are not modified, and the values from Table \ref{Appendix - Table: List of hyperparameters used during search.} are used instead. }
\label{Table: Tests Conducted During Hyperparameter Search - Ablation.}
\end{table}

Importantly, the hyperparameter search conducted in this section, the primary source of computational burden in this work, is not required for the practical use of the MARL model. With the selected hyperparameters, training sessions typically reach stability in the main trackable metrics within 20–40 hours, depending on the number of agents in the system. Once agents are trained, market simulations can be performed with minimal time and computational requirements, partially offsetting the high training costs. Future work could explore methods that allow a single training setup to adapt effectively to a range of policy parameters \textit{(carbon taxes, auction price caps, and other policy options)}, in addition to other methodological improvements that enhance the computational efficiency in the current framework.

\section{Long-term electricity Market Results}
\label{Section - Market results}

Considering the modeling framework and the hyperparameter search presented previously, this Section applies the MARL model to a system inspired by the Italian electricity system. The analysis aims to showcase the models' capabilities while dynamically representing decarbonization pathways under different market designs, policy scenarios, and competition levels. To this end, the Italian electricity system provides a suitable baseline, given its continued reliance on fossil-fuel-based technologies alongside a relatively high penetration of RES. Nonetheless, the analysis can be readily extended to other systems with comparable characteristics \textit{(energy mix, decarbonization policies, demand growth expectations, among others)}.

\begin{figure}[hbt!]
    \centering
    \includegraphics[width=1.0\linewidth]{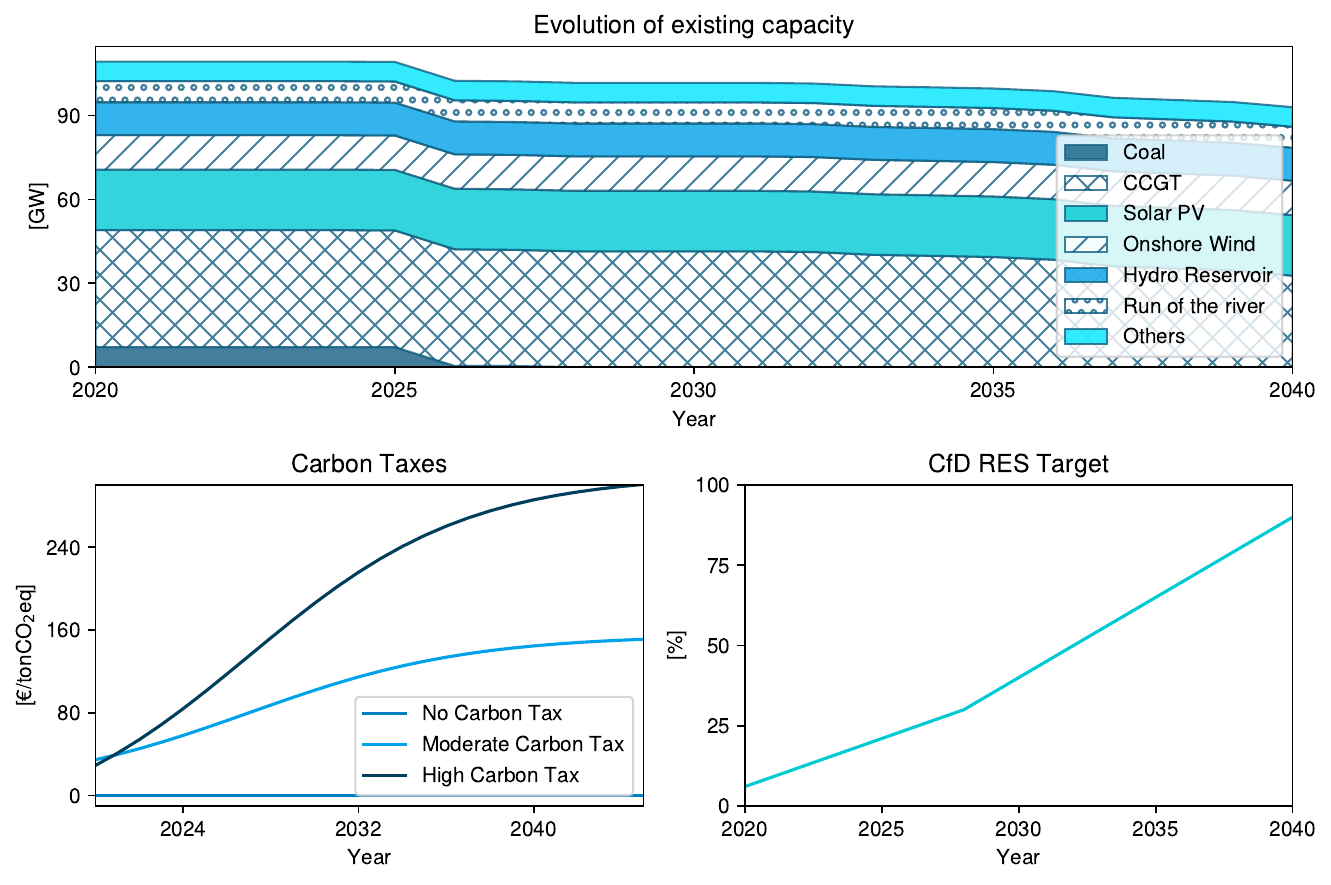} %
    \caption{Existing conditions and main policy assumptions for the study cases. The upper panel showcases the evolution of existing assets, where the key factor is coal power plants decommissioning before 2030.  The lower left panel presents the Carbon Tax scenarios, while the lower right one displays the RES penetration target to be achieved with the Contracts-for-Difference Market.}
\label{fig: Existing conditions.}
\end{figure}

For the scenarios, the period between 2020 and 2040 is selected, using the starting conditions regarding installed capacity and the policy scenarios briefly summarized in Figure \ref{fig: Existing conditions.}. Complementary, demand scenarios are obtained from planning exercises carried out with a central planning energy model (PyPSA-Eur), accounting for the electrification of the energy sector and discounting electricity imports from total consumption. This exercise results in an average demand growth rate of close to 2\% for the period under study. Besides these conditions, the system abides by the characteristics introduced in Section \ref{Section - Long-term Electricity Market Environment}. Importantly, the planning carried out with PyPSA avoids lost-load events. 

Initially, 12 scenarios that combine market designs and competition levels are evaluated. On the long-term market design side, four regulatory frameworks are tested; an Energy-only-Market \textit{(EoM)}, a market combining features from the EoM with CfD auctions to meet a RES target \textit{(CfD)}, a Capacity Market to ensure system adequacy \textit{(CM)}, and simultaneous implementation of a Capacity Market and CfD auctions \textit{(CM+CfD)}. Regarding competition levels, the market designs are tested on environments with 8 \textit{(6 incumbents, 2 entrants)}, 16 \textit{(8 incumbents, 8 entrants)}, and 32 \textit{(16 incumbents, 16 entrants)} agents. For each of the previous configurations, existing assets are allocated to incumbent agents, resulting in HHI coefficients of 3000, 1880, and 1000, respectively, thus representing markets with high, moderate, and low levels of market concentration. Furthermore, the decommissioning shown in Figure \ref{fig: Existing conditions.} is applied uniformly to all incumbents. The maximum investment per technology in the system is maintained constant and divided equally across agents, to allow the number of agents and the emerging competition among them to be driving factors in the simulations. Consequently, the limit for generation technologies and short-term storage, per market and year, is set to 8 GW and 1.6 GW, respectively, which in both cases, as an aggregate, is a comfortable upper bound for the requirements imposed by demand growth. To all previous simulations, a “Moderate Carbon Tax" rising from current levels in the Emission Trading Scheme to around 150\texteuro/tCO2 is applied. This carbon tax schedule reflects a rising ambition in decarbonization plans, though not at the levels expected to reach the EU and Italy's 2040 and 2050 climate targets. Thus, we carry out sensitivity with a higher carbon tax schedule.  Finally, a computational budget of 16, 20, and 32 hours is used for training scenarios with 8, 16, and 32 agents. 

\begin{figure}[hbt!]
    \centering
    \includegraphics[width=1.0\linewidth]{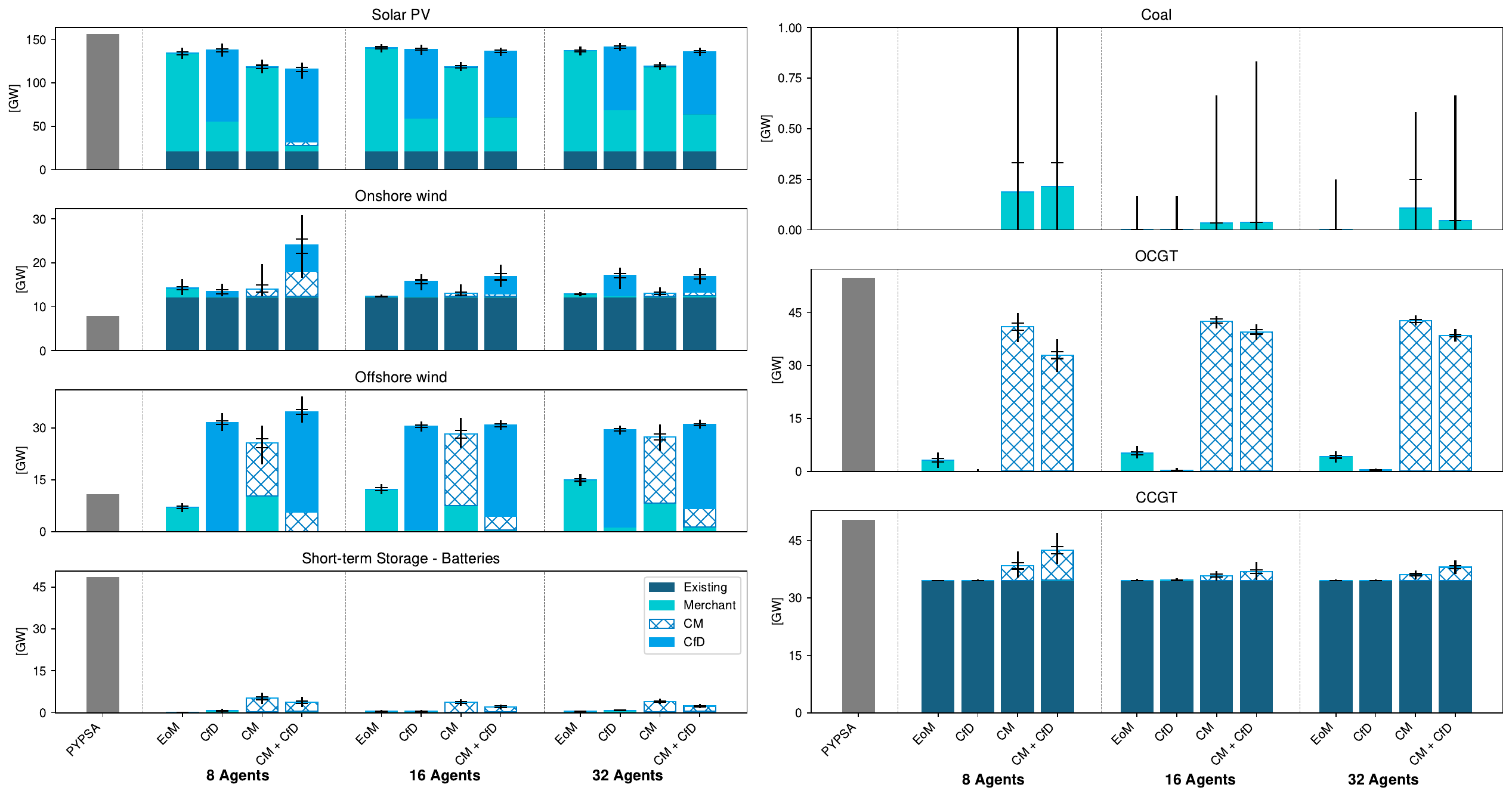} %
    \caption{Installed Capacity of generation and storage technologies in 2040 under different market designs and competition levels. Stacked bars indicate average installed capacities across runs. In each stacked bar, the horizontal markers display the 25th and 75th percentiles, and the vertical markers minimum and maximum values. Columns organize the installed capacities in the four market designs. Furthermore, results are grouped in the column categories, divided by vertical dotted gray lines, according to the number of agents used in the simulation. Different hatching highlights the mechanism used by agents to enter the market. Gray bars, for the corresponding technology, show the output from the central planning exercise carried out in PYPSA. Results from the MARL model are obtained across 200 market simulations using the trained agents.}
\label{fig: Installed Capacity 2040.}
\end{figure}

To analyze market outcomes, Figure \ref{fig: Installed Capacity 2040.} introduces the installed capacity at the end of the study period, selected as 2040. For the EoM scenarios, expansion is driven by solar, offshore wind, and Open-Cycle Gas Turbines (OCGT), while Combined-Cycle Gas Turbines (CCGT) are limited to a few GWs, and Coal plants are negligible. Except for OCGT and short-term storage, these results are relatively close to the central planning exercise carried out with the PYPSA optimization framework, which has been harmonized by sharing, when possible, input parameters, technical assumptions, and the main energy policy with the current implementation of the MARL model. In general, PyPSA allocates greater capacity to dispatchable technologies such as OCGT and CCGT and storage solutions such as batteries. This outcome is largely attributable to its high temporal resolution, hourly representation across all 8760 hours of the year. While the total installed capacity of variable renewable energy sources is broadly comparable across frameworks, PyPSA typically results in a higher share of solar power, reflecting its assumption of progressively declining investment costs for this technology over the modeled period, in addition to the incentives emerging from centralized planning for storage technologies, in contrast to the decentralized market signals in the MARL model. A more detailed discussion of the characteristics of PyPSA and its contrasts with MARL, which account for the observed differences in results, is provided in \ref{Appendix - PyPSA description}.

The energy outcomes from the MARL simulations change under different market designs. In the case when the CfD market is introduced as the only long-term mechanism (CfD scenario), offshore wind penetration increases considerably, while solar penetration remains relatively constant. Yet, the entry mechanism chosen by agents for solar investments shifts from merchant investment towards long-term auctions.  For the case in which the capacity market is the long-term mechanism instead (CM scenario), offshore wind similarly enters the market through these auctions. Still, the main effect of the capacity market is the increase in the share of OCGT in the mix, bringing it closer to the planning model. This similarity can be explained by the fact that the planning model rewards the flexibility of OCGT not through markets, but through the explicit constraint imposed in the system to avoid lost load events. When both long-term auctions are combined (CM + CfD scenario), the composition of the generation mix resembles the case in which only the Capacity Market was in place. However, RES shifts investments from other mechanisms to the long-term CfD auctions. In terms of installed capacity, no significant deviations occurred depending on the number of agents in the simulation.

\begin{figure}[hbt!]
    \centering
    \includegraphics[width=0.7\linewidth]{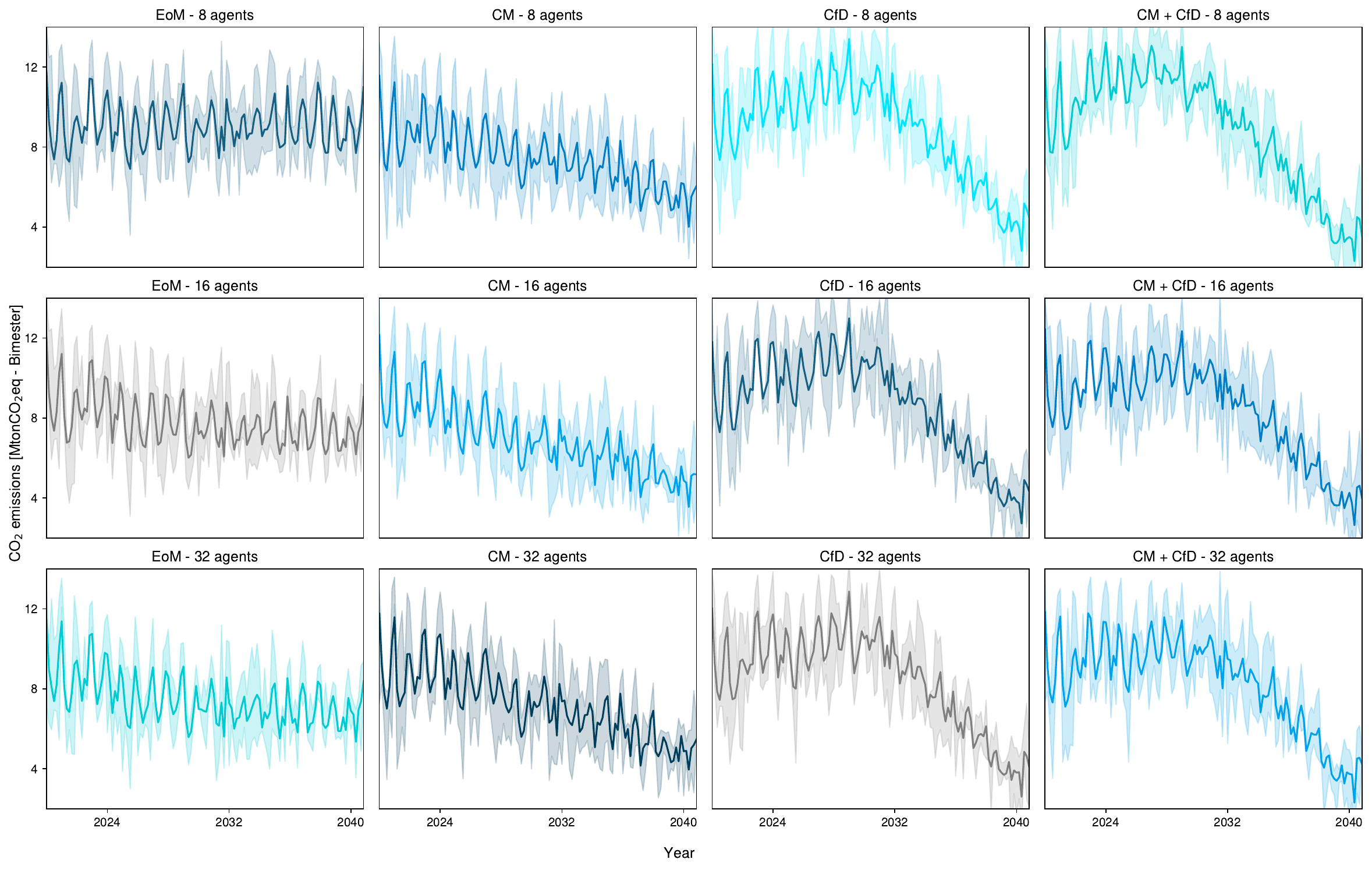} %
    \caption{Bi-monthly \(CO_2\) emissions under different market designs and competition levels. Emissions are obtained using the electricity generated per technology, and its corresponding emission factor. In the plot, rows indicate the number of agents used in the simulation, while columns correspond to the four market designs. Solid lines display average values, while shaded areas are the 25th and 75th percentiles. Results are obtained across 200 market simulations using the trained agents.}
\label{fig: Emissions.}
\end{figure}

\begin{figure}[hbt!]
    \centering
    \includegraphics[width=0.7\linewidth]{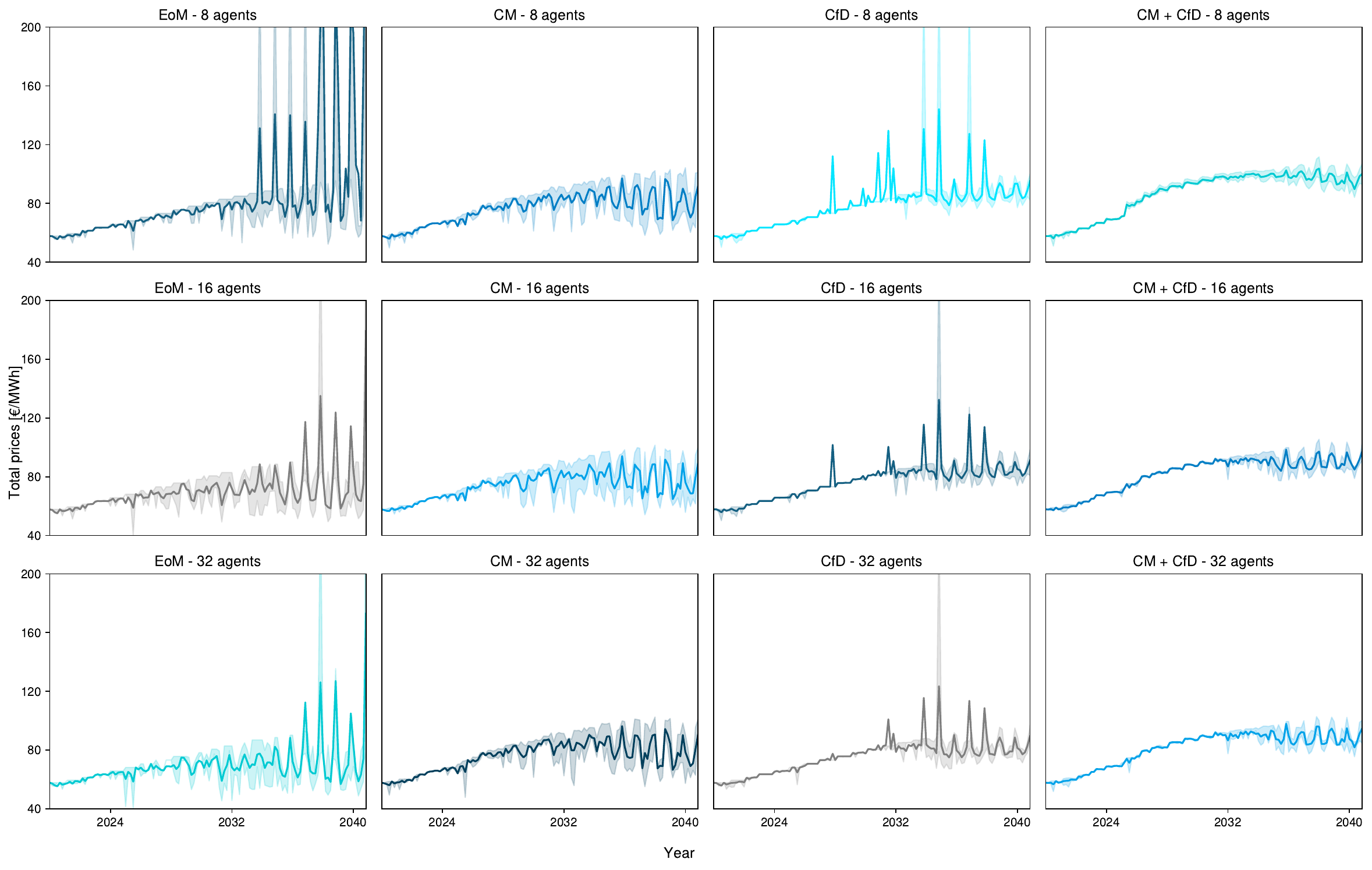} %
    \caption{Total system prices under different market designs and competition levels. Prices are calculated as the net price incurred by demand, including the short-term prices, the premiums from the capacity and contract for difference markets, and any financial settlement derived from these mechanisms. In the plot, rows indicate the number of agents used in the simulation, while columns correspond to the four market designs. Solid lines display average values, while shaded areas are the 25th and 75th percentiles. Results are obtained across 200 market simulations using the trained agents.}
\label{fig: Electricity Prices.}
\end{figure}

To capture the dynamics of the transition toward the generation mix previously described, Figures \ref{fig: Emissions.} and \ref{fig: Electricity Prices.} present the evolution of \(CO_2\) emissions and electricity prices over the study period. Regarding emissions, we find significantly different outcomes depending on the choice of market designs. In addition to the decommissioning of fossil fuel assets and the competitive costs of renewable technologies, particularly solar PV, we find no significant emission reductions by 2040 in the EoM cases, regardless of the level of competition. In contrast, when the long-term mechanisms of the CfD and CM market designs are introduced, with the subsequent increase in total installed capacity illustrated in Figure \ref{fig: Installed Capacity 2040.} from all technologies, but especially RES, emissions are substantially reduced. Notably, the three market designs that incorporate long-term mechanisms (CfD, CM, and CM+CfD) produce similar outcomes in terms of emissions in 2040, mainly due to the opportunities created by both CfD and Capacity Market auctions for high shares of offshore wind deployment. Nonetheless, emission trajectories are more dynamic, with agents in the CfD market designs delaying merchant investments before 2030, before entering the market in mass once the RES target showcased in \ref{fig: Existing conditions.} becomes more stringent.

In terms of electricity prices, two key trends emerge. First, the EoM design is characterized by frequent scarcity events, particularly toward the end of the simulation period. These events, which lead to price spikes and load curtailment, become more prevalent under lower levels of competition. Although CfD auctions are not explicitly designed to address scarcity, they partially mitigate these events and reduce their frequency. In contrast, Capacity Market designs prevent scarcity events altogether during the simulation. This outcome suggests that the Capacity Market's current design procures more adequate services than strictly necessary in an efficient design, especially considering that such markets are intended to meet a Loss of Load Expectation (LOLE) between 3-4 hours per year \cite{department_of_energy_security__net_zero_exploring_2023}. Further details on electricity prices in the short-term market for 2040 and the behavior of CfD and CM premiums are presented in Appendix Figure \ref{Appendix - fig: Detailed prices 2040.}. Crucially, concurrently implementing CfD and CM auctions leads to increased CM premiums. This result is consistent with the exacerbated \textit{Missing Money} problem arising from the additional renewable energy deployment incentivized by CfD schemes. 

To further explain the investment behavior of agents, Table \ref{Table: IRR Agents.} presents the aggregated Internal Rate of Return (IRR) and installed capacities by agent type across simulations. In contrast, Table \ref{Appendix - Table: IRR Techs.} offers more detailed results disaggregated by technology and market design. A key observation is that, in all scenarios, the aggregate IRR exceeds the 8\% discount rate applied in the environment. Nonetheless, the implicit discount rates decreased as the number of agents and the level of competition increased. However, this attenuation effect was not substantial and was less pronounced in the transition from 8 to 16 agents compared to the transition from 16 to 32 agents. These patterns can possibly be attributed to three factors: the emergence of risk-averse strategies during training, limited competition and/or potential collusion induced by the learning algorithm, and the interaction between market design and model input parameters.

\begin{table}[hbt!]
\centering
\resizebox{\textwidth}{!}{%
\begin{tabular}{l|cl|cccccccccccc|}
\cline{2-15}
 &
  \multicolumn{2}{c|}{\multirow{3}{*}{\textbf{Agent Type}}} &
  \multicolumn{12}{c|}{\textbf{Number of Agents and Market design}} \\ \cline{4-15} 
 &
  \multicolumn{2}{c|}{} &
  \multicolumn{4}{c|}{\textbf{8}} &
  \multicolumn{4}{c|}{\textbf{16}} &
  \multicolumn{4}{c|}{\textbf{32}} \\ \cline{4-15} 
 &
  \multicolumn{2}{c|}{} &
  \multicolumn{1}{l|}{\textbf{EoM}} &
  \multicolumn{1}{l|}{\textbf{CfD}} &
  \multicolumn{1}{l|}{\textbf{CM}} &
  \multicolumn{1}{l|}{\textbf{CM + CfD}} &
  \multicolumn{1}{l|}{\textbf{EoM}} &
  \multicolumn{1}{l|}{\textbf{CfD}} &
  \multicolumn{1}{l|}{\textbf{CM}} &
  \multicolumn{1}{l|}{\textbf{CM + CfD}} &
  \multicolumn{1}{l|}{\textbf{EoM}} &
  \multicolumn{1}{l|}{\textbf{CfD}} &
  \multicolumn{1}{l|}{\textbf{CM}} &
  \multicolumn{1}{l|}{\textbf{CM + CfD}} \\ \cline{2-15} 
 &
  \multicolumn{1}{c|}{\multirow{2}{*}{\textbf{Incumbents}}} &
  \textbf{IRR [\%]} &
  15.03 &
  16.63 &
  11.47 &
  \multicolumn{1}{c|}{16.21} &
  11.38 &
  15.51 &
  10.90 &
  \multicolumn{1}{c|}{14.39} &
  10.59 &
  14.53 &
  11.47 &
  14.11 \\ \cline{3-3}
 &
  \multicolumn{1}{c|}{} &
  \textbf{Share of investments [\%]} &
  72.52 &
  77.34 &
  79.44 &
  \multicolumn{1}{c|}{76.40} &
  57.65 &
  56.88 &
  54.44 &
  \multicolumn{1}{c|}{57.94} &
  58.29 &
  56.76 &
  52.29 &
  53.80 \\ \cline{2-3}
 &
  \multicolumn{1}{c|}{\multirow{2}{*}{\textbf{Entrants}}} &
  \textbf{IRR [\%]} &
  14.19 &
  16.47 &
  11.43 &
  \multicolumn{1}{c|}{17.48} &
  10.69 &
  16.38 &
  10.78 &
  \multicolumn{1}{c|}{15.38} &
  10.44 &
  15.13 &
  11.28 &
  14.15 \\ \cline{3-3}
 &
  \multicolumn{1}{c|}{} &
  \textbf{Share of investments [\%]} &
  27.48 &
  22.66 &
  20.56 &
  \multicolumn{1}{c|}{23.60} &
  42.35 &
  43.12 &
  45.56 &
  \multicolumn{1}{c|}{42.06} &
  41.71 &
  43.24 &
  47.71 &
  46.20 \\ \cline{2-15} 
\end{tabular}%
}
\caption{Internal Rate of Return of investments and Share of Investments for Incumbent and Entrant agents under different market designs and competition levels. The Internal Rate of Return is calculated as an aggregate for all investments carried out for the agent during the simulation. The Share of Investments for agents is calculated using the installed capacity [MW] across generation and storage technologies. }
\label{Table: IRR Agents.}
\end{table}

Regarding the environment's design, it is essential to highlight that when investments fail to yield profits, agents quickly learn to avoid investing altogether, since there is no explicit penalty for abstaining from market participation. As a result, agents tend to adopt strategies where they limit their investment commitments to opportunities with near-guaranteed profitability. Such behavior, which translates to agents requiring a positive difference between the implicit internal rate of return derived from their investments and the exogenous discount rate used to calculate the net present value of their portfolios, is commonly associated with risk aversion \cite{cochrane_asset_2005}. This occurs despite the environment optimizing a risk-neutral objective per agent, as described in Section \ref{Section - Multi-Agent Reinforcement Learning applied to the Market Environment}. Regarding the second factor, Section \ref{Section - Training and Validation} shows that the system quickly converges to an equilibrium, with limited exploration or further improvement in agent strategies. This stagnation may promote implicit collusion, as market outcomes become relatively stable. As discussed in \cite{vinyals_grandmaster_2019}, addressing this issue typically requires multiple training restarts to encourage exploration, leaving a stream for future work. Finally, the relatively high endogenous discount rates observed among agents can be linked to active constraints in the model. This is particularly noticeable in the case of solar PV, where investments in 2040 approach the maximum allowable investment level for the study period. 

\begin{figure}[hbt!]
    \centering
    \includegraphics[width=1.0\linewidth]{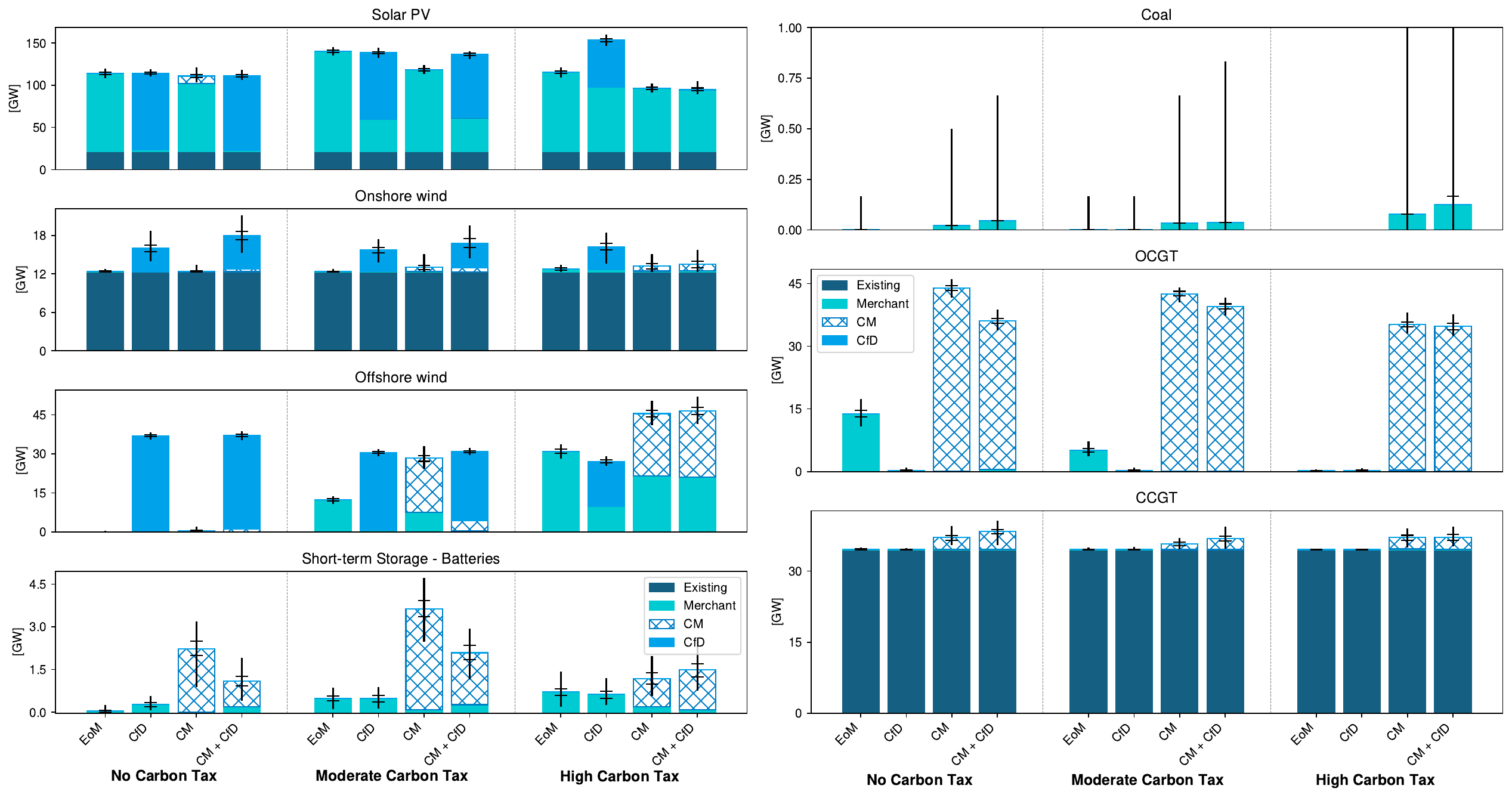} %
    \caption{Installed Capacity of generation and storage technologies in the year 2040 in simulations considering 16 agents and different carbon tax scenarios. Stacked bars indicate average installed capacities across runs. In each stacked bar, the horizontal markers display the 25th and 75th percentiles, and the vertical markers the minimum and maximum values. Columns organize the installed capacities in the four market designs. Furthermore, results are grouped in the column categories, divided by vertical dotted gray lines, according to the carbon tax used in the simulation. Moreover, hatching highlights the mechanism used by agents to enter the market. Results from the MARL model are obtained across 200 market simulations using the trained agents.}
\label{fig: Comparison Installed Capacity 2040.}
\end{figure}

To study the previous hypothesis related to maximum investment quantities, but also to further understand and validate modeling outcomes, a comprehensive set of sensitivities and experiments is carried out. Detailed results are provided in \ref{Appendix - Supplementary information - Long-term Electricity Market Results}, with the following highlighting the main findings from the exercise:

\begin{itemize}
    \item \textbf{Carbon taxes:} Complementing the base case, implemented with the moderate Carbon Tax, the two additional Carbon Tax scenarios \textit{(No Tax and High Carbon Tax)} illustrated in Figure \ref{fig: Existing conditions.} were tested under different market designs. Additionally, the tests were carried out using 16 agents, representing an intermediate level of competition in the market. Installed capacities for the year 2040 under the Carbon Tax scenarios are presented in Figure \ref{fig: Comparison Installed Capacity 2040.}. Overall, the penetration levels of most technologies remain relatively consistent across scenarios, suggesting competitive low-carbon investments remain the dominant factor shaping the generation mix in the coming years. However, carbon taxes had a relevant effect for $CO_2$ emissions reductions, especially in the EoM and CM market designs. Moreover,  the high Carbon Tax scenario facilitates increased investment in offshore wind, resulting in a notable reduction in emissions.  This finding underscores the importance of analysing climate policies such as carbon taxes within the context of complementary policy instruments and market design, which have first-order consequences for the energy sectors and their decarbonization.
    \item \textbf{Maximum investment quantities:} In this sensitivity, the investment cap for solar PV and OCGT technologies doubled compared to the base scenario. The results show an increased share of solar PV in the mix, partially replacing offshore wind investments, especially within CfD auctions. Although this shift leads to a slight reduction in overall IRR, solar PV projects still yield high returns, indicating that agents are still maximizing the share of solar PV eligible for auction participation, which remains highly profitable. These results indicate that all the factors hypothesized to explain the difference between the IRR and the exogenous discount rate have an effect in explaining the observed market results. 
    \item \textbf{Market design:} Variations in auction price caps and market targets for CM and CfD markets are tested and compared to the base scenario \textit{(CM + CfD market environment with a 16-agent configuration, also used for the discount rate and robustness analysis)}. On the one hand, the results demonstrate that the model is less sensitive to auction prices than market targets, indicating that market dynamics and competition are expected to be the primary drivers of system outcomes. Yet, differences are observed for CfD auctions when the uptake of the RES target accelerates, denoting a tendency in the trained agents to approach the price-cap in the auctions in such cases. On the other hand, increasing the targets of these mechanisms leads to over-investment in the system with subsequent trade-offs in total system prices. Notably, changes in the adequacy target for the capacity market had limited effects on emissions, while higher RES penetration levels in CfD auctions yielded additional mitigation potential. Moreover, faster uptakes in the RES target yield higher emission reduction potential, especially around the year 2030, but also increase the pressure on system prices and reduce incentives for merchant investments. These findings underscore the critical need for adaptive decision-making in market design, as the current implementation maintains a fixed market structure. Thus, the analysis also showcases the opportunity to model complementary incentives for short-term storage and other flexibility sources, which maintained a limited participation through the capacity market across sensitivities, remaining the largest differentiating factor with the central-planning model.
    \item \textbf{Discount rates:} A range of exogenous discount rates, uniformly applied across all agents in the system, was tested, leading to three main trends. First, higher discount rates resulted in a rapid shift away from merchant investments, with the subsequent utilization of long-term auctions to meet demand requirements. However, the higher auction utilization and increased prices obtained through these mechanisms resulted in substantially higher system costs. Finally, low discount rate scenarios aligned with rapid decarbonization trends through faster and higher RES investments, with notable emission reduction potentials around 2030. These results further validate the model and reiterate the critical importance of low financing costs for achieving net-zero targets, as highlighted by \cite{calcaterra_reducing_2024, schmidt_adverse_2019}. 
    \item \textbf{Robustness analysis:} Using a base scenario, the MARL training setup was repeated multiple times. Across sessions, the main training metrics and key market variables, such as system prices, emissions, and total installed capacity, remained largely unchanged. However, some degree of substitutability was observed in the technologies selected by agents. From a methodological perspective, this assessment highlights that for in-depth scenario analysis, retraining sessions serve as an additional measure to evaluate the type of solution obtained by the model. From a market perspective, the emergence of different technology portfolios achieving similar aggregate results, even in a relatively static and risk-neutral setting, underscores the opportunity to enhance active planning approaches where the market design considers not only a purely risk-neutral economic perspective but also incorporates differentiated metrics such as resilience to shocks, diversification levels, and other system characteristics to promote and differentiate portfolios. 
\end{itemize}

\section{Conclusions and Future Work}
\label{Section - Conclusions}

With the advent of MARL, recent advances in algorithmic development and increased hardware availability, new frontiers in modeling capabilities are opening across a wide range of applications. Building on these advancements and drawing from previous efforts applying MARL in the electricity sector, this work offers new steps toward applying MARL methodologies to long-term electricity market design. This work helps close the gap between agent-based and partial-equilibrium modeling paradigms, two of the most common methods currently used for long-term market design and assessment. In this sense, the MARL model introduced, implemented, and tested presents an alternative tool for policymakers to test, evaluate, and refine, or even radically redesign, regulatory frameworks in response to the challenges of the energy transition. 

The main results, and their sensitivities, highlighted important caveats regarding the role of long-term markets in stringent decarbonization pathways. Long-term electricity markets provided a pathway for the system to achieve investment levels sufficient to meet demand requirements while maintaining consistent decarbonization trends. This feature remained consistent even under policy misalignment  (for example, in the absence of a carbon tax) or during market pressure (as in scenarios with elevated discount rates). Moreover, a comparison with a centralized planning model indicates that current market structures provide insufficient incentives for flexibility technologies such as short-term storage. However, while the large-scale deployment of storage solutions is expected to generate system-wide benefits, their remuneration within the MARL market framework remains underdeveloped, a market failure commonly observed in real-world systems. This gap underscores the potential value of establishing formal long-term flexibility markets, thus informing regulatory decisions on this issue.

A key strength of MARL in the context of electricity market design is the modeling flexibility that it would offer to practitioners, market actors, and eventually, to policymakers. The trained agents demonstrate the ability to respond to various regulatory configurations while acting and competing against other agents in the system. Particularly noteworthy is the simultaneous evaluation of multiple long-term expansion mechanisms, which require agents to assess investment profitability under market conditions that influence short- and long-term cash flows. More importantly, the model outcome is obtained with minimum assumptions, emerging almost endogenously from the training and model configuration. 

This flexibility, intrinsic to the MARL framework, enables exploring traditionally hard-to-model features in electricity systems, such as generator contracting strategies or bilateral agreements between utilities. Crucially, this flexibility could also be extended to future market designs for which no historical precedent or proxy for comparison exists. This includes hybrid market systems, where agents operate competitively within a centrally planned context that imposes objectives such as system adequacy or emissions reductions. As such, the proposed modeling setup is well-suited to future research efforts to bridge the gap between centralized planning and decentralized market-based decision-making, a key factor for achieving decarbonization targets in energy sectors worldwide. 

Nonetheless, the use of independent learning and Proximal Policy Optimization, an effective method for multi-agent competitive environments, requires careful consideration. In particular, appropriate hyperparameter selection and robustness analysis proved essential for enhancing the quality of the solutions. More importantly, the potential improvements that MARL offers within this modeling framework remain constrained by the inherent limitations of agent-based models. First, the model assumes rational decision-making by agents, which may not reflect real-world behavior. Second, the analytical scope is limited, excluding interactions with additional power sector stakeholders and other economic segments, such as the financing sector. Additionally, the model does not account for transformational and structural changes beyond the current market structure and electricity system organization, a significant limitation given the long-term nature of the analysis. These constraints, along with the broader limitations typical of energy and power sector models, are not directly addressed through the application of MARL.

Future research could explore MARL applications incorporating a broader and more diverse set of electric market agents and stakeholders. Notably, this includes prosumers, driven by digitalisation and democratisation trends, who are expected to play a central role in future energy systems by integrating supply and demand-side incentives into their decision-making. More broadly, the MARL framework can be extended to include active regulatory agents in the system. As demonstrated in prior studies, such a regulator can contribute to the strategic design of electricity markets in response to the strategic behavior of market participants. In particular, the regulator could be modeled as controlling key policy levers such as market caps and carbon pricing, and, specific to the model presented in this work, the targets and objectives of long-term auction mechanisms to pursue system-level goals. 

However, further development is required before this modeling approach can match the performance and realism of established electricity market models in the academic literature. Notably, the implementation lacks several key technical constraints essential for comprehensive market analysis, including unit commitment constraints, transmission system modeling, and a detailed representation of system flexibility requirements and associated markets. These aspects will be addressed in future work. In addition, while risk-averse behavior has emerged organically under specific configurations in the current setup, explicitly modeling agents' risk preferences through dedicated reinforcement learning algorithms designed to optimize risk-sensitive metrics would provide greater control and insight into strategic behavior under uncertainty. From an algorithmic standpoint, improvements are also needed to enhance the applicability of MARL in competitive settings. Large-scale implementations of MARL at the research forefront remain beyond the reach of most academic studies, even this one, which benefited from access to supercomputing resources. Advancements in MARL algorithms that improve convergence, facilitate exploration of alternative equilibria, and expand the diversity of agent responses in competitive environments would significantly strengthen this research. 

\section*{Declaration of generative AI in scientific writing}
During the preparation of this work, the authors used ChatGPT and Grammarly in the writing process to improve the readability and language of the manuscript. After using this tool/service, the author(s) reviewed and edited the content as needed and take full responsibility for the content of the published article
\section*{Funding}
Javier Gonzalez-Ruiz, Carlos Rodriguez-Pardo, and Massimo Tavoni acknowledge support from the European Union ERC Consolidator Grant under project No. 101044703 (EUNICE). Alice Di Bella acknowledges funding from European Union PNRR - Missione 4–Componente 2–Avviso 341 del 15/03/2022 - Next Generation EU, in the framework of the project GRINS - Growing Resilient, INclusive and Sustainable project (GRINS PE00000018 – CUP C83C22000890001). Iacopo Savelli received funding from the European Union’s Horizon Europe programme under the Marie Skłodowska-Curie grant agreement number 101148367. 

\bibliographystyle{elsarticle-num}
\bibliography{references.bib}
\clearpage
\appendix
\section*{Appendix}
\addcontentsline{toc}{section}{Appendix}

The supplementary material is divided into four sections sections. To start, \ref{Appendix - Long-term Electricity Market Environment} describes in detail the Long-term Electricity Market Environment, expanding upon Section \ref{Section - Long-term Electricity Market Environment} from the main document. \ref{Appendix - Supplementary information - Training} includes additional information regarding training and the hyperparameter Selection, complementing Section \ref{Section - Training and Validation} from the main document. \ref{Appendix - Supplementary information - Long-term Electricity Market Results} complements the electricity market results and the sensitivity analysis presented in Section \ref{Section - Market results} from the main document. Last, \ref{Appendix - PyPSA description} provides information regarding the PyPSA model, which is employed as a benchmark for evaluating the MARL electricity market, and complements this with additional information for the model comparison.
\clearpage
\section{Long-term Electricity Market Environment}
\label{Appendix - Long-term Electricity Market Environment}

This Appendix presents a thorough two-part description of the Long-term Electricity Market Environment. First, an overview of the long-term electricity market is provided, highlighting the main structural components, the information currently used to set up the simulations, and how the framework can be extended to new study cases. Second, the approach taken for implementing the Long-term electricity market into a Multi-Agent Reinforcement Learning environment is described, focusing on available modeling alternatives and relevant trade-offs in the options finally selected for the current version of the model. In short, the Appendix should be read as a translation from standard long-term electricity market models, especially Agent-based ones, to the MARL framework.

\subsection{Long-term electricity market}
\label{Appendix - Long-term electricity market}

The model presented in this work is targeted towards mid and long-term (\textit{several years or, utmost, a couple of decades}) decarbonization analysis in wholesale electricity markets. As such, most attention is dedicated towards utility-scale investment decisions and market mechanisms designed to promote them (\textit{e.g. Capacity Markets}), while giving lower priority to short-term operational considerations. Abiding by these principles, the following sections describe the building blocks of the long-term electricity market representation. 

\subsubsection{Generation Companies}
\label{Appendix - GENCOs}

\begin{figure}[h!]
    \centering
    \includegraphics[width=0.95\linewidth]{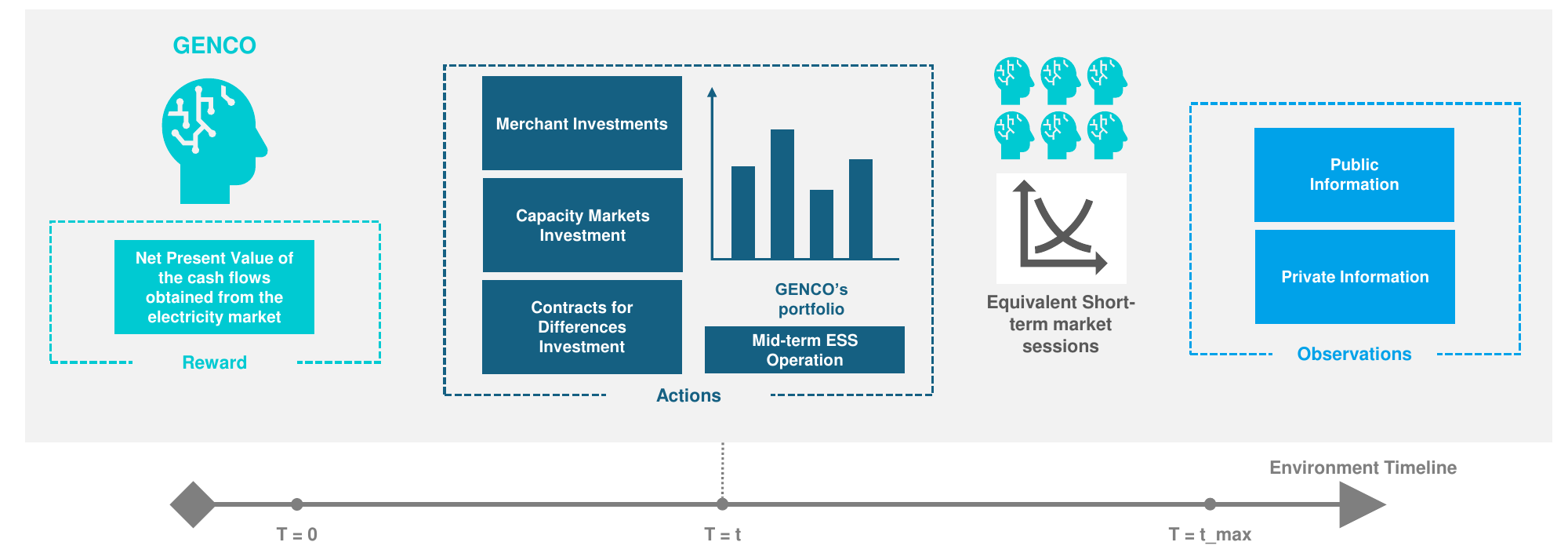} %
    \caption{GENCO structure in the reinforcement learning model, highlighting the interactions between agents, including observed states, selected actions, and received rewards within the environment.}
    \label{fig: GENCO}
\end{figure}

In the model, Generation Companies (GENCOs) are the market agents trained using Reinforcement Learning. Specifically, GENCOs invest and operate generation and storage assets while responding to their individual drivers, system conditions, policy signals, and competition from other actors. Figure \ref{fig: GENCO} schematically presents the GENCO structure implemented in the model. 

GENCOs are assumed to be profit-driven private enterprises, thus aiming to maximize the Net Present Value of the cash flows obtained from the electricity market. This assumption is aligned with Reinforcement Learning principles (\textit{see \ref{Appendix - Agent Reward}}) and widely used in the literature \cite{bublitz_survey_2019}. The study of GENCOs with other strategic objectives (\textit{increasing market share, eliminating competitors, maximizing social welfare, among others}), although not inherently limited by the modeling framework, is left out of the scope of this work.

Individual GENCOs are characterized by their initial portfolio \textit{(enabling the representation of both incumbent and entrant actors)}, decommissioning curves for those existing assets, and the set of available technologies for new investments \textit{(allowing for agents to be exclusively dedicated to specific market categories)}. The current modeling framework can introduce further heterogeneity in GENCOs using differentiated discount rates, technology costs, fuel prices, decommissioning curves, and timings for entering or exiting the market.

To participate in the electricity market, GENCOs make investment decisions in generation plants \textit{(\ref{Appendix - Generation technologies})} and ESS \textit{(\ref{Appendix - Energy Storage Systems})} through the available, and mutually exclusive, investment channels \textit{(Merchant Investments - \ref{Appendix - Merchant Investments}, Capacity Market - \ref{Appendix - Capacity Market}, and Contract for Difference Market - \ref{Appendix - Contract for Difference Market})}. Investment decisions are lumped, with discretization steps equal to all market players and independent across technologies. Once assets finish construction and enter operation, they take part in equivalent short-term market sessions (\textit{\ref{Appendix - Equivalent short-term market sessions}});  generation plants present bids according to their availability and marginal production costs, while ESSs follow the dispatching rules detailed in \ref{Appendix - Energy Storage Systems}. Considering the results from the short-term markets, GENCOs sell their plants' production to the system, while complying with the commitments derived from the previously mentioned investment mechanisms, such as the financial settlements from Contracts for Differences.

To represent investments and operational assets, a stylized approach models their cash flows. The framework assumes unlimited access to financing and no equity constraints, thus enabling GENCOs to invest up to the maximum allowed quantities per technology in each time step. Moreover, agents incurred the total amount of CAPEX for any particular investment during construction. Finally, only before-tax cash flows are considered, but corporate tax rates could be included if needed for the question under study. GENCOs make investment decisions across markets and technologies to build their portfolios within this structure. It is worth noting that, unlike the single-asset evaluations implemented in \cite{anwar_modeling_2022, marc_melliger_phasing_2022} and similar models, the GENCOs' investment decisions do not depend on internal profit projections per technology and/or asset. Instead, investments rely on the portfolio dynamics learned during the training process.

Lastly, individual GENCOs rely on two distinct sets of system information for decision-making \textit{(and training)}, as detailed in \ref{Appendix - Agent Observations}. On the one hand, agents have access to public information, shared across market players, describing the general state of the electricity system (\textit{expected demand growth, availability of generation assets, market prices, total installed capacity per technology, among others}). On the other hand, agents also use their private information (\textit{market participation and financial performance per generation assets and markets}), which is not shared across market actors. The selection of private and public variables is coherent with real electricity systems. A central unit (\textit{Independent System and Market Operators, for example}) provides key system-wide figures and projections, while GENCOs safeguard their internal data. 

\subsubsection{Equivalent Short-term market}
\label{Appendix - Equivalent short-term market sessions}
The model uses equivalent short-term sessions to represent day-ahead markets, and the interactions emerging from resource availability, price formation, demand profiles, and settlement of financial/commitments in the system. Specifically, equivalent short-term market sessions are defined as sequential \textit{(e.g. the session representing the third and fourth months in a year follows the one for the first and second months)} 24-hour windows, with an hourly resolution, intended to capture daily seasonal patterns in electricity systems. The aggregation process \textit{(transforming X-months into a 24-hour window} is carried out using standard representative period methodologies, as described in \ref{Appendix - Resource, demand, and representative periods}. 

In each hour, GENCOs participate in the market by presenting quantity and price bids for their portfolio. These bids are input for double-sided marginal price auctions, dispatching resources based on their merit order and setting a unique system price. Once Equivalent Short-term market outcomes are obtained, other market mechanisms are settled. Taking from real electricity market designs \cite{cramton_electricity_2017}, the maximum price for the short-term auctions is limited by a market cap intended to represent the Value of Lost Load (VoLL), allowing the system to reflect scarcity conditions. The VoLL is set to $4000 \, \text{\texteuro}/\text{MWh}$ in the test bench. This, acknowledging that real VoLL is not observable. 

Given the copper plate assumption \textit{see \ref{Appendix - Transmission Infraestructure}} and the lack of unit-commitment constraints, the market is solved by ordering demand and supply bids according to their prices and later matching quantities to find the clearing price. This solution method reduces overall wall time by increasing the environment throughput and facilitating parallelization, relevant characteristics for realistic RL implementations given the available hardware. Yet, for model versions that intend to represent more complex system dynamics and inter-temporal constraints within the equivalent short-term markets, framing and solving the market as an optimization problem would be necessary. 

\subsubsection{Generation Technologies}
\label{Appendix - Generation technologies}

In the model, GENCOs can invest in assets from a pool of electricity generation technologies, both fossil-based and Renewable Energy Sources (RES). Generally, any technology is described by its CAPEX, OPEX, Fixed O\&M Costs, Variable O\&M Costs, a given Construction Time, and a Downtime Index. While the first parameters are all standard in energy-system models, further discussion is needed for the latter group. 

In the case of the Construction Time, it is set to represent the duration of the civil and electrical works exclusively, referred to as commissioning time in \cite{gumber_global_2024}. This is equivalent to assuming that the generation asset to be constructed is selected by the agent \textit{(or bought from a secondary market)} from a pipeline of projects under development, which have completed their environmental, social, and legal licensing, in addition to holding connection rights equivalent to their expected installed capacity \textit{(Ready to Build stage)}. In particular, the last assumption should be revisited in versions of the model which include transmission systems and network constraints \textit{(See the discussion in \ref{Appendix - Transmission Infraestructure})}.

The Downtime Index represents the individual plant's availability during the Equivalent short-term market sessions, given prolonged maintenance and forced/unforced failures. For the implementation, the Downtime Index is directly modeled as a discrete random variable with realizations of 0, 50, and 100\% availability, corresponding to not operational, partially operational, and fully operational conditions, respectively. As such, the number of realizations in the discrete random variable allows for a straightforward method for controlling both the expectation and the standard deviation for asset availabilities. 

For Conventional fossil-fuel technologies, time-dependent Efficiency/Heat Rates and Fuel Costs can be included in the Variable O\&M, but are kept constant in this work. Furthermore, Conventional plants are considered available at their rated capacity during the Equivalent Short-term market. Instead, RES technologies' hourly capacity factor is described using time series and representative periods \textit{(See \ref{Appendix - Equivalent short-term market sessions})}, thus allowing the model to reflect their short-term variability. In both cases, the effect derived from the Downtime Index is aggregated on top of the hourly availability of generation assets. 

Given these considerations, Table \ref{Table: Generation Technology characteristics} presents the characteristics of available generation technologies. In particular, the Python for Power System Analysis (PyPSA) database \cite{brown_pypsa-eur_2024} has been used for costs and efficiency parameters in three specific years \textit{(2020-2030-2040)}, with lineal interpolations in between, while construction times are based on \cite{international_energy_agency_average_2019, gumber_global_2024}. For the test bench system, the Downtime Index for all technologies is adjusted to result in an expected availability during representative periods equal to 95\% and a standard deviation of 21\%. Yet, these parameters can be adjusted if information for asset availability is available. 

\begin{table}[hbt!]
\small
\centering
\resizebox{0.9\columnwidth}{!}{
\begin{tabular}{@{}|c|c|c|c|c|c|c|@{}}
\toprule
\textbf{\begin{tabular}[c]{@{}c@{}}Feature  \\ Technology\end{tabular}}             & \textbf{Solar} & \textbf{\begin{tabular}[c]{@{}c@{}}Onshore \\ Wind\end{tabular}} & \textbf{\begin{tabular}[c]{@{}c@{}}Offshore\\ Wind\end{tabular}} & \textbf{Coal} & \textbf{\begin{tabular}[c]{@{}c@{}}Open Cycle \\ Gas Turbine\end{tabular}} & \textbf{\begin{tabular}[c]{@{}c@{}}Combined Cycle\\ Gas Turbine\end{tabular}} \\ \midrule
\begin{tabular}[c]{@{}c@{}}\textbf{CAPEX} \\ {[}\texteuro/MW{]}\\ \textit{2020 - 2040}\end{tabular}             & 562 - 323      & 1183 - 1033                                                      & 2437 - 2009                                                      & 3827 - 3827   & 480 - 443                                                                  & 931 - 860                                                                     \\ \midrule
\begin{tabular}[c]{@{}c@{}}\textbf{OPEX - Fixed}\\ {[}\% CAPEX / MW - year{]}\end{tabular}  & 2.5            & 1.7                                                              & 2.3                                                              & 1.31          & 3.3                                                                        & 1.77                                                                          \\ \midrule
\begin{tabular}[c]{@{}c@{}}\textbf{OPEX - Variable}\\ {[}\texteuro/MWh{]}\end{tabular}              & 1              & 1.5                                                              & 1.5                                                              & 32            & 46                                                                         & 64                                                                            \\ \midrule
\begin{tabular}[c]{@{}c@{}}\textbf{Emission factor} \\ {[}\(\text{tCO}_2\text{eq}\)/MWh{]}\end{tabular}            & 0              & 0                                                                & 0                                                                & 1.01          & 0.34                                                                       & 0.495                                                                         \\ \midrule
\begin{tabular}[c]{@{}c@{}}\textbf{Construction time} \\ {[}\# - years{]}\end{tabular}       & 2              & 3                                                                & 4                                                                & 3             & 3                                                                          & 2                                                                             \\ \midrule
\begin{tabular}[c]{@{}c@{}}\textbf{Maximum Investment} \\ {[}MW{]}\end{tabular}              & -           & -                                                             & -                                                             & -          & -                                                                       & -                                                                          \\ \midrule
\begin{tabular}[c]{@{}c@{}}\textbf{Operational Lifetime}\\ {[}\# - years{]}\end{tabular}     & 35             & 28                                                               & 30                                                               & 40            & 25                                                                         & 25                                                                            \\ \midrule
\begin{tabular}[c]{@{}c@{}}\textbf{Expected availability} \\ {[}\%{]}\end{tabular}           & 92.5           & 92.5                                                             & 92.5                                                             & 92.5          & 92.5                                                                       & 92.5                                                                          \\ \midrule
\begin{tabular}[c]{@{}c@{}}\textbf{Standard dev. availability} \\ {[}\%{]}\end{tabular} & 23             & 23                                                               & 23                                                               & 23            & 23                                                                         & 23                                                                            \\ \bottomrule
\end{tabular}%
}
\caption{Characteristics and relevant parameters for Generation technologies. Based on \cite{international_energy_agency_average_2019, gumber_global_2024}}
\label{Table: Generation Technology characteristics}
\end{table}

To integrate additional technologies into the model, Table \ref{Table: Generation Technology characteristics} should be extended with the corresponding parameters \textit{(which are available directly from \cite{brown_pypsa-eur_2024}, for instance)}. Yet, as described in \textit{\ref{Appendix - Agent Actions}} any newly available technology comes with substantial increments in the agents' action space and subsequent increases in modeling complexity. Thus, the current model concentrates on key and representative investment options for the Energy Transition in the short and mid-term. Moreover, depending on the specific policy question under study, the available technologies could be easily switched between cases/scenarios. 

\subsubsection{Energy Storage Systems}
\label{Appendix - Energy Storage Systems}

Currently, the model includes two types of Energy Storage Systems (ESS). On the first hand, short-term battery energy storage systems (BESS), which are characterized by a time-duration \textit{(Energy to Power ratio)} between 3-4 hours. On the other hand, pumped-storage hydropower plants are described by their corresponding hydrologic inflows and time durations ranging from hours to weeks. Long-term Energy Storage Systems, those with time durations longer than the periods represented by the short-term equivalent market sessions, are left out of the scope of this work. 

Similarly to generation technologies, GENCOs have an existing ESS capacity at the beginning of the simulation. However, given possible constraints for deployments of new Pumped-Storage facilities, the current model only enables BESS for investments. Also analogous to generation technologies, the model allows for controlling which and when agents can invest in BESS, although the time component is not explored in the current work. Given these considerations, Table \ref{Table: Energy Storage Technology characteristics} summarizes the ESS features and cost assumptions, based on \cite{brown_pypsa-eur_2024,terna_study_2023}. 

\begin{table}[hbt!]
\small
\centering
\resizebox{0.5\columnwidth}{!}{
\begin{tabular}{@{}|c|c|c|@{}}
\toprule
\textbf{\begin{tabular}[c]{@{}c@{}}Feature  \\ Technology\end{tabular}}                    & \textbf{\begin{tabular}[c]{@{}c@{}}ESS \\ short-term\end{tabular}} & \textbf{\begin{tabular}[c]{@{}c@{}}ESS\\ mid-term\end{tabular}} \\ \midrule
\begin{tabular}[c]{@{}c@{}}CAPEX \\ {[}MW{]}\\ 2020 - 2040\end{tabular}                    & 562 - 323                                                          & N/A                                                             \\ \midrule
\begin{tabular}[c]{@{}c@{}}OPEX - Fixed \\ {[}Eur/ MW - year{]}\\ 2020 - 2040\end{tabular} & 32,000 - 117,00                                                    & 32,000-11,700                                                   \\ \midrule
\begin{tabular}[c]{@{}c@{}}Time Duration\\ {[}\# - hours{]}\end{tabular}                   & 4                                                                  & 34                                                              \\ \midrule
\begin{tabular}[c]{@{}c@{}}Emission factor \\ {[}CO2/MWh{]}\end{tabular}                   & 0                                                                  & 0                                                               \\ \midrule
\begin{tabular}[c]{@{}c@{}}Construction time \\ {[}\# - years{]}\end{tabular}              & 2                                                                  & N/A                                                             \\ \midrule
\begin{tabular}[c]{@{}c@{}}Maximum Investment \\ {[}MW{]}\end{tabular}                     & -                                                               & N/A                                                             \\ \midrule
\begin{tabular}[c]{@{}c@{}}Operational Lifetime\\ {[}\# - years{]}\end{tabular}            & 25                                                                 & N/A                                                             \\ \midrule
\begin{tabular}[c]{@{}c@{}}Expected availability \\ {[}\%{]}\end{tabular}                  & 92.5                                                               & 92.5                                                            \\ \midrule
\begin{tabular}[c]{@{}c@{}}Standard deviation availability \\ {[}\%{]}\end{tabular}        & 23                                                                 & 23                                                              \\ \bottomrule
\end{tabular}%
}
\caption{Characteristics and relevant parameters for Energy Storage System Technologies. Based on \cite{brown_pypsa-eur_2024,terna_study_2023}}.
\label{Table: Energy Storage Technology characteristics}
\end{table}

\begin{figure}[h!]
    \centering
    \includegraphics[width=0.9\linewidth]{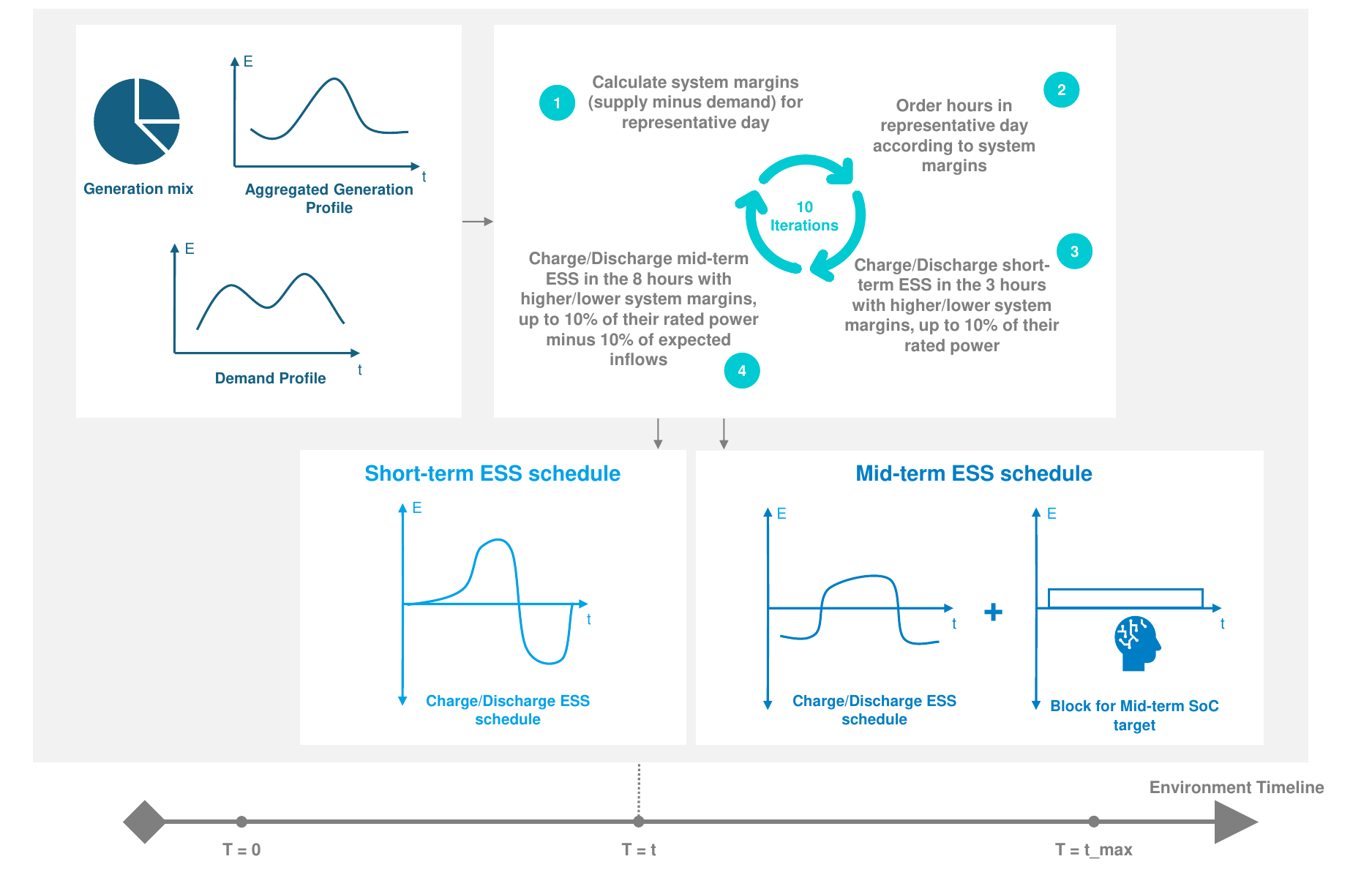} %
    \caption{Scheduling for short and mid-term ESS. First, demand and aggregated generation profile are obtained \textit{(upper-left panel)}. Second, the scheduling is calculated via an iterative process that charges/discharges ESS in system conditions with maximum/minimum system margins \textit{(upper-right panel)}. This procedure directly results in the short-term ESS schedule \textit{(lower-left panel)}. Instead, to obtain the mid-term ESS schedule, the required quantities to achieve the desired SoC level for the next environment step are integrated into the profile \textit{(lower-right panel)}}. 
    \label{fig: ESS scheduling}
\end{figure}

In the model, all ESSs are operated during the equivalent short-term market sessions abiding by a predefined schedule, which is calculated based on margins between available generation resources and demand. This calculation, which is part of the procedure used in the Capacity Market to define capacity credits and system requirements, is illustrated in Figure \ref{fig: ESS scheduling}. For all ESS, the scheduling is intended to maximize ESS flexibility in daily system operation by ensuring ESS charge/recharge in conditions with maximum/minimum resource availability. Yet, following guidelines for ESS modeling \cite{mantegna_establishing_2024}, slight variations in the calculation of the predefined schedules are necessary to account for the difference in time durations across technologies:

\begin{itemize}[noitemsep, left=8pt]
\item In the case of short-term ESS, the schedule allows for charging/discharging up to the ESS-rated power capacity. Yet, charge/discharge events can be distributed across the equivalent period depending on system margins. Considering this, to account for the ESS energy capacity constraint the schedule ensures a complete operational cycle is carried out \textit{(charge and discharge energy are equal, minus efficiency differences)}. Finally, short-term ESSs begin all equivalent periods at a state of charge that enables such operation, a reasonable assumption considering their relatively low time duration. 
\item For mid-term ESS, the scheduling process adjusts the previous calculations to account for the plant's expected hydrologic inflows and their longer time duration:
\begin{itemize}[noitemsep, left=16pt]
\item First, GENCOs set the desired state of charge for their ESS in the next equivalent period \textit{(from 0 to 100\% state of charge to allow for full flexibility)}. 
\item Once this objective is defined, the charging/discharging cycles are modified to achieve it while considering both the energy balance and the power capacity in the pumped-storage facility. 
\item Given the relationship between the time duration of mid-term ESS and the duration of equivalent periods, the model assumes such inter-period operation is always feasible. This assumption is no longer valid for ESS with time durations longer than the representative periods, as the objective set by the Agent in the current method might be unfeasible. 
\end{itemize}
\end{itemize}

It is worth mentioning that this implementation now abides by all the high-level guidelines defined in \cite{mantegna_establishing_2024}, and some technical constraints may be not respected if the same policies are applied to a setup without representative periods. Moreover, to comply with the predefined schedule, ESSs are operated as price-takers in the equivalent short-term market \textit{(thus submitting maximum/minimum price bids when buying/selling energy)}. As such, the active role of ESS in short-term market sessions is not included in the model, an issue that will gain more relevance with increasing RES penetration. However, the selected implementation can capture ESS flexibility in the time frame relevant to the technologies' characteristics, while maintaining the problem tractable for the current RL implementation. 

Further improvements to the ESS module can be envisioned for research-specific model versions. To start, investments on ESS with different characteristics \textit{(different time-durations)} could be enabled, thus allowing agents to select assets depending on implicit/explicit flexibility requirements and incentives. Moreover, without significant modifications in the current framework, the model could allow agents to choose, in each equivalent market session, pre-calculated operation profiles for their ESS that respect their internal constraints \textit{(e.g. aggressively charge during peak-solar hours, or charge evenly distributed across the day)}. Finally, for studies with a short-term focus, allowing agents to select an hourly state of charge across the equivalent short-term sessions is feasible. 

To achieve this last objective, two alternatives can be implemented. In the first, equivalent market sessions are abandoned, allowing agents to make decisions hourly. In this setup, GENCOs would directly control their ESS's state of charge, while the technical feasibility of decisions could be enforced via action-masking. Yet, the consequent increase in the planning horizon when modeling investment dynamics is dramatic \textit{(moving from few representative periods per year to thousands)}, raising feasibility questions considering available hardware and algorithms. In the second alternative, representative periods are maintained. Still, agents are allowed to select hourly charge/discharge profiles using either safe-RL techniques, on the one hand, or traditional RL algorithms but applying penalties when not feasible states are reached, on the other. However, it should be noted that the former option increases algorithm complexity \cite{liu_constrained_2022, ray_benchmarking_nodate}, and the latter significantly amplifies the training challenges compared to action masking \cite{huang_closer_2022}. 

\subsubsection{Transmission Infrastructure \textit{(or lack thereof)}}
\label{Appendix - Transmission Infraestructure}

The current version of the model is based on a copper plate assumption for the electricity system, equivalent to a condition where generation and transmission expansion are ideally coordinated. Consequently, when investing in new assets, GENCOs can connect their plants as soon as they finish commissioning. They are guaranteed not to face curtailments caused by network constraints during the project's lifetime. For certain studies, the copper plate assumption might be the biggest shortcoming of the current model, as transmission congestion and RES curtailments have been highlighted as one of the key barriers for massive RES deployment. More so, electricity market design can be a tool used by policymakers to mitigate the impact of such physical constraints \cite{savelli_putting_2022, newbery_efficient_2023, pollitt_locational_2023}. 

Yet, the choice of using a copper plate assumption was made considering trade-offs between the main methodological objectives in this work, the policy questions to be addressed with this type of model, the capabilities of existing MARL methods and implementations, and the available hardware. To start, the model is built towards mid- and long-term assessments in which particular aspects of the market design are tested to identify if they represent a constraint for building electricity systems compliant with stringent decarbonization targets. If constraints emerge from this analysis, it is reasonable to infer that the challenges would intensify when transmission systems are explicitly represented. From an implementation standpoint, the copper plate assumption enables all auctions within market sessions to be solved algorithmically, thus reducing both the training loop runtime and the complexity of parallelization, two critical considerations for RL. Last, to make the training setup and its results graspable, the current state of MARL applied to competitive environments demands a simplified configuration to ensure comprehensibility. In particular, including a network representation in a context where GENCOs are changing dramatically the composition of the Generation Mix during training \textit{(in terms of technologies, installed capacities, timings, and market outcomes)} could add excessive complexity to the framework. 

Nonetheless, future model versions could integrate different representations of transmission system infrastructure. Initially, a DC network representation \textit{(including both dispatch and expansion mechanisms)} could be part of the environment, and thus run inside the RL training loop. Such an addition would reduce the RL algorithm throughput and increase parallelisation challenges, negatively impacting learning efficiency and even affecting the choice of training architecture. Still, it could remain feasible for the type of implementation presented in this work. Alternatively, the transmission system infrastructure could be represented via a soft coupling with a detailed network model, which would run outside the RL training loop and periodically update the market environment. Finally, in the case of large-scale studies with ample computational resources, such as \cite{openai_dota_2019, vinyals_grandmaster_2019}, a detailed network model could be integrated as part of the training environment and run in the training loop itself. 

\subsubsection{Resource availability, Electricity demand, and representative periods}
\label{Appendix - Resource, demand, and representative periods}
For the model, time series projections from \cite{antonini_weather-_2024,antonini_weather-_2024-1} describe RES hourly capacity factor. This database provides information for onshore/offshore wind, solar, and hydrologic inflows for a year \textit{(between 2020 and 2100)}, EU country, and Representative Concentration Pathways (RCP). Information for the RCP 2.6 and the Italian system is used for the test bench. 

In the case of electricity demand, projections for the 2030 Italian power system are used to describe the yearly profile \cite{di_bella_mitigation_2025}. Notably, the demand time series includes the effects of Climate Change adaptation measures, such as increased air-conditioning adoption. From this time series, the model obtains electricity demand in a particular year via an exogenous demand growth rate \textit{(treated as a scenario parameter)}. Yet, if additional hourly demand profiles are available, other methods to include electricity demand could be incorporated \textit{(e.g. interpolating historical and projected profiles for different years)}. It is worth noting electricity demand is assumed to be inelastic to both short and long-term electricity prices. Although demand bids that reflect elastic and inelastic blocks could be incorporated into the equivalent market sessions as a proxy for short-term price elasticity, this feature is not currently used in the model. Moreover, no explicit mechanism is contemplated to include long-term price elasticity.  

Once the time series for RES availability and electricity demand are obtained, they are aggregated to form representative periods, a methodology commonly used to reduce the size in energy-system models \cite{poncelet_selecting_2017}. In particular, aggregation is carried out using the TSAM Python library, developed and discussed in \cite{hoffmann_typical_2021}, to transform hourly a group of time series from any given length \textit{(monthly, bi-monthly, six-monthly, etc.)} to create a set amount of 24-hour representative days. For the test bench, bi-monthly time series are jointly transformed into 4 representative days, while an extra representative day is calculated using maximum demand conditions to be used in the Capacity Market \textit{(see \ref{Appendix - Capacity Market})}.

During simulations \textit{(training and execution)}, one representative day is randomly selected for each time step and applied to all generation resources and demand. Consequently, GENCOs face \textit{mid-term variability}, as the electricity system is affected by both changes in resource and demand profiles, besides the individual asset availability described in \ref{Appendix - Generation technologies}. In contrast, other sources of system variability and/or uncertainty for the short-term \textit{(e.g. variations in solar/wind availability within the representative periods)} and long-term \textit{(the realization of one or another RCP pathway or uncertainty regarding technological changes)} are excluded from the current model. 

\subsubsection{Merchant Investments}
\label{Appendix - Merchant Investments}

The first mechanism for GENCOs to invest and access the electricity system is through Merchant Investments. Yearly, any GENCO can decide to place projects from different technologies under construction, with capacities limited by the maximum investment enabled per agent shown in \ref{Table: Generation Technology characteristics}. No centralized process is implemented to constrain Merchant Investments in the system, resulting in much higher investments than are financially reasonable during the early stages of training.  

Once a merchant investment enters operation, the project earns revenues directly from the short-term market. In practice, during excess supply situations, the plant has no protections against sustained low prices, which might prove insufficient to justify the investment. However, when scarcity conditions occur, the plant will harness the fully-fledged scarcity rents. As mentioned earlier, investment mechanisms are mutually exclusive, and thus, Merchant Investments cannot participate in the Capacity or CfD Markets. 

Currently, the model allows activating/deactivating merchant investments or any other two investment mechanisms. This tool, which can be applied to any subset of agents or periods in the simulation, enables testing different market designs. For instance, an Energy-Only Market can be implemented by deactivating the CfD and the Capacity Markets. 

Finally, from a financial perspective, the model considers existing generators as passive assets, receiving income exclusively from day-ahead markets. As such, these plants do not participate in the expansion auctions available to new investments or the Capacity Market. A more active integration of existing generators, as needed when modeling forward markets, is left out of the scope of this work.  

\subsubsection{Investments in the Contract for Difference Market}
\label{Appendix - Contract for Difference Market}

The second investment channel for GENCOs is a dedicated RES market built around a two-way Contract for Differences. In this market, a regulator \textit{(or central planner)} establishes a penetration target for RES in the system, defined as the share between renewable production and total demand. When GENCOs have built insufficient RES projects to achieve the desired penetration target, an auction takes place to cover the missing renewable share. Winning projects in the auction are awarded a two-way Contract for Difference. The design of the two-way CfDs is inspired by the proposals to modify the original one-way CfDs after the Ukraine War \cite{european_parliament_improving_2024}. Still, CfD designs that focus on locational and comparative yardstick aspects, as discussed in \cite{schlecht_financial_2023,newbery_efficient_2023}, are left out of the scope of this work. 

\begin{figure}[h!]
    \centering
    \includegraphics[width=0.9\linewidth]{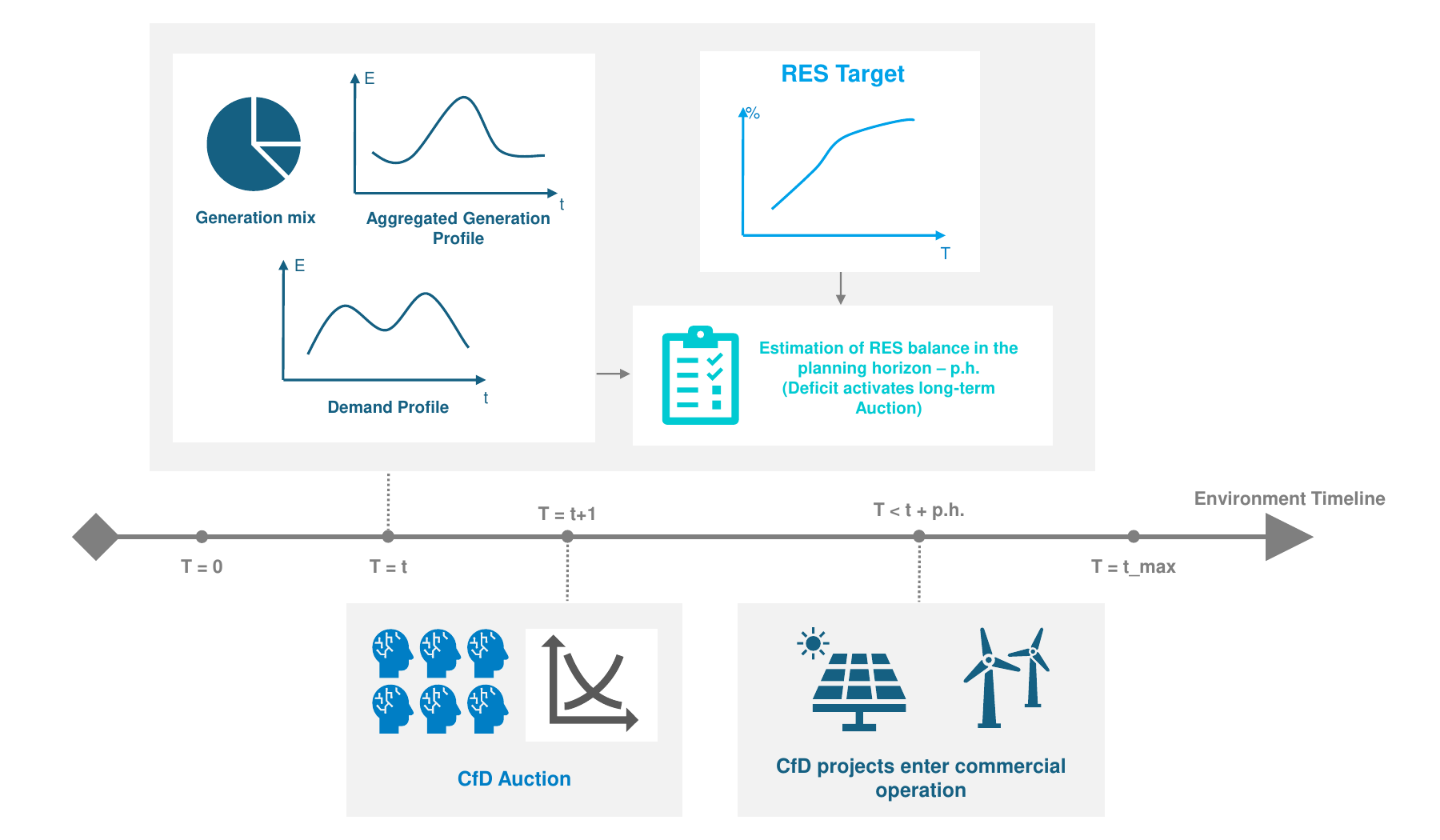} %
    \caption{Operations in the Contract for Difference Market. First, demand and aggregated generation profile are obtained \textit{(upper-left panel)}. Then, given the Renewable Energy target defined by the regulator \textit{(upper right panel)} the balance of available resources concerning the target is calculated \textit{(center panel)}. When a RES deficit is found, an auction occurs in the following period \textit{(lower left panel)}. Selected projects from the RES auction enter operation, depending on the specific time of their technologies, but before the regulator's planning horizon \textit{(lower right panel)}}.
\label{fig:CfD Market}
\end{figure}

From a GENCO perspective, the two-way CfD ensures a fixed price for the total output of the participant plant. Equivalently, the GENCO hedges its project against low-price conditions in the system. Moreover, the GENCO is enforced to build the project presented in the auction and withhold the financial obligation associated with the CfD during the project's lifetime. The following describes the most relevant characteristics in the Contract for Difference Market, shown schematically in Figure \ref{fig:CfD Market}.

\begin{itemize}[noitemsep, left=8pt]
\item The exogenous RES penetration target for the system is set as a curve, extended for all the simulation duration, pairing years and percentages of total system demand that should be covered with renewable energy \textit{(setting all values in the curve equal to zero would avoid any CfD auctions)}. This target, which would be directly linked to mitigation efforts, country ambitions, and carbon pricing schemes, is treated in the model as an exogenous parameter. 
\item To accomplish the predefined target on time, and considering the construction time of the generation assets, the regulator verifies the conditions to carry out RES auctions years in advance. The planning horizon parameter controls this behavior.  
\item To evaluate the RES penetration target, the regulator compares system demand, including its expected growth, with the average output of the RES installed capacity. In this calculation, the regulator subtracts any project undergoing decommissioning and includes plants under construction \textit{(from all investment channels)}. Importantly, investments under all other mechanisms in the system may be sufficient to accomplish the RES target without calling additional CfD auctions. However, an auction will occur in the subsequent year when RES scarcity conditions are detected. 
\item The auction follows a single-end design with pricing that allows both Marginal or Pay-as-Bid closing rules. The regulator sets the ceiling price \textit{(treated by the model as an exogenous parameter)}, while quantities to be auctioned correspond to the value needed to accomplish the RES target.
\item For the auction, GENCOs present bids, pairs of quantities and prices, for each RES technology enabled in their portfolio: 
\begin{itemize}[noitemsep, left=16pt]
\item On the one hand, quantities in bids are the average expected production of the corresponding plants, which is a function of three factors. First is the average availability of the specific technology during the simulation horizon and, more specifically, the time series used to model their availability. Second is the capacity of the individual assets, modeled through lumped steps between zero and the maximum allowed value per technology, as outlined in Table \ref{Table: Generation Technology characteristics}). Third is the average Downtime Index. In practice, GENCOs in the model bid quantities by selecting the asset's capacity, from which the average expected production is automatically calculated. 
\item On the other, prices in bids are values between zero and the ceiling price of the auction, freely chosen by GENCOs without further interaction or constraints. In its current version, the ceiling price of the auction is disclosed to agents \textit{(and used as the maximum price for bidding)}. This feature could be modified to accommodate simulations where the ceiling price is unknown, and GENCO bid based on an alternative maximum value \textit{(see \ref{Appendix - Agent Actions})}.  
\item Bidding for enabled technologies in the portfolio is mandatory. However, GENCOs have two strategies for avoiding undesired allocations in the auction: bidding zero quantities \textit{(rapidly found during training)} or high prices.  
\end{itemize}
\item Winning projects in the auctions enter the agent's construction pipeline and are awarded two-way CfDs. CfDs provide a stable, fixed price for a plant's energy output, mitigating the risks associated with short-term market volatility. 
\item To be dispatched, plants emerging from the CfD auctions participate in the short-term markets without any priority or special treatment. Combined with the absence of active bidding in the short-term market sessions, this means that during generator curtailment caused by excess supply \textit{(e.g. duck-curve)}, GENCOs may lose their CfD income
\end{itemize}

\subsubsection{Investments in the Capacity Market}
\label{Appendix - Capacity Market}

The third investment channel in the model is the Capacity Market. Like the CfD market, the regulator seeks to promote investments in the system through dedicated expansion auctions. However, the objective of the Capacity Market is to ensure resource adequacy, thus guaranteeing available resources are sufficient to cover demand requirements in \textit{close to} all system conditions. The Capacity Market implementation takes inspiration from the Reliability Option design \cite{cramton_colombia_2007, cramton_capacity_2013, mastropietro_reliability_2024}, while focusing on the key factors applicable to new investments. 

From a GENCO's perspective, participating in the Capacity Market provides a premium tied to the project's contribution to system adequacy \textit{(the equivalent capacity credits explained below)}. In return, a GENCO must construct the generation project and shield demand from high-price events through the Reliability Option. Like a financial call option, the Reliability Option requires the GENCO to supply electricity at a capped price during scarcity periods. This capped price, the Scarcity Price, is an integral component of the Capacity Market design and is treated as an exogenous parameter within the model. Importantly, the Reliability Option is enforced even when the asset is not in operation in the short-term market or when the plant's output is lower than the committed contribution to system adequacy. The following describes the Capacity Market in detail, also illustrated in Figure \ref{fig: Capacity Market - Calculation of capacity credits, storage operation, and adequacy margins}.

\begin{itemize}[noitemsep, left=8pt]
\item In real-world Capacity Market implementations, regulators aim to allocate sufficient resources to cover the X\% percentile of worst system conditions, which can be translated to supplying Z\% of hours in the year. Yet, considering the use of representative periods in this model \textit{(one representative day models two months of operation)} a direct hour-by-hour calculation is impossible. As a stylized alternative, the criteria employed in the model to ensure system adequacy is for available supply to be sufficient to meet system requirements during these representative days, assuming maximum demand conditions. 
\item To consider maximum demand conditions, a set of representative days \textit{(one per time step in the model)} is obtained, following the procedure explained in \ref{Appendix - Resource, demand, and representative periods}, but enforcing the algorithm to select the combination of time series that are concurrent with the maximum demand profile. 
\item Once these representative days are obtained, the regulator verifies the balance between supply and demand in the system over the planning horizon \textit{(controlled by the same parameter as in the CfD market)}: 
\begin{itemize}[noitemsep, left=16pt]
\item On the demand side, the profile obtained from the representative days is adjusted according to the expected demand growth for the planning horizon. 
\item On the supply side, the balance calculation considers generation and storage units expected to be in operation by the end of the planning horizon \textit{(existing assets, plus construction pipeline, minus plants undergoing decommissioning)}. 
\item The Generation plants' expected output, in line with the process established for the CfD market, depends on its installed capacity, the Downtime Index, and the time series associated with its hourly production.
\item For ESS instead, the procedure to dispatch them in the equivalent market sessions, as explained in \ref{Appendix - Energy Storage Systems} is applied. As a result, the production/consumption profiles for the ESSs in the system are obtained. 
\item Given all resources in the system are accounted for, the regulator verifies system margins for the representative days. The hour with the lowest system margin is defined as the critical period in the system and serves to identify the scarcity condition for the Capacity Market. 
\item If the margin for the critical period is negative, an expansion auction is scheduled for the subsequent year. 
\item Moreover, the expected availability of generation and storage assets during the critical period is considered their marginal contribution towards system adequacy or their capacity credits for the capacity market. Capacity credits per technology are normalized to be in percentage terms \textit{(a capacity credit of 50\% entails the plant had 50\% of its capacity available during the system critical condition)}. 
\end{itemize}
\item Similar to the CfD market, the auction follows a single-end design with pricing that allows both Marginal or Pay-as-Bid closing rules. The regulator sets the ceiling price \textit{(treated by the model as an exogenous parameter)}, while quantities to be auctioned correspond to the margin between supply and demand in the system's critical period. 
\item For the auction, GENCOs present bids, pairs of quantities and prices, for technologies in their portfolios with capacity credits higher than zero \textit{(those available during the system's critical condition)}: 
\begin{itemize}[noitemsep, left=16pt]
\item On the one hand, quantities in bids correspond to the equivalent capacity credits contributed to the system by a given plant. This value depends on the technology \textit{(and thus the capacity credits per technology calculated in previous steps)}, the capacity \textit{(lumped steps between zero and the maximum allowed value per technology)}, and the Downtime Index. In practice, GENCOs in the model bid quantities by selecting the asset's capacity, from which the project's contribution to system adequacy is automatically calculated. 
\item On the other, prices in bids are values between zero and the ceiling price of the auction, freely chosen by GENCOs without further interaction or constraints. In its current version, the ceiling price of the auction is disclosed to agents \textit{(and used as the maximum price for bidding)}. This feature could be modified to accommodate simulations where the ceiling price is unknown, and GENCO bid based on an alternative maximum value \textit{(see \ref{Appendix - Agent Actions})}.  
\item Bidding for enabled technologies in the portfolio is mandatory. However, GENCOs have two strategies for avoiding undesired allocations in the auction: bidding zero quantities \textit{(rapidly found during training)} or high prices.
\end{itemize}
\item Winning projects in the auctions enter the agent's construction pipeline and are awarded Capacity Obligations. As such, GENCOs receive a premium \textit{(the resulting price from the auction)} for their equivalent capacity credit allocation. In exchange, prices for energy traded in the short-term market are capped \textit{(at the Scarcity Price of the market)}. In cases where the plant's output is lower than the Firm Capacity committed in the auction, GENCOs are also responsible for the excess costs incurred by demand. 
\end{itemize}

\begin{figure}[h!]
    \centering
    \includegraphics[width=0.9\linewidth]{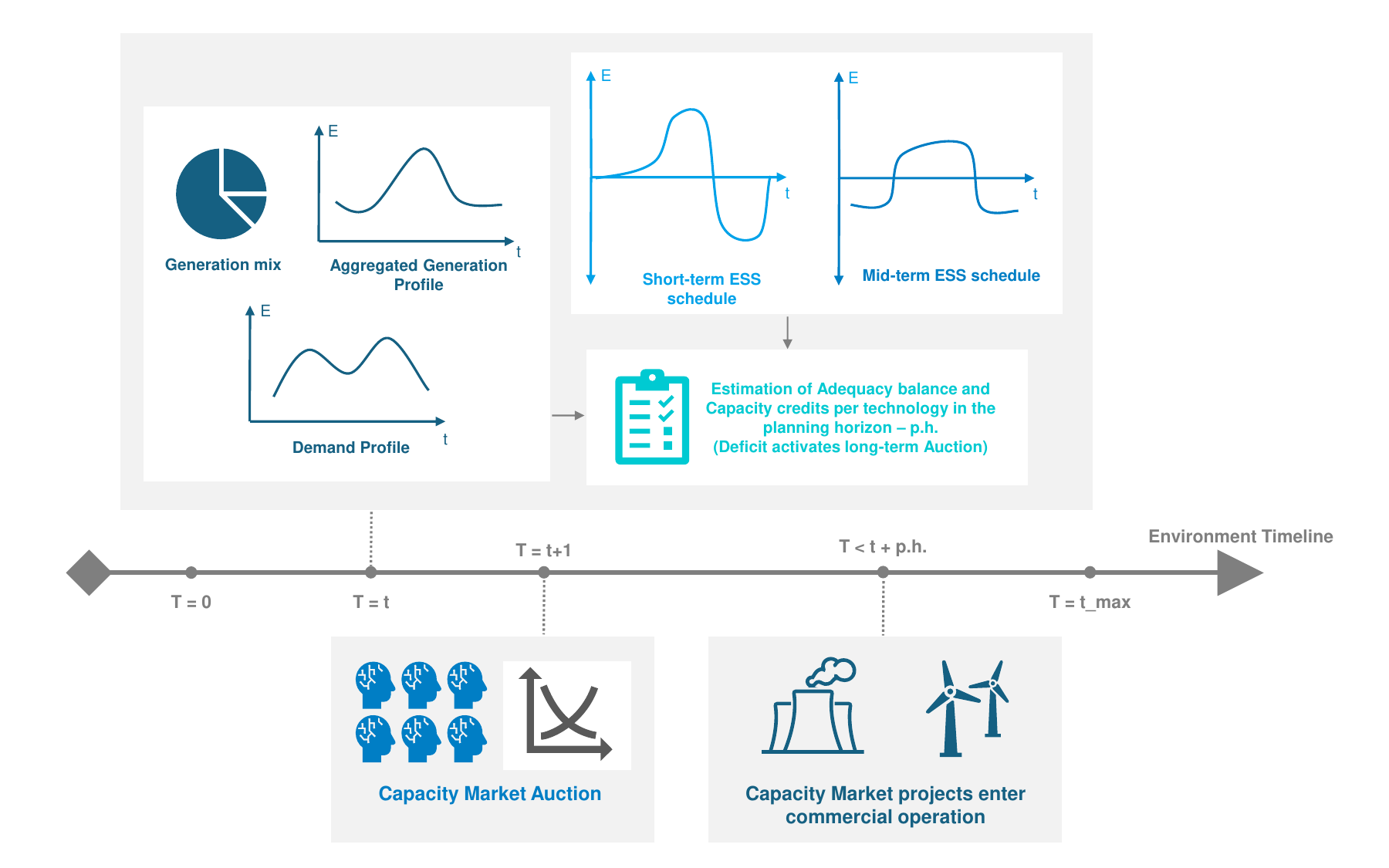}  %
    \caption{Operations in the Capacity Market. To start, demand, aggregated generation profile, and ESS schedules are obtained \textit{(upper-left panel)}. With this information, the regulator estimates the system's adequacy balance and the Capacity credits per technology \textit{(center panel)}. An auction occurs in the following period \textit{(lower left panel)} when an adequacy deficit is found. Selected projects from the Capacity Market auction enter operation, depending on the specific time of their technologies, but before the regulator's planning horizon \textit{(lower right panel)}}.
\label{fig: Capacity Market - Calculation of capacity credits, storage operation, and adequacy margins}
\end{figure}

\subsubsection{Additional Policy instruments}

Apart from market design, the model includes two additional policy instruments that could help to shape and define scenarios. The first is a Carbon Tax, directly linked to the \( CO_2\) produced by the different generation technologies, as indicated in \ref{Table: Generation Technology characteristics}. In the presence of a Carbon Tax, GENCOs include in their short-term market bids the emissions' equivalent cost, effectively acting as a pass-through to final consumers. In the model, the carbon tax is provided in \texteuro\(/\text{tCO}_2\text{eq}\). The second instrument is an exogenous limit for investing in specific technologies. This limit can be enforced at any point during the simulation and applied discriminately to market players and investment channels \textit{(e.g. Coal investments through the Capacity Markets are forbidden from the year 2030 and beyond)}. 

\subsection{Electricity Market as a Multi-Agent Reinforcement Learning Environment}
\label{Appendix - MARL Environment}

In this section, the integration of the Long-term Electricity Market into the Gymnasium standard \cite{towers_gymnasium_2023}, and more specifically, the version of Multi-Agent environments developed by the RLLIB team \cite{liang_rllib_2018}, is presented. The general structure in the environment is described first, while the main building blocks \textit{(Rewards, Actions, Observations, and Initial/Terminal State)} are detailed next. 

\subsubsection{Environment General Structure}
\label{Appendix - Environment Steps}

In the Gymnasium standard, environments for Reinforcement Learning use steps to simulate the transitions in the underlying Markov Decision Process. In each step, agents observe the system's state, take action, and receive the corresponding reward. Once this process has been completed, the environment transitions to a new state, and the stepping process is repeated.  

In the context of the market model, these environment steps are directly linked to Equivalent Short-Term Market sessions, as detailed in \ref{Appendix - Equivalent short-term market sessions}. Each step is designed to represent a fixed period of operation by condensing interactions into a single 24-hour representation. This is achieved using representative periods, as described in \ref{Appendix - Resource, demand, and representative periods}. Under this framework, the total number of steps defines the simulation length. For instance, the test bench uses 120 steps, each reducing two calendar months into 24-hour periods, to simulate 20 years operation in the electricity market. 

During each environment step, the GENCOs' portfolios participate in the Equivalent Short-Term Market, generating market outcomes that serve as the basis for the environment's observations and rewards. However, aside from operating mid-term ESS, agents do not actively make decisions affecting the short-term market, as their participation in it is automated. Instead, investment decisions are enabled every year, allowing agents to take actions to expand their portfolios. Investment in the different markets occurs \textit{(if enabled)} every two environment steps and in sequence across the year \textit{(e.g. investments in the CfD market follow investment decisions in the capacity market)}. The occurrence of investments in particular periods and markets is controlled through action masking, as explained in \ref{Appendix - Agent Actions}. 

\begin{figure}[h!]
    \centering
    \includegraphics[width=0.9\linewidth]{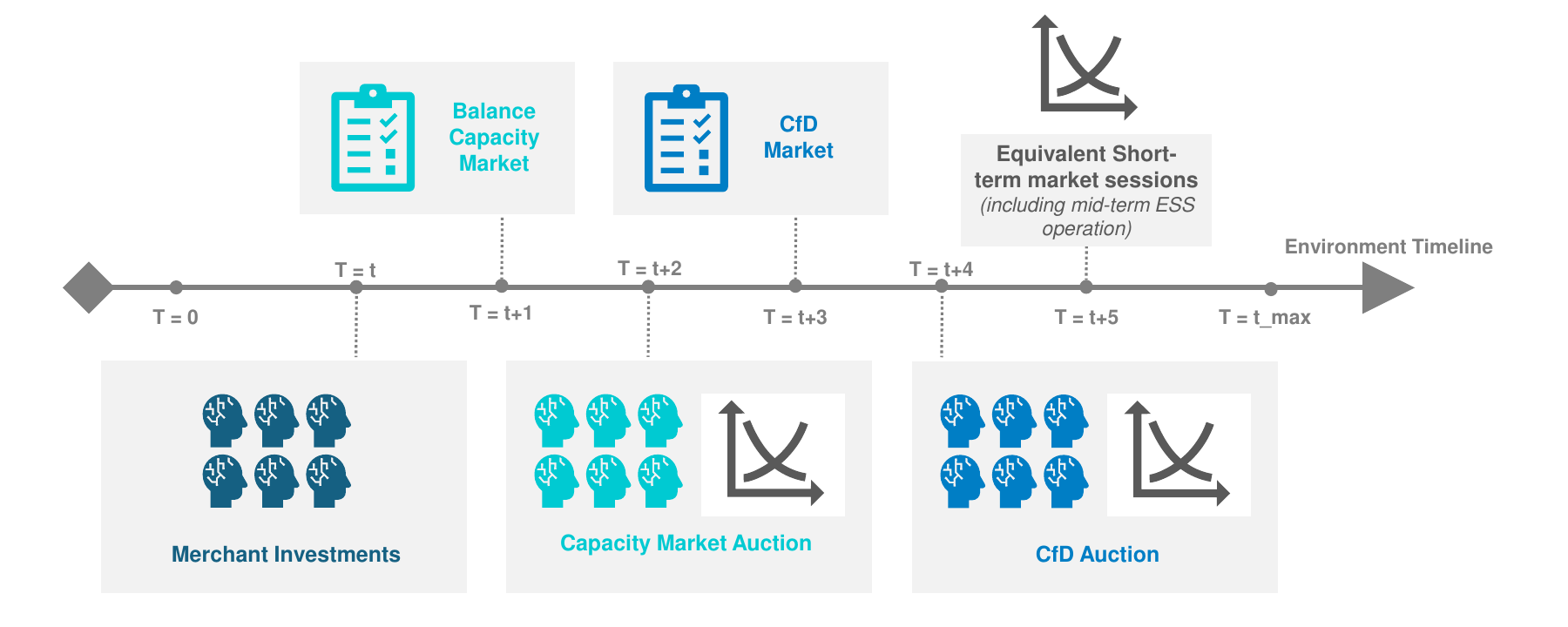} %
    \caption{Structure of the long-term electricity market, illustrating one year of operation. Agents operate in the equivalent short-term market \textit{(taking place in every environment step)}, while investment decisions are taken yearly. Investment decisions occur in sequence \textit{(merchant - Capacity Market - CfD market)}, and the case of the Capacity and CfD market depending on the balance calculations for the system.}
\label{fig: One-year operation in the environment}
\end{figure}

This setup provides a structured approach where environment steps are associated with two-month intervals for detailed observations and rewards. At the same time, investment decisions are aligned with the longer timeframes typical of portfolio adjustments. By combining granular market data with appropriately spaced decision-making, the model balances operational detail and investment timing. Figure \ref{fig: One-year operation in the environment} summarizes the environment's structure by presenting its evolution for an equivalent calendar year. 

\subsubsection{Reward Function}
\label{Appendix - Agent Reward}

In the model, agents are assumed to maximize the net present value of the cash flows obtained from the electricity market, as explained in \ref{Appendix - GENCOs}. This involves calculating and aggregating revenues and costs from all assets and markets relevant to the agents' portfolio across the simulation length. Yet, to be consistent with real company operations, where there is exposure to short-term variations in cash flows, the reward function is designed to be passed to agents step-by-step. 

Abiding by this principle, Equation (\ref{Appendix - eq:total_reward}) presents the reward harnessed by agents from each environment step. The reward is divided into two parts: profits and investment costs. On the one hand, the profits \(\left(P_{\text{M}},P_{\text{CM}},P_{\text{CfD}}\right) \) derived from all markets in the system, including the revenues earned from energy traded, but also the fixed and variable costs incurred while operating the asset. On the other hand, the investment costs \(\left(IC_{\text{M}},IC_{\text{CM}}, IC_{\text{CfD}}\right)\) of new portfolio additions, similarly disaggregated across markets. Apart from the previous terms, and to improve clarity in the financial modeling across the formulation, the discounting of cash flows is carried out inside the environment itself, as indicated by the last term in the formula. Thus, the discount rate used in RL implementations for training purposes \textit{(\(\gamma\) in the RLLIB library)} should be equal to one across tests and simulations. This approach enables the differentiation of discount rates across agents and/or markets, a feature not explored in the current work demonstrates the capabilities and flexibilities in the RL modeling framework.  

\begin{equation}
R_{\text{t}} = \left( P_{\text{M}} + P_{\text{CM}} + P_{\text{CfD}} - \left( IC_{\text{M}} + IC_{\text{CM}} + IC_{\text{CfD}} \right) \right) \left(\frac{1}{(1 + r)^t}\right) 
\label{Appendix - eq:total_reward}
\end{equation}

Expressions (\ref{Appendix - eq:profit_M}) to (\ref{Appendix - eq:profit_CM}) disaggregate the revenues and costs for the market profits. To improve readability, expressions are formulated for one period, one technology, and one generation asset. Still, they should be extended according to the GENCOs portfolio and all hours in the Equivalent short-term market. Starting from the profit for merchant investments, shown in equation (\ref{Appendix - eq:profit_M}), it is assumed that GENCO produces and dispatches quantities \( Q_{\text{M}} \) using an asset with installed capacity \(\overline{Q_{\text{M}}}\) and fixed costs \(C_{\text{fix}}\). Energy from this asset is sold to the market at a price \( \pi_{\text{M}} \), while total variable production costs are given by the sum of standard variable costs \( C_{\text{var}} \), and the carbon tax pass-through \( C_{\text{CO}_2 \text{tax}} \). The carbon tax depends on the emission rate from the particular technology, as indicated in Table \ref{Table: Generation Technology characteristics}. Importantly, as explained in \ref{Appendix - Merchant Investments}, the energy from merchant investments is paid at a price that, depending on market conditions, can reach up to the VoLL. Moreover, this formulation applies to existing and new assets in the merchant case given the undifferentiated treatment they receive in the model. 

\begin{equation}
P_{\text{M}} = \pi_{\text{M}} Q_{\text{M}} - \left(C_{\text{var}} + C_{\text{CO}_2 \text{tax}} \right) Q_{\text{M}} - C_{\text{fix}}\overline{Q_{\text{M}}}
\label{Appendix - eq:profit_M}
\end{equation}

The expression (\ref{Appendix - eq:profit_CfD}) presents the profit from the CfD market. As before, it is assumed that GENCO produces and dispatches in the market renewable energy in quantities \( Q_{\text{CfD}} \) using an asset with installed capacity \(\overline{Q_{\text{CfD}}}\) and fixed costs \(C_{\text{fix}}\). Yet, instead of prices from the short-term market, the RES production is paid at the CfD auction price \( \pi_{\text{CfD}} \) used by the asset to enter the market. Similarly, total variable production costs are given by the sum of standard variable costs \( C_{\text{var}} \), and the carbon tax pass-through \( C_{\text{CO}_2 \text{tax}} \). This formulation is equivalent to assuming the GENCO sells its energy in the short-term market. It later settles the Contract for Difference financially, using the CfD auction price as a reference.  Furthermore, it is worth noting the CfD hedges the GENCO against price variations but not against potential reductions in the plant's output \( Q_{\text{CfD}} \), which might be caused by unexpected failures or curtailment from the market itself.  

\begin{equation}
P_{\text{CfD}} = \pi_{\text{CfD}} Q_{\text{CfD}} - \left(C_{\text{var}} + C_{\text{CO}_2 \text{tax}} \right) Q_{\text{CfD}} - C_{\text{fix}}\overline{Q_{\text{CfD}}}
\label{Appendix - eq:profit_CfD}
\end{equation}

Equation (\ref{Appendix - eq:profit_CM}) represents the profit from the Capacity Market. As before, the GENCO produces quantities \( Q_{\text{CM}} \) using an asset with installed capacity \(\overline{Q_{\text{CM}}}\) and fixed costs \(C_{\text{fix}}\). Additionally, total variable production costs are given by the sum of standard variable costs \( C_{\text{var}} \), and the carbon tax pass-through \( C_{\text{CO}_2 \text{tax}} \). However, extra terms are included to represent the Reliability option from the Capacity Market, and its interaction with the short-term market. Starting from \( Q_{\text{FirmCapacity}}\), corresponding to the Firm Capacity presented and allocated by the GENCO to the Capacity Market Auction, \( \pi_{\text{Premium}\text{CM}} \), representing the Premium of the Reliability Option obtained from such auction, and \( \pi_{\text{Scarcity}\text{CM}}\), associated to the Reliability Option Strike Price. Within this notation, it is assumed that GENCO sells all its production \( Q_{\text{CM}} \) at the short-term market price \( \pi_{\text{M}} \). Moreover, the Firm Capacity \( Q_{\text{FirmCapacity}}\) is remunerated at price \( \pi_{\text{Premium}\text{CM}} \). Furthermore, the last term in equation (\ref{Appendix - eq:profit_CM}) settles the Call Option, which becomes active when system prices \( \pi_{\text{M}} \) surpass the Strike Price \(\pi_{\text{Scarcity}\text{CM}}\). In short, this formulation partially limits rents received by the GENCO from energy sold in the short-term market, while making it liable for the quantities committed in the Capacity Market when a scarcity event occurs. 

\begin{equation}
P_{\text{CM}} = \pi_{\text{M}} Q_{\text{CM}} - \left(C_{\text{var}} + C_{\text{CO}_2 \text{tax}} \right) Q_{\text{CM}} - C_{\text{fix}}\overline{Q_{\text{CM}}} + Q_{\text{FirmCapacity}} \pi_{\text{Premium}\text{CM}} - MAX[\pi_{\text{M}}-\pi_{\text{Scarcity}\text{CM}}, 0]Q_{\text{FirmCapacity}}
\label{Appendix - eq:profit_CM}
\end{equation}

In the presence of ESSs in the GENCOs portfolio, which are enabled to participate in the system both as merchant investments and through the Capacity Market, the previous expressions are adjusted to accommodate the bidirectional flow of energy, as shown in Equation (\ref{Appendix - eq:profit_M_battery}) and (\ref{Appendix - eq:profit_CM_battery}). Specifically, it is assumed GENCOs discharge \(\left( Q_{\text{M}\text{\_Discharge}},Q_{\text{CM}\text{\_Discharge}} \right) \) and recharge \(\left( Q_{\text{M}\text{\_Charge}},Q_{\text{CM}\text{\_Charge}} \right) \) their ESS by trading electricity in the short-term market. In this context, quantities are defined by the ESS scheduler, ensuring no charging and discharging occur concurrently. Moreover, no variable costs are included for ESS cycling, while fixed costs depend on the ESS's capacity \( Q_{\text{M}\text{\_ESS}} \).

\begin{equation}
P_{\text{M}\text{\_ESS}} = \pi_{\text{M}}  Q_{\text{M}\text{\_Discharge}}  - \pi_{\text{M}}  Q_{\text{M}\text{\_Charge}} - C_{\text{fix}}\overline{Q_{\text{M}\text{\_ESS}}}
\label{Appendix - eq:profit_M_battery}
\end{equation}

\begin{equation}
P_{\text{CM}\text{\_ESS}} = \pi_{\text{M}}  \left( Q_{\text{M}\text{\_Discharge}} - Q_{\text{M}\text{\_Charge}} \right) - C_{\text{fix}}\overline{Q_{\text{CM}\text{\_ESS}}} + \pi_{\text{Premium}\text{CM}}Q_{\text{FirmCapacity}\text{\_ESS}} - MAX[\pi_{\text{M}}-\pi_{\text{Scarcity}\text{CM}}, 0]Q_{\text{FirmCapacity}\text{\_ESS}}
\label{Appendix - eq:profit_CM_battery}
\end{equation}

It is worth noting that terms \(IC_{\text{M}}, IC_{\text{CM}},\) and \(IC_{\text{CfD}}\) have values different than zero when the GENCO has assets under construction. If such is the case, the amount payable by GENCOs at each environment step is a proportional share of the total investment, determined by dividing the total investment cost by the number of equivalent periods required for construction. For example, on the test bench, an asset with a three-year construction timeline would distribute its total CAPEX across 36 smaller payments.

Finally, to improve the readability of reward results obtained from the environment, a normalization factor, \( nf = VoLL^2 Env_{\text{length}} \), based on the VoLL and the environment length in years, \( Env_{\text{length}} \), is applied. No effect from such normalization has been observed during training. 

\subsubsection{Action Space}
\label{Appendix - Agent Actions}

In the current implementation, the actions available to GENCO for interacting with their environment are: I.) Merchant Investments, where GENCOs decide the asset capacity; II.) Investments in the Capacity Market, where GENCOs decide the asset capacity and the price for bidding in the corresponding auction; III.) Investments in the CfD, where again GENCOs decide the asset capacity and the bidding price for the auction; and IV.) the reservoir control for mid-term ESS. 

To model these actions in an RL environment, three options from existing Gymnasium implementations can be considered \cite{towers_gymnasium_2023}: continuous variables, multi-discrete variables, and a hybrid scheme between continuous and discrete variables. Continuous action spaces are well-suited for modeling price bids in auctions but lack realism for investment decisions, given that generation and ESS technologies exhibit limited modularity. Conversely, a multi-discrete representation effectively captures investment constraints but may restrict agent flexibility in auction price bidding. A hybrid approach, employing continuous variables for price representation and discrete variables for investment quantities, would ideally balance both requirements.

While the hybrid approach appears optimal at first analysis, its integration within the model presents structural challenges. Specifically, the availability of certain auctions to GENCOs depends on market conditions and must be dynamically incorporated into the learning process. Discrete action masking, which cancels unavailable actions during the gradient calculations of RL algorithms, has demonstrated superior performance compared to penalties and other constraint-handling methods \cite{huang_closer_2022}. However, extending this concept to continuous action spaces is nontrivial and remains underexplored in the literature. Consequently, the model adopts a fully multi-discrete action space, increasing the number of discretization steps for continuous variables to enhance the realism of agent strategies, following the recommendations of \cite{delalleau_discrete_2019}.

In this context, Table \ref{Appendix - Table: List of Actions} provides a comprehensive overview of the actions available to GENCOs, detailing the minimum and maximum values for discretization, the number of discretization steps, and the conversion process to market strategies. However, two key limitations emerge from the definition of these actions, and the implementation in the Gymnasium framework. First, the model regulates the maximum level of investment per agent solely through the maximum rate available for each technology, as specified in Tables \ref{Table: Generation Technology characteristics} and \ref{Table: Energy Storage Technology characteristics}. This approach does not account for other critical factors, such as an agent’s access to financing, which plays a significant role in real-world investment decisions. Second, agents in the model have prior knowledge of auction price caps, a condition that does not typically exist in real markets. Addressing these limitations by incorporating more intrinsic and realistic representations of agent actions remains an important direction for future work.

\begin{table}[hbt!]
\small
\centering
\resizebox{0.9\columnwidth}{!}{
\begin{tabular}{@{}|l|c|c|l|@{}}
\toprule
\multicolumn{1}{|c|}{\textbf{Actions}}   & \textbf{Number of Actions} & \textbf{Discretization steps} & \multicolumn{1}{c|}{\textbf{Value for Normalization}} \\ \midrule
Quantities for Merchant Investments      & 7                          & 4                             & Maximum allowed investment per technology             \\ \midrule
Quantities for CfD auction               & 3                          & 4                             & Maximum allowed investment per technology             \\ \midrule
Prices for CfD auction                   & 3                          & 12                            & Ceiling price for CfD auctions                        \\ \midrule
Quantities for Capacity Market auction   & 7                          & 4                             & Maximum allowed investment per technology             \\ \midrule
Prices for Capacity Market auction       & 7                          & 12                            & Ceiling price for Capacity Market auctions            \\ \midrule
Desired state of charge for mid-term ESS & 1                          & 7                             & Maximum State of Charge (100\%)                        \\ \bottomrule
\end{tabular}%
}
\caption{Action Space for GENCOs, following definitions and descriptions from the Gymnasium terminology \cite{towers_gymnasium_2023}.}
\label{Appendix - Table: List of Actions}
\end{table}

It is worth noting that the number of Actions \textit{(second column)} in Table \ref{Appendix - Table: List of Actions} directly depends on the number of technologies participating in the available markets. As a result, any addition of new technologies to the GENCOs' portfolio comes with a significant increment in the action space, justifying the selection of a relatively small number of investment options, but still representative of the main trends expected during the energy transition. 

\subsubsection{Observation Space}
\label{Appendix - Agent Observations}

The design of the observation set available to GENCOs in the model follows three key principles. First, the selected variables should align with real-world market conditions, where agents can access publicly shared system information while specific details remain private and inaccessible to competitors. Second, no internal forecasting tools are included in the observation set, ensuring that the model retains full autonomy in decision-making. Finally, the selection process adheres to the principle that any information available in the market is also provided to the agents.

Considering these principles, Table \ref{Appendix - Table: List of Observations} lists the observations made available to agents across environment steps, subdivided according to their thematic category. The set of observations includes indicative values for the time series modeling demand and variable resource availability, the composition of the energy mix via the assets in operation in the system, individual and system-wide reservoir levels, market information \textit{(prices, balances, scarcity)}, the aggregated reward per technology in these markets, policy-relevant information \textit{(the carbon tax level)}, and the time associated to the environment step. The framework designed for the GENCO's observation does not depend on the number of agents in the market, as they harness aggregated system information given that the number of agents in each simulation is constant.

\begin{table}[hbt!]
\centering
\resizebox{\textwidth}{!}{%
\begin{tabular}{@{}|l|c|c|l|@{}}
\toprule
\multicolumn{1}{|c|}{\textbf{Observations}}                             & \textbf{Number of observations} & \textbf{Type of normalization} & \multicolumn{1}{c|}{\textbf{Reference for Normalization}}        \\ \midrule
Short term prices                                                       & 24                              & (\ref{eq:norm_1})                            & VoLL                                                             \\ \midrule
Short-term Solar and Wind Resource   Availability                       & 2                               & (\ref{eq:norm_4})                            & N/A                                                              \\ \midrule
Long-term Solar and Wind Resource   Availability                        & 2                               & (\ref{eq:norm_4})                            & N/A                                                              \\ \midrule
Short-term hydrologic inflows                                           & 1                               & (\ref{eq:norm_2})                            & Short-term demand projection                                     \\ \midrule
Long-term hydrologic inflows                                            & 1                               & (\ref{eq:norm_2})                            & Long-term demand projection                                      \\ \midrule
Short-term demand projection                                            & 1                               & (\ref{eq:norm_1})                            & Maximum system demand                                            \\ \midrule
Long-term demand projection                                             & 1                               & (\ref{eq:norm_1})                            & Maximum system demand                                            \\ \midrule
Individual Capacity per Generation and   ESS technology                 & 8                               & (\ref{eq:norm_2})                            & Long-term demand projection divided per number of agents         \\ \midrule
Total    Capacity per Generation and ESS technology                     & 8                               & (\ref{eq:norm_2})                            & Long-term demand projection                                      \\ \midrule
Individual Existing Capacity per   Generation and ESS technology        & 8                               & (\ref{eq:norm_2})                            & GENCO total installed Capacity per Generation and ESS technology \\ \midrule
Total Existing Capacity per Generation   and ESS technology             & 8                               & (\ref{eq:norm_2})                            & Total installed Capacity per Generation and ESS technology       \\ \midrule
Individual Merchant Capacity per   Generation and ESS technology        & 7                               & (\ref{eq:norm_2})                            & GENCO total installed Capacity per Generation and ESS technology \\ \midrule
Total Merchant Capacity per Generation   and ESS technology             & 7                               & (\ref{eq:norm_2})                            & Total installed Capacity per Generation and ESS technology       \\ \midrule
Individual CfD Capacity per Generation   and ESS technology             & 3                               & (\ref{eq:norm_2})                            & GENCO total installed Capacity per Generation and ESS technology \\ \midrule
Total CfD Capacity per Generation and ESS   technology                  & 3                               & (\ref{eq:norm_2})                            & Total installed Capacity per Generation and ESS technology       \\ \midrule
Individual Capacity Market Capacity per   Generation and ESS technology & 7                               & (\ref{eq:norm_2})                            & GENCO total installed Capacity per Generation and ESS technology \\ \midrule
Total Capacity Market Capacity per   Generation and ESS technology      & 7                               & (\ref{eq:norm_2})                            & Total installed Capacity per Generation and ESS technology       \\ \midrule
System Margins in the CfD and Capacity   Market                         & 2                               & (\ref{eq:norm_2})                            & Maximum system demand                                            \\ \midrule
Price in the CfD and Capacity Market                                    & 2                               & (\ref{eq:norm_1})                            & Maximum price from the corresponding market                      \\ \midrule
Scarcity in the CfD and Capacity Market                                 & 2                               & (\ref{eq:norm_4})                            & N/A                                                              \\ \midrule
Capacity Credits per technology in the   Capacity Market                & 7                               & (\ref{eq:norm_4})                            & N/A                                                              \\ \midrule
Carbon Tax                                                              & 1                               & (\ref{eq:norm_1})                            & Maximum Carbon Tax                                               \\ \midrule
Bimester, Year, and total simulation   length                           & 3                               & (\ref{eq:norm_1})                            & Maximum value per time category                                  \\ \midrule
Accumulated Reward of existing assets                                   & 8                               & (\ref{eq:norm_3})                            & Normalization factor                                             \\ \midrule
Accumulated Reward of Merchant   Investments                            & 7                               & (\ref{eq:norm_3})                            & Normalization factor                                             \\ \midrule
Accumulated Reward of CfD investments                                   & 3                               & (\ref{eq:norm_3})                            & Normalization factor                                             \\ \midrule
Accumulated Reward of Capacity Market   investments                     & 7                               & (\ref{eq:norm_3})                            & Normalization factor                                             \\ \bottomrule
\end{tabular}%
}
\caption{Observation Space for GENCOs, following definitions and descriptions from the Gymnasium terminology \cite{towers_gymnasium_2023}.}
\label{Appendix - Table: List of Observations}
\end{table}

All observations listed in Table \ref{Appendix - Table: List of Observations} are defined inside the \([-1,1]\), with four normalization methods used to enforce such boundaries, showcased in the third and fourth columns. First, when maximum values for the observations are well defined, such as the participation of particular technologies in the energy mix, normalization is performed against these values, as indicated by Equation (\ref{eq:norm_1}). Second, when an observation involves the comparison of two positive quantities, \( Obs_{\text{a}}\) and \( Obs_{\text{b}}\), expression (\ref{eq:norm_2}) is applied. Third, for aggregated reward observations, \( Obs_{\text{r}}\), where predefined minimum/maximum values are unknown beforehand, normalization is performed relative to the reward normalization factor, as defined in (\ref{eq:norm_3}). Finally, for observation of variables ranging between zero and one, such as the availability of renewable resources, the normalization from Equation (\ref{eq:norm_4}) is used. Nonetheless, to further safeguard the implementation, all observations are clipped to be within the interval before they are fed to the RL algorithm. Apart from these treatments, no additional changes, filters, or importance weighting are applied to system observations. 

\begin{equation}
Obs_{\text{adjusted}\text{\_1}} = \frac{2Obs_{\text{true}}}{Obs_{\text{max}}}-1
\label{eq:norm_1}
\end{equation}

\begin{equation}
Obs_{\text{adjusted}\text{\_2}} = \frac{Obs_{\text{a}} - Obs_{\text{b}}}{Obs_{\text{a}} + Obs_{\text{b}}}
\label{eq:norm_2}
\end{equation}

\begin{equation}
Obs_{\text{adjusted}\text{\_3}} = \frac{Obs_{\text{r}}}{nf} = \frac{Obs_{\text{r}}}{VoLL^2 Env_{\text{length}}}
\label{eq:norm_3}
\end{equation}

\begin{equation}
Obs_{\text{adjusted}\text{\_1}} = 2Obs_{\text{true}}-1
\label{eq:norm_4}
\end{equation}

\subsubsection{Initial and Absorbing states}
\label{Appendix - Final state}
For the simulation and the environment to be coherent with the intended modeling objectives, further considerations are needed for the initial and absorbing states in the environment. Regarding the start of simulations, it is worth noting that the environment does not allow to represent an initial construction pipeline. However, similar behavior can be achieved if the simulation starts and maintains stable demand and generation mix conditions for a few years, thus allowing the agents to generate the pipeline of projects under construction by themselves. 

Regarding the simulation ending, initial testing revealed that agents tended to avoid investments when no compensation was provided for assets that had not yet reached the end of their operational lifetime. An equivalent remuneration mechanism is introduced in the absorbing state to address this issue. This payment, denoted \(\ P{\text{absorbing}}\),  compensates assets for their remaining operational lifetime at the end of the simulation, as described in expression (\ref{eq:annuity absorbing state}). The payment represents the net present value of an average income, \(\ I{\text{mean}}\), when such an income is applied over the remaining lifetime of projects, \(\ t_{\text{r}\text{\_l}} \). The system-wide average profit from all assets is considered to approximate the average income. Furthermore, to better capture the heterogeneity of assets in GENCOs' portfolios, the calculation in Equation (\ref{eq:annuity absorbing state}) is extended to be independently performed for each technology and market. This refinement enhances the differentiation of asset valuations, ensuring a more realistic representation of investment incentives.

\begin{equation}
P{\text{absorbing}}=I{\text{mean}}  \left( \frac{1 - (1 + r)^{-t_{\text{r}\text{\_l}}}}{r} \right)
\label{eq:annuity absorbing state}
\end{equation}

Finally, two further modifications are introduced to the final environment steps to ensure a smooth end of simulations. First, agents are restricted from initiating investments that cannot reach operational status before the simulation terminates. For example, if three years remain in the simulation, GENCOs can only invest in assets with less than three years of construction. Second, similar to the initialization phase, the final periods maintain stable demand and a balanced generation mix. This prevents artificial scarcity or overproduction conditions from distorting market dynamics, and thus decisions taken by agents in the simulation. 

Overall, these adjustments to the initial and absorbing states enhance the simulation dynamics and facilitate smoother transitions within the environment. However, they come at the cost of additional environment steps. Specifically, to accurately simulate 20 years of operation, a further five years are incorporated \textit{(two at the beginning of the simulation and three at the end)}. While these extra steps improve stability, they extend the total simulation time without directly contributing to the modeling of relevant agent dynamics.

\clearpage

\section{Training and Hyperparameter selection}
\label{Appendix - Supplementary information - Training}

\begin{figure}[hbt!]
    \centering
    \includegraphics[width=1\linewidth]{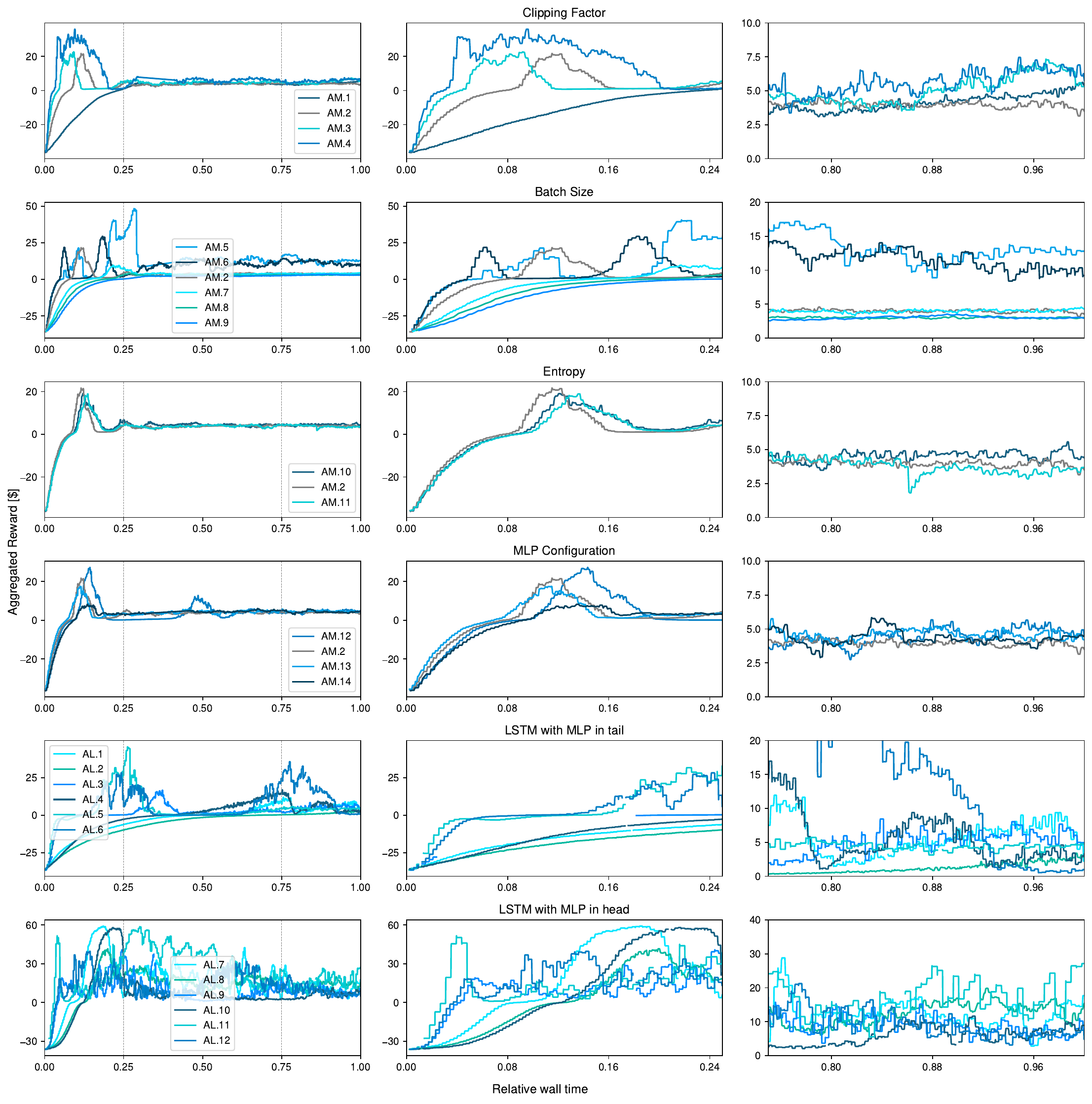} %
    \caption{Average system reward during training for hyperparameter configurations in the EoM environment. In the left panels, the complete training curve is presented. Center panels truncate relative wall time between 0 and 0.25 to better showcase the system's behavior in the early stages of training. In contrast, right panels limit relative wall time between 0.75 and 1 to represent the last stages of training.}
\label{Appendix - fig: Training Behavior - EoM}
\end{figure}

\begin{figure}[hbt!]
    \centering
    \includegraphics[width=1\linewidth]{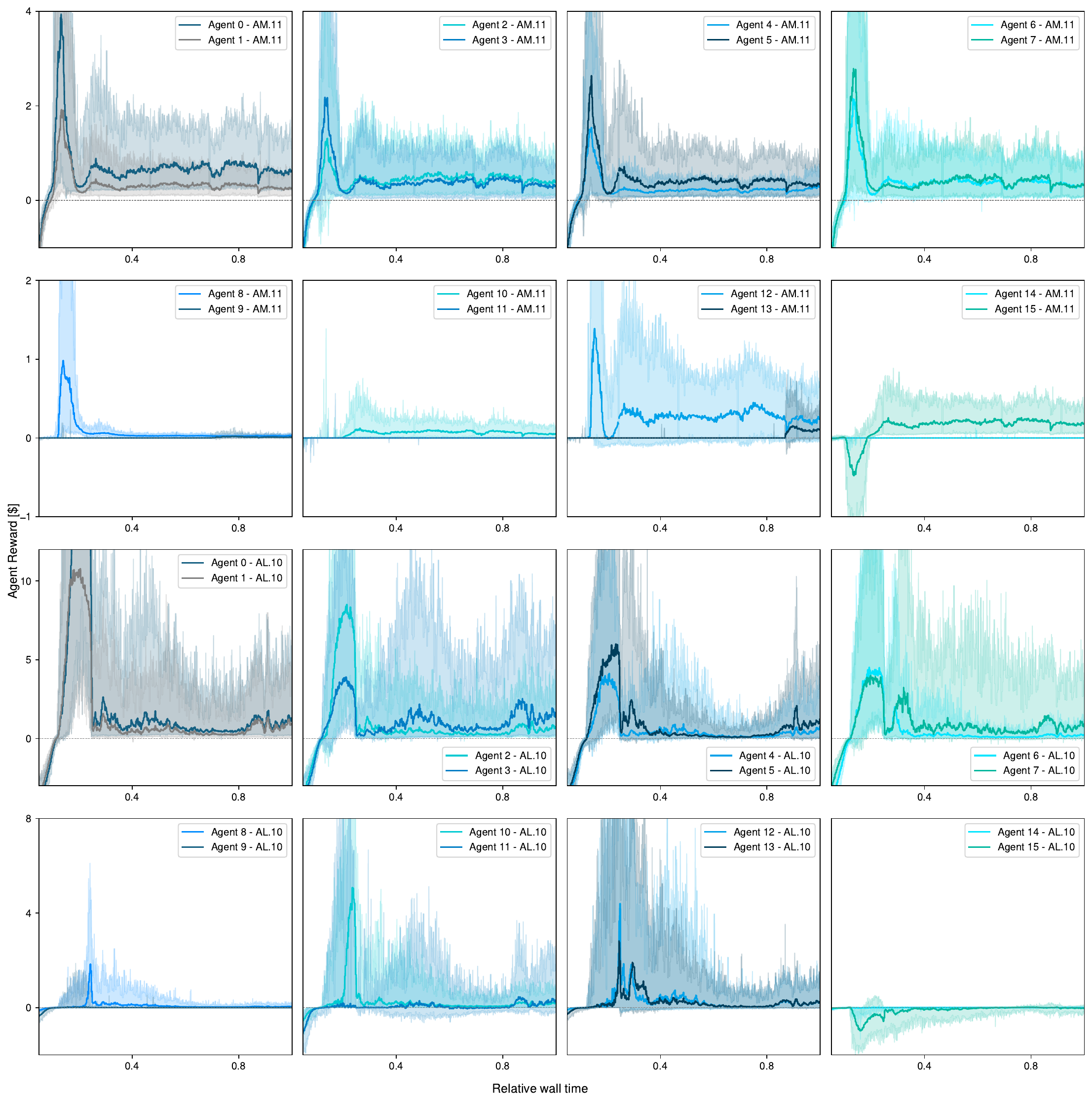} %
    \caption{Agent Reward during training for hyperparameter configurations AM. 11 and AL. 10 in the EoM environment. The solid line represents the average reward, while shaded areas indicate the maximum and minimum reward sampled along the training. Relative wall time is truncated between 0.1 and 1 to better showcase the agents' behavior in the last stages of training.}
\label{Appendix - fig: Agent training - EoM}
\end{figure}

\begin{figure}[hbt!]
    \centering
    \includegraphics[width=1\linewidth]{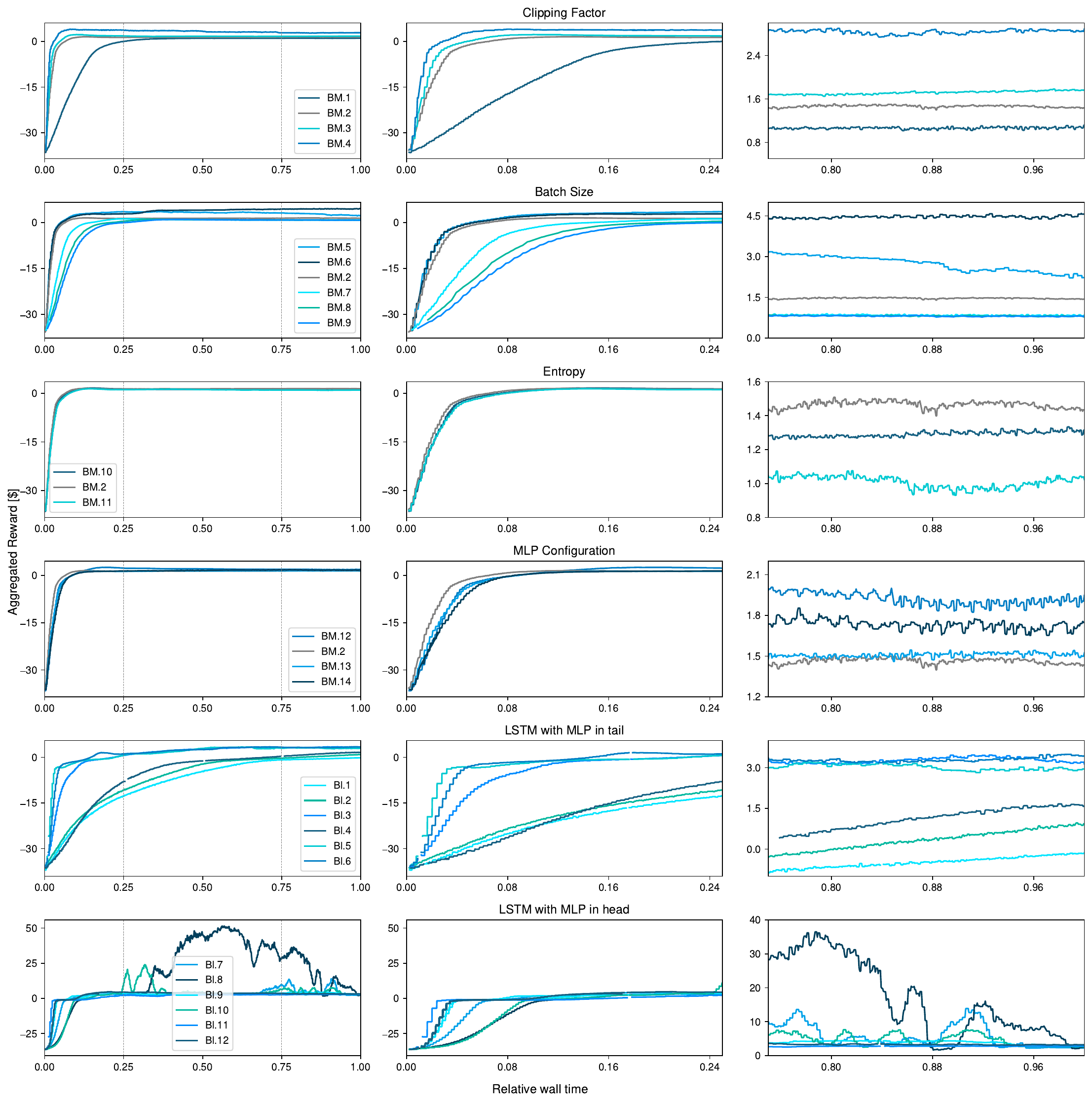} %
    \caption{Average system reward during training for hyperparameter configurations in the CM + CfD environment. In the left panels, the complete training curve is presented. Center panels truncate relative wall time between 0 and 0.25 to better showcase the system's behavior in the early stages of training. In contrast, right panels limit relative wall time between 0.75 and 1 to represent the last stages of training.}
\label{Appendix - fig: Training Behavior - CM + CfD}
\end{figure}

\begin{figure}[hbt!]
    \centering
    \includegraphics[width=1\linewidth]{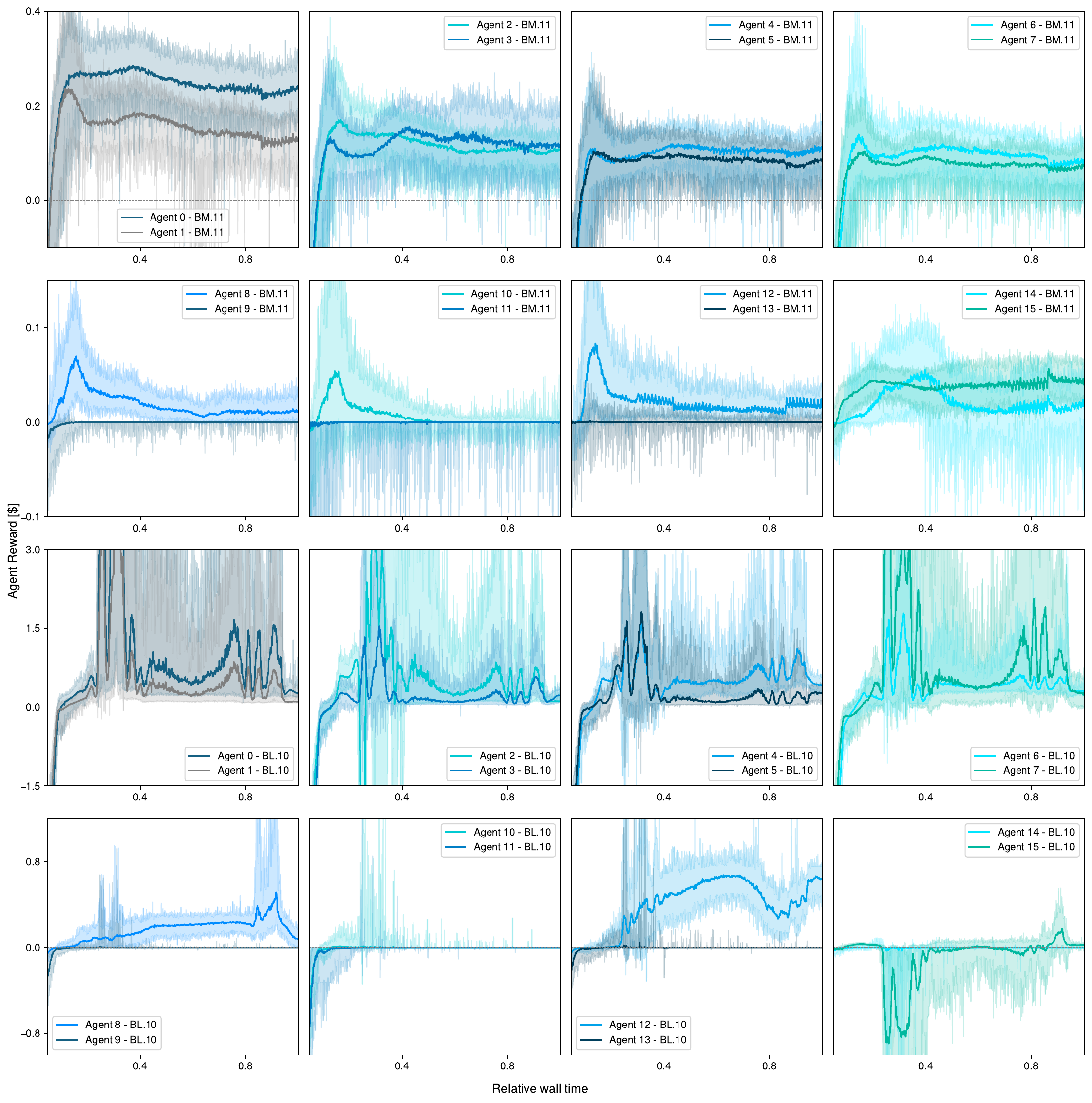} %
    \caption{Agent Reward during training for hyperparameter configurations BM. 11 and BL. 10 in the CM + CfD environment. The solid line represents the average reward, while shaded areas indicate the maximum and minimum reward sampled along the training. Relative wall time is truncated between 0.1 and 1 to better showcase the agents' behavior in the last stages of training.}
\label{Appendix - fig: Agent training - CM + CfD}
\end{figure}

\begin{figure}[hbt!]
    \centering
    \includegraphics[width=1.0\linewidth]{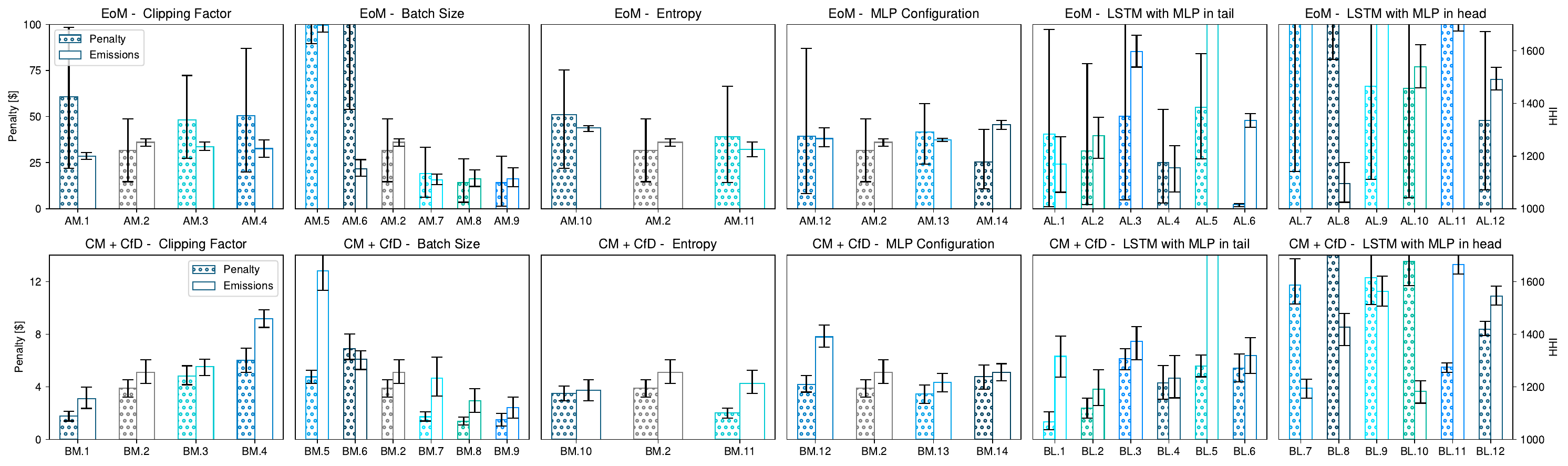} %
    \caption{Penalty and HHI for hyperparameter configurations in the EoM, upper panel, and CM + CfD, lower panel, environments.  Results are obtained using 100 episodes in the environment with the most updated agents' versions. Error bars showcased the 10th and 90th percentile from the 100 episodes. The Y axes in the Figure are adjusted to facilitate comparison among the most relevant hyperparameter configurations.}
\label{Appendix - fig: Penalty and HHI for hyperparameter configurations in the EoM and CM + CfD environments.}
\end{figure}

This section presents relevant information for the hyperparameter search. Particularly:
\begin{itemize}
    \item Table \ref{Appendix - Table: List of hyperparameters used during search.} introduces the base hyperparameters used for training agents using the IPPO algorithm. Notation follows the structure of RLLIB. Changes are applied over the base configuration for the hyperparameter search according to the information presented in Table \ref{Appendix - Table: Full Notation for Tests Conducted During Hyperparameter Search.}. 
    \item Figures \ref{Appendix - fig: Training Behavior - EoM} and \ref{Appendix - fig: Agent training - EoM} introduce the training profiles obtained during the hyperparameter search for the Energy-only-market environment at the system and agent level, respectively. 
    \item Figures \ref{Appendix - fig: Training Behavior - CM + CfD} and \ref{Appendix - fig: Agent training - CM + CfD} similarly present the training profiles, but for the environment with the capacity and contract for difference markets. 
    \item Figure \ref{Appendix - fig: Penalty and HHI for hyperparameter configurations in the EoM and CM + CfD environments.} and Table \ref{Appendix - Table: League competition results.} presents additional results regarding the comparison metrics for the hyperparameter runs. 
\end{itemize}

\begin{table}[hbt!]
\small
\centering
\resizebox{\textwidth}{!}{%
\begin{tabular}{|c|c|c|c|}
\hline
\textbf{Parameter}  & \textbf{Value}      & \textbf{Parameter}                & \textbf{Value} \\ \hline
Clipping factor - \( \epsilon\)     & 0.05                & Number of SGA iterations          & 10             \\ \hline
Batch Size          & 17664               & Weight of Value function loss - \( v \) & -inf           \\ \hline
Entropy Coefficient - \(h\) & 0.00001             & Discount Factor - \( \gamma\)                  & 1              \\ \hline
Network Architecture for MLP tests & [256-256] & Discount factor in Generalized Advantage Estimation - \( \lambda\) & 0.995 \\ \hline
MLP for head or tail in LSTM       & [128-128] & Shared layers between Value and Policy networks    & No    \\ \hline
Learning rate       & 0.0003              & Parallel environments             & 69             \\ \hline
Size of minibatch   & Equal to batch size & Workers per parallel environments & 8              \\ \hline
\end{tabular}%
}
\caption{List of hyperparameters used during search. Notation is kept in line with definitions from the RLLIB Library.}
\label{Appendix - Table: List of hyperparameters used during search.}
\end{table}

\begin{table}[hbt!]
\small
\centering
\resizebox{\textwidth}{!}{%
\begin{tabular}{|c|c|c|c|c|}
\hline
\textbf{\begin{tabular}[c]{@{}c@{}}Code\\ EoM\end{tabular}} &
  \textbf{\begin{tabular}[c]{@{}c@{}}Code\\ CM + CfD\end{tabular}} &
  \textbf{Network Type} &
  \textbf{Test description} &
  \textbf{Test ranges} \\ \hline
\begin{tabular}[c]{@{}c@{}}AM.1, AM.2,\\ AM.3, AM.4\end{tabular} &
  \begin{tabular}[c]{@{}c@{}}BM.1, BM.2, \\ BM.3, BM.4\end{tabular} &
  MLP &
  Clipping Factor - \( \epsilon\)&
  {[}0.01, 0.05, 0.1, 0.2{]} \\ \hline
\begin{tabular}[c]{@{}c@{}}AM.5, AM.6, AM.2, \\ AM.7, AM.8, AM.9\end{tabular} &
  \begin{tabular}[c]{@{}c@{}}BM.5, BM.6, BM.2, \\ BM.7, BM.8, BM.9\end{tabular} &
  MLP &
  Batch Size &
  \begin{tabular}[c]{@{}c@{}}{[}4416, 8832, 17664, \\ 35328, 70656, 88320{]}\end{tabular} \\ \hline
AM.10, AM.2, AM.11 &
  BM.10, BM.2, BM.11 &
  MLP &
  Entropy - \(h\)&
  {[}0,0.00001, 0.01{]} \\ \hline
\begin{tabular}[c]{@{}c@{}}AM.12, AM.2, \\ AM.13, AM.14\end{tabular} &
  \begin{tabular}[c]{@{}c@{}}BM.12, BM.2, \\ BM.13, BM.14\end{tabular} &
  MLP &
  Network Configuration &
  \begin{tabular}[c]{@{}c@{}}{[}{[}64-64{]}, {[}256-256{]}, \\ {[}512-512{]}, {[}1024-1024{]}{]}\end{tabular} \\ \hline
AL.1, AL.2 &
  BL.1, BL.2 &
  LSTM &
  \begin{tabular}[c]{@{}c@{}}MLP in tail\\ \(\ h_{LSTM}=\) 16 + \(\ L_{max}=\) 8\\ Batch Size\end{tabular} &
  {[}8832, 17664{]} \\ \hline
AL.3, AL.4 &
  BL.3, BL.4 &
  LSTM &
  \begin{tabular}[c]{@{}c@{}}MLP in tail\\  \(\ h_{LSTM}=\) 32 + \(\ L_{max}=\) 16  \\ Batch Size\end{tabular} &
  {[}4416, 8832{]} \\ \hline
AL.5, AL.6 &
  BL.5, BL.6 &
  LSTM &
  \begin{tabular}[c]{@{}c@{}}MLP in tail\\  \(\ h_{LSTM}=\) 64 + \(\ L_{max}=\) 32\\ Batch Size\end{tabular} &
  {[}2208, 4416{]} \\ \hline
AL.7, AL.8 &
  BL.7, BL.8 &
  LSTM &
  \begin{tabular}[c]{@{}c@{}}MLP in head \\ \(\ h_{LSTM}=\) 16 + \(\ L_{max}=\) 8  \\ Batch Size\end{tabular} &
  {[}8832, 17664{]} \\ \hline
AL.9, AL.10 &
  BL.9, BL.10 &
  LSTM &
  \begin{tabular}[c]{@{}c@{}}MLP in head\\  \(\ h_{LSTM}=\) 32 + \(\ L_{max}=\) 16  \\ Batch Size\end{tabular} &
  {[}4416, 8832{]} \\ \hline
AL.11, AL.12 &
  BL.11, BL.12 &
  LSTM &
  \begin{tabular}[c]{@{}c@{}}MLP in head\\ \(\ h_{LSTM}=\) 64 + \(\ L_{max}=\) 32  \\ Batch Size\end{tabular} &
  {[}2208, 4416{]} \\ \hline
\end{tabular}%
}
\caption{Relevant information for tests conducted during hyperparameter search. For individual tests, hyperparameters not mentioned in the corresponding field are not modified, and the values from Table \ref{Appendix - Table: List of hyperparameters used during search.} are used instead. The parameter \(\ h_{LSTM}\) refers to the LSTM cell size, and \(\ L_{max}\) is the maximum sequence length. Consequently, apart from the network configuration, LSTM tests are carried out using the Entropy and Clipping factor from the M.2 case.}
\label{Appendix - Table: Full Notation for Tests Conducted During Hyperparameter Search.}
\end{table}

\begin{table}[hbt!]
\centering
\resizebox{\textwidth}{!}{%
\begin{tabular}{|c|c|cc|cc|cc|cc|cc|}
\hline
\multirow{2}{*}{\textbf{Category}} &
  \multirow{2}{*}{\textbf{Run/Score}} &
  \multicolumn{2}{c|}{\textbf{Round 1}} &
  \multicolumn{2}{c|}{\textbf{Round 2}} &
  \multicolumn{2}{c|}{\textbf{Round 3}} &
  \multicolumn{2}{c|}{\textbf{Round 3}} &
  \multicolumn{2}{c|}{\textbf{Overall Rank}} \\ \cline{3-12} 
 &
   &
  \multicolumn{1}{c|}{\textbf{A - EoM}} &
  \textbf{B - CM + CfD} &
  \multicolumn{1}{c|}{\textbf{A - EoM}} &
  \textbf{B - CM + CfD} &
  \multicolumn{1}{c|}{\textbf{A - EoM}} &
  \textbf{B - CM + CfD} &
  \multicolumn{1}{c|}{\textbf{A - EoM}} &
  \textbf{B - CM + CfD} &
  \multicolumn{1}{c|}{\textbf{A - EoM}} &
  \textbf{B - CM + CfD} \\ \hline
\multirow{4}{*}{Clipping Factor}                                                       & M.1  & 0.29 & 0.36 & 0.59 & 0.50 & -    & -    & -    & -    & 16 & 11 \\
                                                                                       & M.2  & 0.41 & 0.37 & 0.44 & 0.47 & 0.54 & 0.58 & -    & -    & 6  & 8  \\
                                                                                       & M.3  & 0.37 & 0.31 & 0.45 & 0.47 & 0.52 & 0.76 & 0.66 & -    & 4  & 9  \\
                                                                                       & M.4  & 0.45 & 0.36 & 0.54 & 0.57 & -    & -    & -    & -    & 13 & 12 \\ \hline
\multirow{6}{*}{Batch Size}                                                            & M.5  & 0.50 & 0.47 & 0.56 & 0.60 & -    & -    & -    & -    & 15 & 13 \\
                                                                                       & M.2  & 0.44 & 0.49 & 0.62 & 0.64 & -    & -    & -    & -    & 17 & 14 \\
                                                                                       & M.6  & 0.41 & 0.37 & 0.44 & 0.47 & 0.54 & 0.58 & -    & -    & 6  & 8  \\
                                                                                       & M.7  & 0.25 & 0.24 & 0.27 & 0.24 & 0.32 & 0.38 & 0.59 & 0.41 & 3  & 1  \\
                                                                                       & M.8  & 0.21 & 0.24 & 0.30 & 0.23 & 0.29 & 0.32 & 0.35 & 0.44 & 1  & 2  \\
                                                                                       & M.9  & 0.21 & 0.26 & 0.23 & 0.26 & 0.29 & 0.27 & 0.39 & 0.45 & 2  & 3  \\ \hline
\multirow{3}{*}{Entropy}                                                               & M.10 & 0.42 & 0.31 & 0.52 & 0.40 & -    & 0.56 & -    & -    & 12 & 6  \\
                                                                                       & M.2  & 0.41 & 0.37 & 0.44 & 0.47 & 0.54 & 0.58 & -    & -    & 6  & 8  \\
                                                                                       & M.11 & 0.40 & 0.33 & 0.48 & 0.42 & 0.51 & 0.49 & 0.73 & 0.79 & 5  & 4  \\ \hline
\multirow{4}{*}{\begin{tabular}[c]{@{}c@{}}MLP   \\      Configuration\end{tabular}}   & M.12 & 0.45 & 0.26 & 0.46 & 0.45 & 0.57 & 0.76 & -    & -    & 8  & 9  \\
                                                                                       & M.2  & 0.41 & 0.37 & 0.44 & 0.47 & 0.54 & 0.58 & -    & -    & 6  & 8  \\
                                                                                       & M.13 & 0.45 & 0.26 & 0.40 & 0.47 & 0.56 & 0.51 & -    & 0.83 & 7  & 5  \\
                                                                                       & M.14 & 0.40 & 0.30 & 0.46 & 0.36 & 0.64 & 0.56 & -    & -    & 9  & 6  \\ \hline
\multirow{6}{*}{\begin{tabular}[c]{@{}c@{}}LSTM with \\      MLP in tail\end{tabular}} & L.1  & 0.43 & 0.77 & 0.50 & -    & 0.73 & -    & -    & -    & 10 & 25 \\
                                                                                       & L.2  & 0.51 & 0.73 & 0.68 & -    & -    & -    & -    & -    & 18 & 24 \\
                                                                                       & L.3  & 0.60 & 0.52 & -    & 0.80 & -    & -    & -    & -    & 21 & 18 \\
                                                                                       & L.4  & 0.62 & 0.80 & -    & -    & -    & -    & -    & -    & 22 & 26 \\
                                                                                       & L.5  & 0.66 & 0.59 & -    & -    & -    & -    & -    & -    & 26 & 21 \\
                                                                                       & L.6  & 0.65 & 0.56 & -    & -    & -    & -    & -    & -    & 25 & 19 \\ \hline
\multirow{6}{*}{\begin{tabular}[c]{@{}c@{}}LSTM with \\      MLP in head\end{tabular}} & L.7  & 0.64 & 0.61 & -    & -    & -    & -    & -    & -    & 24 & 22 \\
                                                                                       & L.8  & 0.45 & 0.64 & 0.52 & -    & -    & -    & -    & -    & 11 & 23 \\
                                                                                       & L.9  & 0.63 & 0.56 & -    & 0.74 & -    & -    & -    & -    & 23 & 16 \\
                                                                                       & L.10 & 0.50 & 0.56 & 0.56 & 0.75 & -    & -    & -    & -    & 14 & 17 \\
                                                                                       & L.11 & 0.58 & 0.57 & -    & -    & -    & -    & -    & -    & 20 & 20 \\
                                                                                       & L.12 & 0.53 & 0.53 & -    & 0.73 & -    & -    & -    & -    & 19 & 15 \\ \hline
\end{tabular}%
}
\caption{League competition results for the hyperparameter configurations showcased in \ref{Appendix - Table: Full Notation for Tests Conducted During Hyperparameter Search.}. In the League, each round consists of 3750 simulations with a random selection of networks for individual agents. Simulations are repeated for 10 episodes each. For each round, the first columns showcase the score during the test, normalized by the maximum score possible \textit{(higher is worse)}. The last columns display the overall ranking of the particular set of hyperparameters. The M.2 run results are repeated in each category to improve clarity.} 
\label{Appendix - Table: League competition results.}
\end{table}

\clearpage

\section{Long-term Electricity Market Results}
\label{Appendix - Supplementary information - Long-term Electricity Market Results}

\begin{figure}[hbt!]
    \centering
    \includegraphics[width=0.7\linewidth]{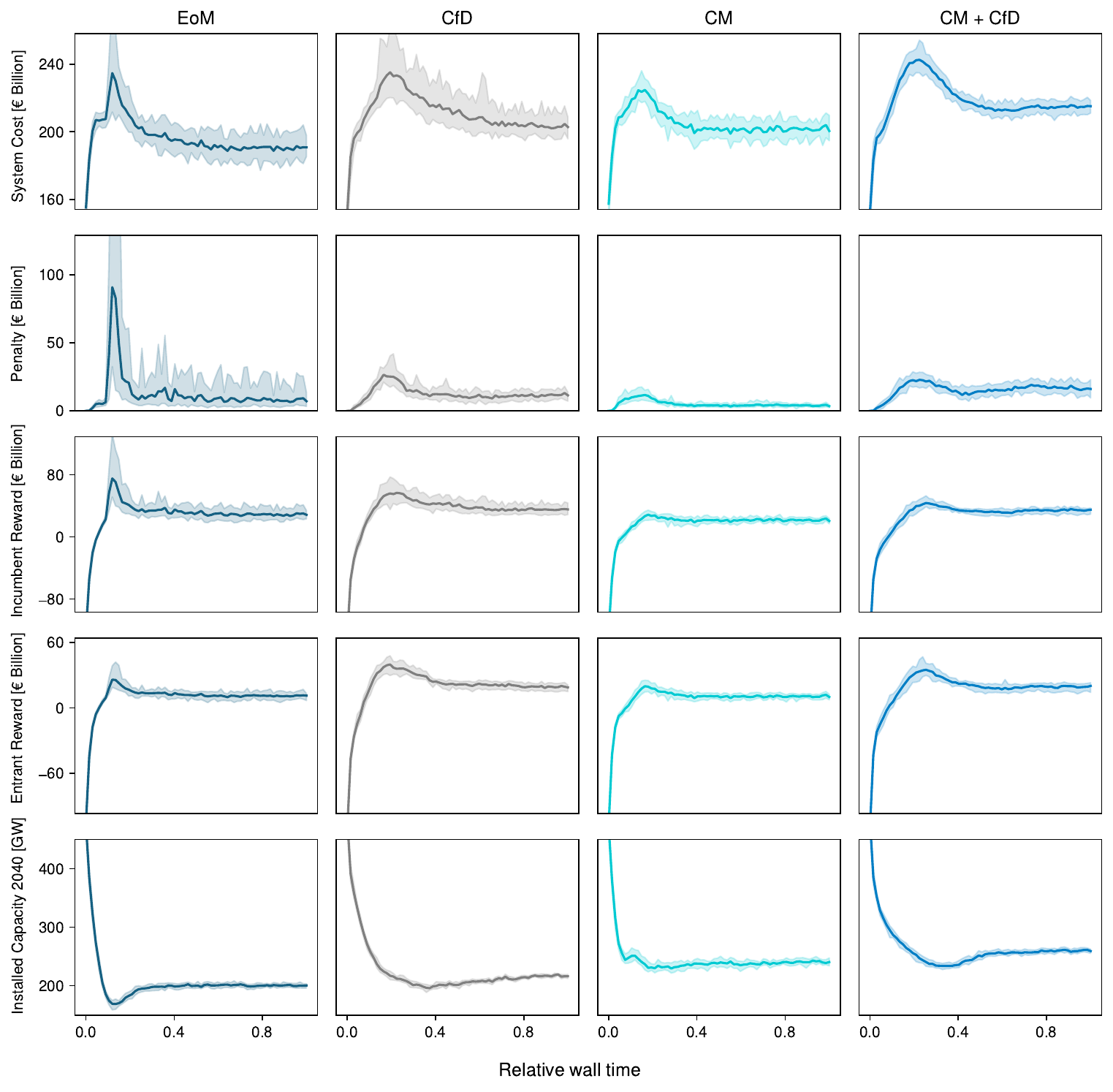} %
    \caption{ Relevant Training Metrics for Different Market Configurations considering 16 agents. From top to bottom, the graph showcases the total electricity system cost, the agents' Penalty, calculated using the procedure described in Section \ref{SubSection - Hyperparameter Selection}, the aggregated reward of agents grouped in incumbent and entrants categories, and the total installed capacity in the year 2040. Solid lines display average values, while shaded areas are the minimum and maximum values during sampling. Results are obtained using intermediate checkpoints during training, and sampled from 25 market simulations.}
\label{Appendix - fig: Training - electricity markets.}
\end{figure}

This section complements the tests and results presented in \ref{Section - Market results} from the main document. Starting with Figure \ref{Appendix - fig: Training - electricity markets.}, which showcases the evolution of relevant metrics during training, across the four market environments, tested with a 16-agent configuration. The three prominent faces of training, briefly described in Section \ref{SubSection - Hyperparameter Selection}, come into play. Training starts with installed capacities considerably higher than those reasonable for the market conditions, leading to consistent negative rewards for agents, and comparatively low system prices. From that point, agents rapidly adjust their investment, shifting the system towards a condition with positive agent rewards and relatively high system prices. Next, the agents re-adjust their strategies, increasing the installed capacity until an equilibrium is reached. This behavior remains largely unaltered across the market designs evaluated. However, introducing long-term market sessions reduces reward levels and increases installed capacities in the system.

\begin{figure}[hbt!]
    \centering
    \includegraphics[width=0.7\linewidth]{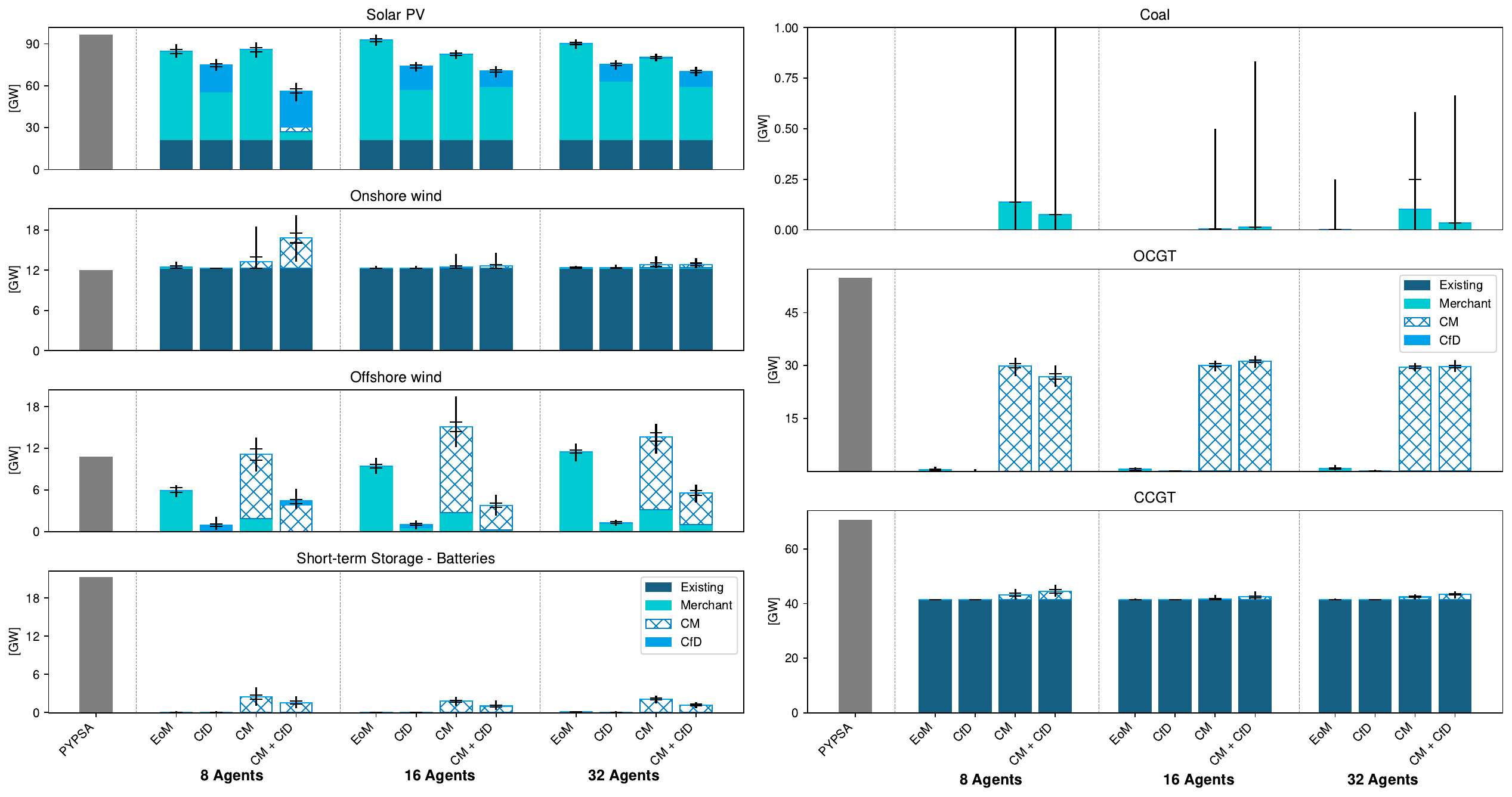} %
    \caption{Installed Capacity of generation and storage technologies in the year 2030 under different market designs and competition levels. Stacked bars indicate average installed capacities across runs. In each stacked bar, the horizontal markers display the 25th and 75th percentiles, and the vertical markers the minimum and maximum values. Columns organize the installed capacities in the four market designs. Furthermore, results are grouped in the column categories, divided by vertical dotted gray lines, according to the number of agents used in the simulation. Moreover, hatching highlights the mechanism used by agents to enter the market. Gray bars, for the corresponding technology, show the output from the central planning exercise carried out in PYPSA. Results from the MARL model are obtained across 200 market simulations using the trained agents.}
\label{Appendix - fig: Installed Capacity 2030.}
\end{figure}

Using trained agents, Figure \ref{Appendix - fig: Installed Capacity 2030.} presents the installed capacities for 2030 according to market design and competition level, complementing the 2040 results showcased in Figure \ref{fig: Installed Capacity 2040.} Across markets, similar trends to those observed by 2040 are present a decade earlier: I) substantial investments in solar PV and OCGT technologies, the two drivers of capacity expansion in the model, II) penetration of offshore wind, mainly through long-term markets, and III) a lag of short-term storage investments when compared to PyPSA results.  Interestingly, compared to 2040, the introduction of CfD auctions leads to a slowdown of Solar PV and offshore Wind merchant investment, which reduces the emission reduction potential in the transition. 

\begin{figure}[hbt!]
    \centering
    \includegraphics[width=0.7\linewidth]{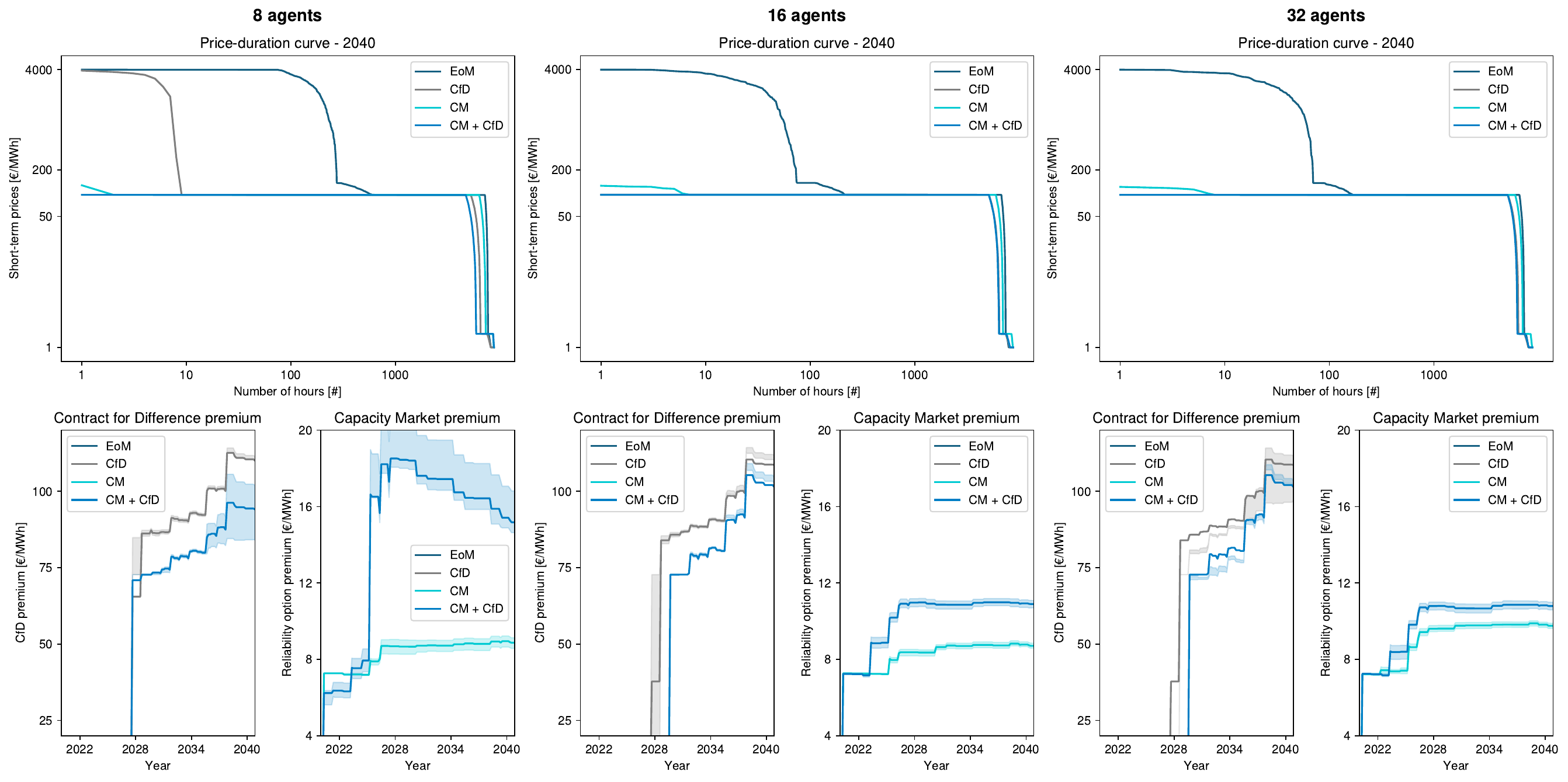} %
    \caption{Price duration curve for the short-term market in 2040, and Prices for the Contract for Difference and Capacity markets for the study period under different market designs and competition levels. The upper panels present short-term market prices, while the lower panels indicate prices for the Contract for Difference and Capacity Markets. The graph's three main columns are dedicated to the tested competition levels. The price duration curve is obtained by interpolating the sampled prices from the simulation and displayed on logarithmic scales to emphasize high-price events. Premiums for the Contract for Difference and Capacity Markets consider the cumulative allocations. Solid lines display average values, while shaded areas are the 25th and 75th percentiles. All results are obtained across 200 market simulations using the trained agents.}
\label{Appendix - fig: Detailed prices 2040.}
\end{figure}

Complementary, Figure \ref{Appendix - fig: Detailed prices 2040.} presents system prices across the simulations, where the difference in long-term market premiums denotes using these auctions to mitigate missing money problems.

\begin{figure}[hbt!]
    \centering
    \includegraphics[width=0.7\linewidth]{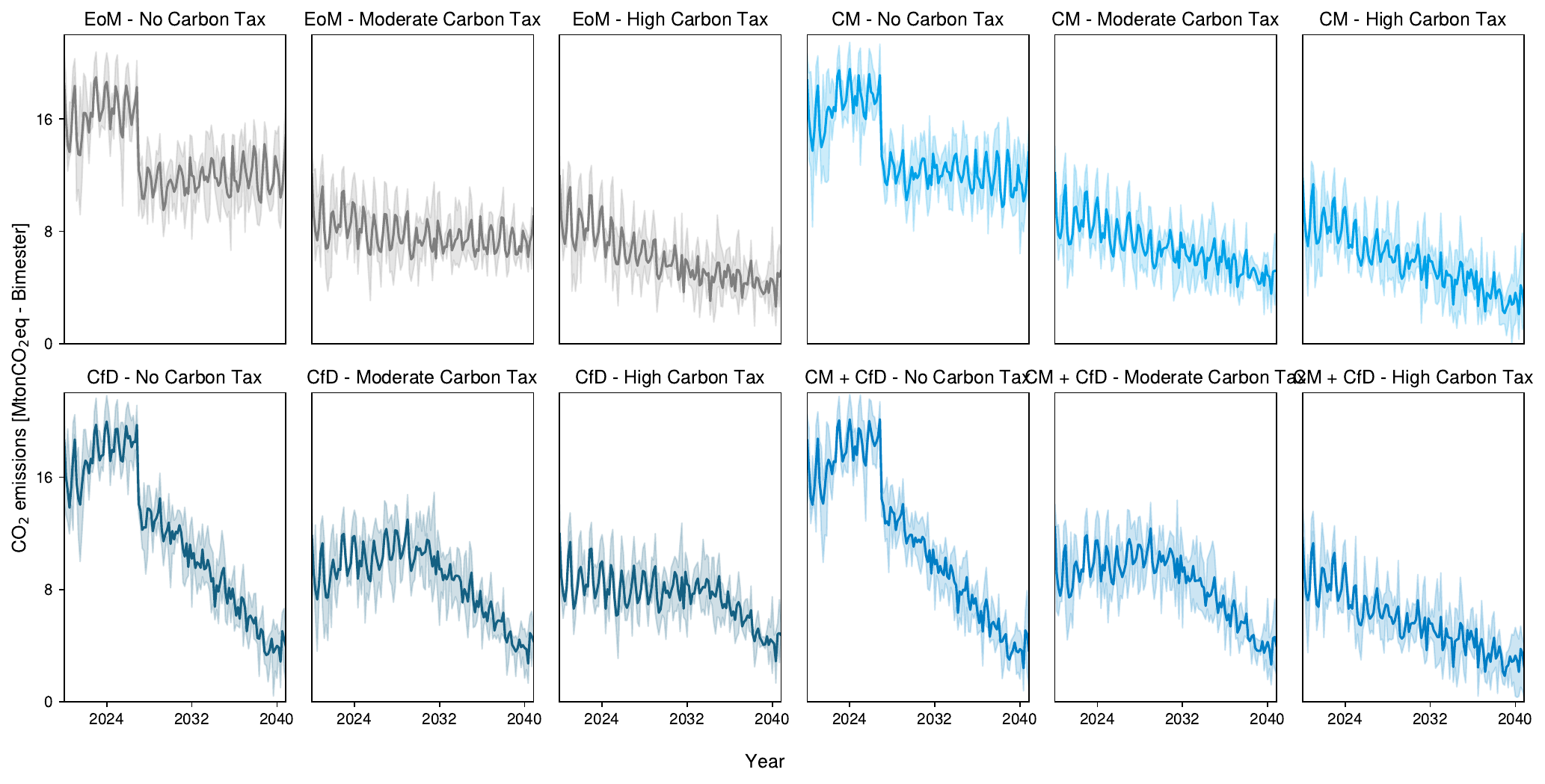} %
    \caption{Bi-monthly greenhouse gas in simulations considering 16 agents and different carbon tax scenarios. Emissions are obtained using the electricity generated per technology, and its corresponding emission factor. In the plot, the upper left, upper right, lower left, and lower right panels show results for the EoM, CM, CfD, and CM + CfD market designs. The outcome of applying the three carbon tax scenarios is displayed among these groups. Solid lines display average values, while shaded areas are the 25th and 75th percentiles. Results are obtained across 200 market simulations using the trained agents.}
\label{Appendix - fig: Emissions no tax.}
\end{figure}

\begin{figure}[hbt!]
    \centering
    \includegraphics[width=0.7\linewidth]{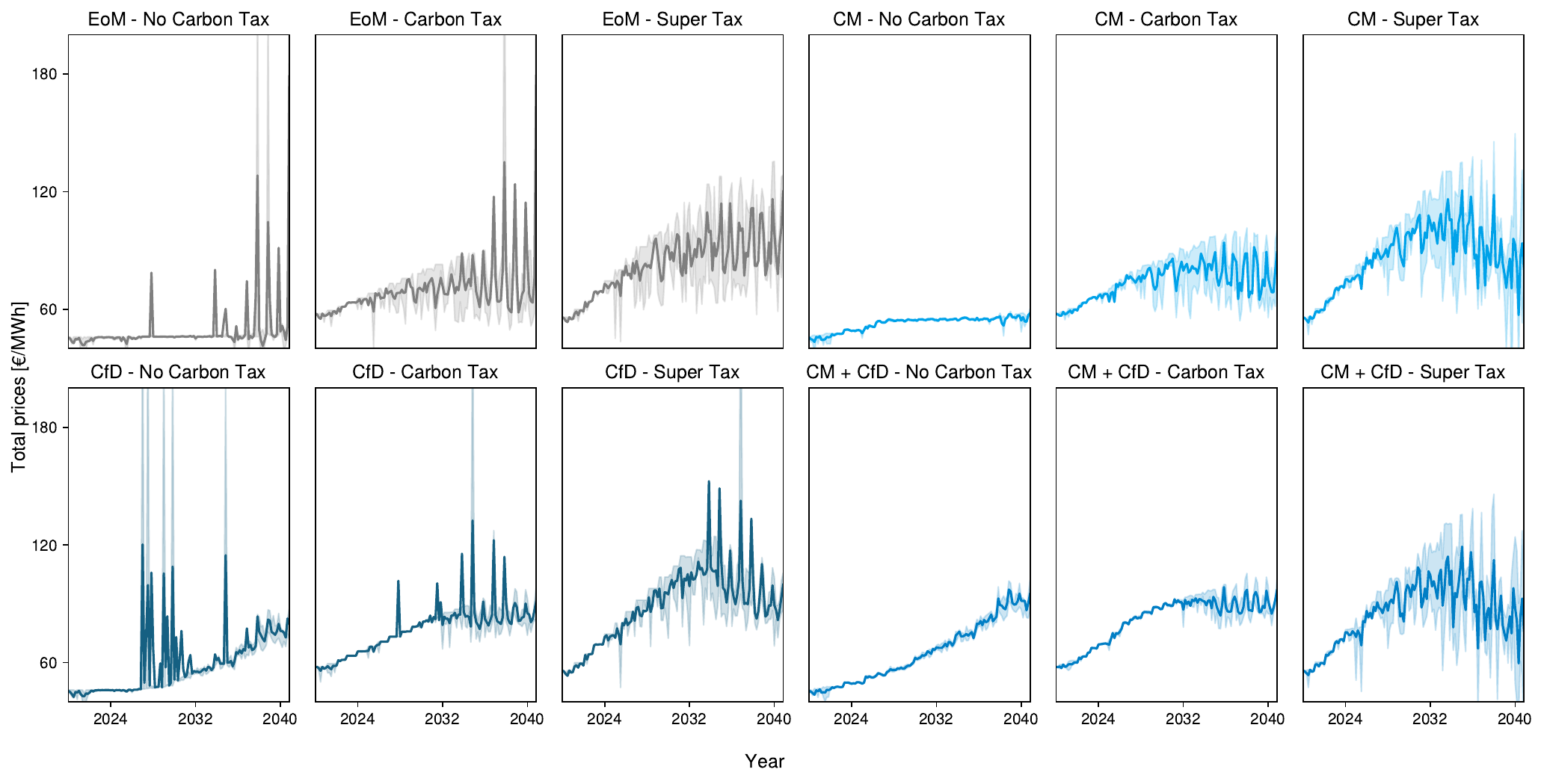} %
    \caption{Total system prices in simulations considering 16 agents and different carbon tax scenarios. Prices are calculated as the net price incurred by demand, including the short-term prices, the premiums from the capacity and contract for difference markets, and any financial settlement derived from these mechanisms. In the plot, the upper left, upper right, lower left, and lower right panels show results for the EoM, CM, CfD, and CM + CfD market designs. The outcome of applying the three carbon tax scenarios is displayed among these groups. Solid lines display average values, while shaded areas are the 25th and 75th percentiles. Results are obtained across 200 market simulations using the trained agents.}
\label{Appendix - fig: Prices no tax.}
\end{figure}

Considering the three carbon tax scenarios tested, Figures \ref{Appendix - fig: Emissions no tax.} and \ref{Appendix - fig: Prices no tax.} present the evolution of $CO_2$ emissions and total system prices across market designs. Regarding emissions, the scenarios with the highest carbon tax applied lead to a consistent decarbonization trend across markets. This is especially relevant in the EoM scenario, where system emissions remain relatively flat without applying a carbon tax.  Yet, the boost to low-carbon technologies emerges from the upward pressure on system prices across all market designs. 

Next, Tables \ref{Appendix - Table: IRR Techs.} and \ref{Appendix - Table: Capacity IRR sensitivities.} show the comparison of Internal Rate of Return (IRR) calculations when the sensitivity to the maximum investment capacity of Solar PV and OCGT technologies is applied to the environment. As stated in the main document, the IRR from individual technologies obtained by agents after training is consistently above the parametric discount rate. This result is consistent with the training patterns observed in Figure \ref{Appendix - fig: Training - electricity markets.}, and the market design. When the maximum investment sensitivity is studied, the variation in final installed capacities, particularly for Solar PV, does not significantly alter the IRR from agents, demonstrating a degree of substitutability across technologies without significant variations in the agents' rewards.
\begin{table}[hbt!]
\centering
\resizebox{\textwidth}{!}{%
\begin{tabular}{|c|c|cccccccccccc|}
\hline
\multirow{3}{*}{\textbf{Entry Market Mechanism}} &
  \multirow{3}{*}{\textbf{Technologies}} &
  \multicolumn{12}{c|}{\textbf{Number of Agents and Market design}} \\ \cline{3-14} 
 &
   &
  \multicolumn{4}{c|}{\textbf{8}} &
  \multicolumn{4}{c|}{\textbf{16}} &
  \multicolumn{4}{c|}{\textbf{32}} \\ \cline{3-14} 
 &
   &
  \multicolumn{1}{c|}{\textbf{EoM}} &
  \multicolumn{1}{c|}{\textbf{CfD}} &
  \multicolumn{1}{c|}{\textbf{CM}} &
  \multicolumn{1}{c|}{\textbf{CM + CfD}} &
  \multicolumn{1}{c|}{\textbf{EoM}} &
  \multicolumn{1}{c|}{\textbf{CfD}} &
  \multicolumn{1}{c|}{\textbf{CM}} &
  \multicolumn{1}{c|}{\textbf{CM + CfD}} &
  \multicolumn{1}{c|}{\textbf{EoM}} &
  \multicolumn{1}{c|}{\textbf{CfD}} &
  \multicolumn{1}{c|}{\textbf{CM}} &
  \textbf{CM + CfD} \\ \hline
\multirow{7}{*}{\textbf{Merchant}} &
  \textbf{Solar PV} &
  14.59 &
  12.38 &
  11.91 &
  \multicolumn{1}{c|}{12.55} &
  11.16 &
  11.86 &
  10.69 &
  \multicolumn{1}{c|}{11.80} &
  10.87 &
  11.54 &
  11.41 &
  11.58 \\ \cline{2-2}
 &
  \textbf{Onshore Wind} &
  12.62 &
  0.04 &
  3.41 &
  \multicolumn{1}{c|}{-} &
  5.76 &
  0.19 &
  2.75 &
  \multicolumn{1}{c|}{2.23} &
  5.42 &
  1.36 &
  3.38 &
  1.13 \\ \cline{2-2}
 &
  \textbf{Offshore Wind} &
  12.92 &
  9.37 &
  10.59 &
  \multicolumn{1}{c|}{3.13} &
  10.58 &
  8.84 &
  9.17 &
  \multicolumn{1}{c|}{8.83} &
  10.23 &
  8.76 &
  9.22 &
  8.58 \\ \cline{2-2}
 &
  \textbf{Coal} &
  - &
  - &
  - &
  \multicolumn{1}{c|}{-} &
  - &
  - &
  - &
  \multicolumn{1}{c|}{-} &
  - &
  - &
  - &
  - \\ \cline{2-2}
 &
  \textbf{OCGT} &
  23.44 &
  - &
  - &
  \multicolumn{1}{c|}{-} &
  13.49 &
  - &
  - &
  \multicolumn{1}{c|}{-} &
  - &
  - &
  - &
  - \\ \cline{2-2}
 &
  \textbf{CCGT} &
  - &
  - &
  - &
  \multicolumn{1}{c|}{-} &
  - &
  - &
  - &
  \multicolumn{1}{c|}{-} &
  - &
  - &
  - &
  - \\ \cline{2-2}
 &
  \textbf{Short-term Storage} &
  4.20 &
  9.56 &
  - &
  \multicolumn{1}{c|}{-} &
  6.58 &
  8.53 &
  - &
  \multicolumn{1}{c|}{2.17} &
  4.08 &
  6.59 &
  - &
  1.23 \\ \hline
\multirow{7}{*}{\textbf{CfD}} &
  \textbf{Solar PV} &
  - &
  24.94 &
  - &
  \multicolumn{1}{c|}{20.65} &
  - &
  24.74 &
  - &
  \multicolumn{1}{c|}{22.70} &
  - &
  24.31 &
  - &
  22.20 \\ \cline{2-2}
 &
  \textbf{Onshore Wind} &
  - &
  14.78 &
  - &
  \multicolumn{1}{c|}{13.18} &
  - &
  14.40 &
  - &
  \multicolumn{1}{c|}{13.25} &
  - &
  13.37 &
  - &
  12.94 \\ \cline{2-2}
 &
  \textbf{Offshore Wind} &
  - &
  14.92 &
  - &
  \multicolumn{1}{c|}{13.67} &
  - &
  14.54 &
  - &
  \multicolumn{1}{c|}{14.25} &
  - &
  14.19 &
  - &
  14.35 \\ \cline{2-2}
 &
  \textbf{Coal} &
  - &
  - &
  - &
  \multicolumn{1}{c|}{-} &
  - &
  - &
  - &
  \multicolumn{1}{c|}{-} &
  - &
  - &
  - &
  - \\ \cline{2-2}
 &
  \textbf{OCGT} &
  - &
  - &
  - &
  \multicolumn{1}{c|}{-} &
  - &
  - &
  - &
  \multicolumn{1}{c|}{-} &
  - &
  - &
  - &
  - \\ \cline{2-2}
 &
  \textbf{CCGT} &
  - &
  - &
  - &
  \multicolumn{1}{c|}{-} &
  - &
  - &
  - &
  \multicolumn{1}{c|}{-} &
  - &
  - &
  - &
  - \\ \cline{2-2}
 &
  \textbf{Short-term Storage} &
  - &
  - &
  - &
  \multicolumn{1}{c|}{-} &
  - &
  - &
  - &
  \multicolumn{1}{c|}{-} &
  - &
  - &
  - &
  - \\ \hline
\multirow{7}{*}{\textbf{CM}} &
  \textbf{Solar PV} &
  - &
  - &
  13.03 &
  \multicolumn{1}{c|}{13.46} &
  - &
  - &
  10.11 &
  \multicolumn{1}{c|}{6.38} &
  - &
  - &
  10.76 &
  7.12 \\ \cline{2-2}
 &
  \textbf{Onshore Wind} &
  - &
  - &
  9.32 &
  \multicolumn{1}{c|}{11.41} &
  - &
  - &
  5.90 &
  \multicolumn{1}{c|}{6.31} &
  - &
  - &
  9.11 &
  7.32 \\ \cline{2-2}
 &
  \textbf{Offshore Wind} &
  - &
  - &
  10.79 &
  \multicolumn{1}{c|}{14.69} &
  - &
  - &
  10.28 &
  \multicolumn{1}{c|}{11.13} &
  - &
  - &
  10.90 &
  11.04 \\ \cline{2-2}
 &
  \textbf{Coal} &
  - &
  - &
  1.38 &
  \multicolumn{1}{c|}{-} &
  - &
  - &
  - &
  \multicolumn{1}{c|}{-} &
  - &
  - &
  - &
  - \\ \cline{2-2}
 &
  \textbf{OCGT} &
  - &
  - &
  13.08 &
  \multicolumn{1}{c|}{19.69} &
  - &
  - &
  13.78 &
  \multicolumn{1}{c|}{16.54} &
  - &
  - &
  14.80 &
  16.05 \\ \cline{2-2}
 &
  \textbf{CCGT} &
  - &
  - &
  9.46 &
  \multicolumn{1}{c|}{16.24} &
  - &
  - &
  6.57 &
  \multicolumn{1}{c|}{8.63} &
  - &
  - &
  9.32 &
  9.23 \\ \cline{2-2}
 &
  \textbf{Short-term Storage} &
  - &
  - &
  11.79 &
  \multicolumn{1}{c|}{19.22} &
  - &
  - &
  12.00 &
  \multicolumn{1}{c|}{14.45} &
  - &
  - &
  11.53 &
  13.95 \\ \hline
\end{tabular}%
}
\caption{Internal Rate of Return of investments for generation and storage technologies under different market designs and competition levels. The Internal Rate of Return is calculated as an aggregate for all investments in a particular technology by all agents in the simulation. Values are indicated for cases where the Internal Rate of Return can be computed \textit{(Investments in a particular technology are placed, and the resulting IRR is positive)}. Otherwise, a hyphen is reported instead. }
\label{Appendix - Table: IRR Techs.}
\end{table}

\begin{table}[hbt!]
\centering
\resizebox{\textwidth}{!}{%
\begin{tabular}{|c|l|cc|cc|cc|}
\hline
\multirow{2}{*}{\textbf{Entry Market Mechanism}} &
  \multicolumn{1}{c|}{\multirow{2}{*}{\textbf{Technologies}}} &
  \multicolumn{2}{c|}{\textbf{\begin{tabular}[c]{@{}c@{}}Installed Capacity 2040\\ $[MW]$\end{tabular}}} &
  \multicolumn{2}{c|}{\textbf{\begin{tabular}[c]{@{}c@{}}Technology IRR\\ $[\%]$\end{tabular}}} &
  \multicolumn{2}{c|}{\textbf{\begin{tabular}[c]{@{}c@{}}Agents IRR\\ $[\%]$\end{tabular}}} \\ \cline{3-8} 
 &
  \multicolumn{1}{c|}{} &
  \multicolumn{1}{c|}{\textbf{Base Case}} &
  \textbf{SMI} &
  \multicolumn{1}{c|}{\textbf{Base Case}} &
  \textbf{SMI} &
  \multicolumn{1}{c|}{\textbf{Base Case}} &
  \textbf{SMI} \\ \hline
\multirow{7}{*}{\textbf{Merchant}} &
  \textbf{Solar PV} &
  5981.7 &
  34213.3 &
  12.55 &
  9.79 &
  \multicolumn{1}{c|}{\multirow{21}{*}{\textbf{16.52}}} &
  \multirow{21}{*}{\textbf{13.96}} \\ \cline{2-6}
                              & \textbf{Onshore Wind}       & 52.5    & 8.1      & --     & 0.18  & \multicolumn{1}{c|}{} &  \\ \cline{2-6}
                              & \textbf{Offshore Wind}      & 1.4     & 472.5    & 3.13  & 8.41  & \multicolumn{1}{c|}{} &  \\ \cline{2-6}
                              & \textbf{Coal}               & 0       & 0        & --     & --     & \multicolumn{1}{c|}{} &  \\ \cline{2-6}
                              & \textbf{OCGT}               & 118.3   & 6.7      & --     & --     & \multicolumn{1}{c|}{} &  \\ \cline{2-6}
                              & \textbf{CCGT}               & 55.6    & 5.6      & --     & --     & \multicolumn{1}{c|}{} &  \\ \cline{2-6}
                              & \textbf{Short-term Storage} & 538.3   & 712.3    & --     & 5.26  & \multicolumn{1}{c|}{} &  \\ \cline{1-6}
\multirow{7}{*}{\textbf{CfD}} & \textbf{Solar PV}           & 83853.4 & 131269.8 & 20.65 & 21.00 & \multicolumn{1}{c|}{} &  \\ \cline{2-6}
                              & \textbf{Onshore Wind}       & 5835.1  & 1795.0   & 13.18 & 9.78  & \multicolumn{1}{c|}{} &  \\ \cline{2-6}
                              & \textbf{Offshore Wind}      & 28870.4 & 12418.1  & 13.67 & 13.33 & \multicolumn{1}{c|}{} &  \\ \cline{2-6}
                              & \textbf{Coal}               & 0       & 0        & --     & --     & \multicolumn{1}{c|}{} &  \\ \cline{2-6}
                              & \textbf{OCGT}               & 0       & 0        & --     & --     & \multicolumn{1}{c|}{} &  \\ \cline{2-6}
                              & \textbf{CCGT}               & 0       & 0        & --     & --     & \multicolumn{1}{c|}{} &  \\ \cline{2-6}
                              & \textbf{Short-term Storage} & 0       & 0        & --     & --     & \multicolumn{1}{c|}{} &  \\ \cline{1-6}
\multirow{7}{*}{\textbf{CM}}  & \textbf{Solar PV}           & 4301.8  & 0        & 13.46 & --     & \multicolumn{1}{c|}{} &  \\ \cline{2-6}
                              & \textbf{Onshore Wind}       & 5786.6  & 627.1    & 11.41 & 2.88  & \multicolumn{1}{c|}{} &  \\ \cline{2-6}
                              & \textbf{Offshore Wind}      & 5751.2  & 5467.0   & 14.69 & 9.82  & \multicolumn{1}{c|}{} &  \\ \cline{2-6}
                              & \textbf{Coal}               & 214.8   & 53.3     & --     & --     & \multicolumn{1}{c|}{} &  \\ \cline{2-6}
                              & \textbf{OCGT}               & 32786.0 & 44538.7  & 19.69 & 12.17 & \multicolumn{1}{c|}{} &  \\ \cline{2-6}
                              & \textbf{CCGT}               & 7874.2  & 964.8    & 16.24 & 9.64  & \multicolumn{1}{c|}{} &  \\ \cline{2-6}
                              & \textbf{Short-term Storage} & 3129.5  & 4624.6   & 19.22 & 19.58 & \multicolumn{1}{c|}{} &  \\ \hline
\end{tabular}%
}
\caption{Installed Capacities per technology in 2040, Internal Rate of Return of investments for generation and storage technologies, and aggregated Internal Rate of Return in the market, comparing the Base scenario and the sensitivity implemented by doubling the maximum installed capacity parameters for Solar PV and OCGT, under the CM+CfD market design and 8 agents. The sensitivity results are indicated in columns denoted as SMI. The Internal Rate of Return is calculated as an aggregate for all investments in a particular technology by all agents in the simulation. Values are indicated for cases where the Internal Rate of Return can be computed \textit{(Investments in a particular technology are placed, and the resulting IRR is positive)}. Otherwise, a hyphen is reported instead. }
\label{Appendix - Table: Capacity IRR sensitivities.}
\end{table}

\begin{figure}[hbt!]
    \centering
    \includegraphics[width=0.7\linewidth]{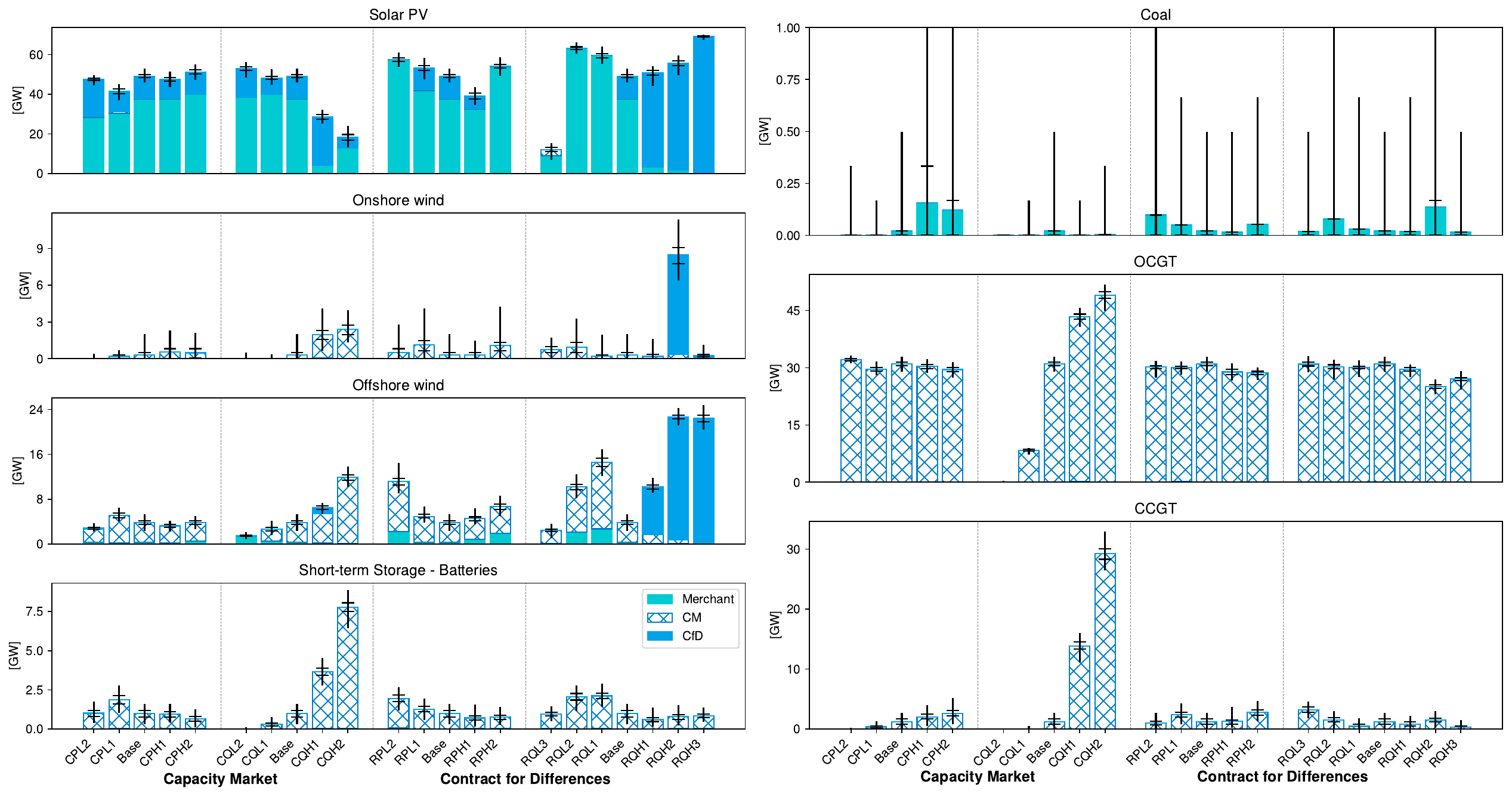} %
    \caption{Installed Capacity of generation and storage technologies in the year 2030 under sensitivities for the long-term electricity market design. Stacked bars indicate average installed capacities across runs. In each stacked bar, the horizontal markers display the 25th and 75th percentiles, and the vertical markers the minimum and maximum values. Results are grouped in column categories and divided by vertical dotted gray lines according to the sensitivity type and the market under consideration. Moreover, hatching highlights the mechanism used by agents to enter the market. Results from the MARL model are obtained across 200 market simulations using the trained agents.}
\label{Appendix - fig: Installed Capacity 2030 market sensitivities.}
\end{figure}

\begin{figure}[hbt!]
    \centering
    \includegraphics[width=0.7\linewidth]{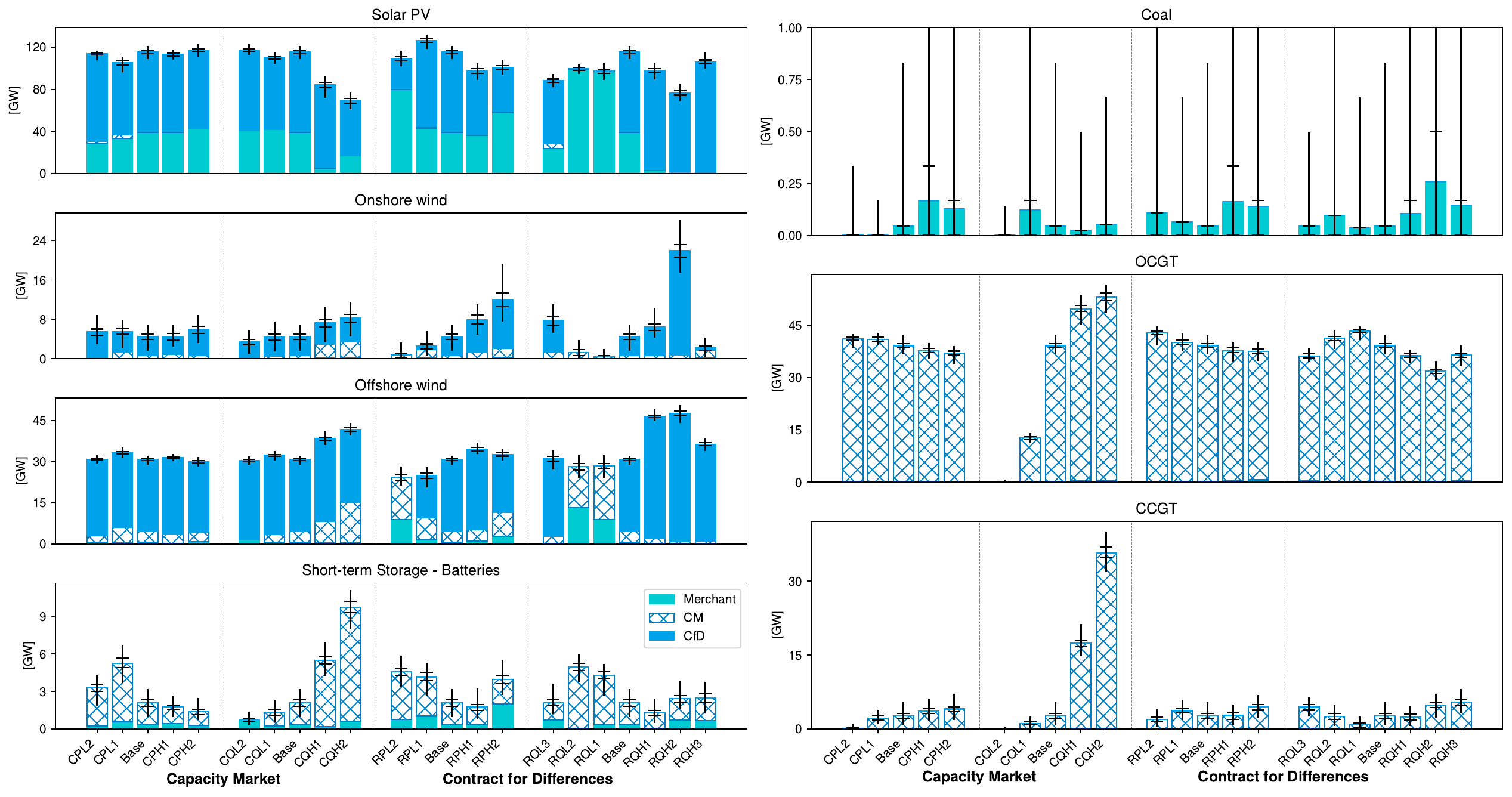} %
    \caption{Installed Capacity of generation and storage technologies in the year 2040 under sensitivities for the long-term electricity market design. Stacked bars indicate average installed capacities across runs. In each stacked bar, the horizontal markers display the 25th and 75th percentiles, and the vertical markers the minimum and maximum values. Results are grouped in column categories and divided by vertical dotted gray lines according to the sensitivity type and the market under consideration. Moreover, hatching highlights the mechanism used by agents to enter the market. Results from the MARL model are obtained across 200 market simulations using the trained agents.}
\label{Appendix - fig: Installed Capacity 2040 market sensitivities.}
\end{figure}

\begin{figure}[hbt!]
    \centering
    \includegraphics[width=0.7\linewidth]{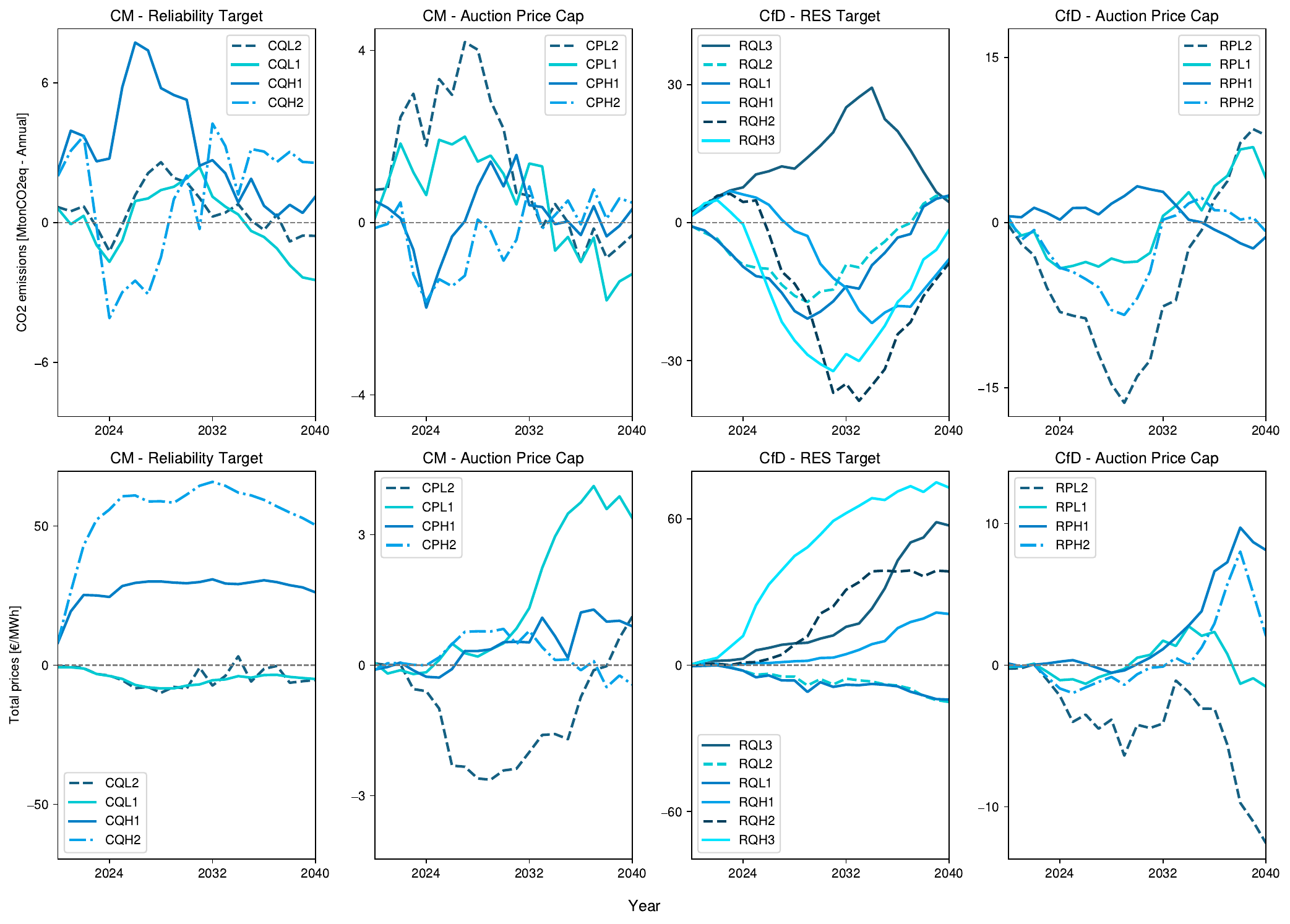} %
    \caption{ Differences between market sensitivities and base case of annual greenhouse gas and total electricity prices. Emissions are obtained using the electricity generated per technology, and its corresponding emission factor. Prices are calculated as the net price incurred by demand, including the short-term prices, the premiums from the capacity and contract for difference markets, and any financial settlement derived from these mechanisms. Columns are organized depending on the sensitivity type and the market under consideration. Solid lines display average values. Results are obtained across 200 market simulations using the trained agents.}
\label{Appendix - fig: Price and emissions market sensitivities.}
\end{figure}

For the market and discount rate sensitivities instead, Table \ref{Appendix - Table: Market sensitivities.} presents the complete testing configuration. Importantly, all of these sensitivities are applied in the CM + CfD environment, using a 16-agent configuration. After applying the sensitivities, the resulting installed capacities for 2030 and 2040 are shown in Figures  \ref{Appendix - fig: Installed Capacity 2030 market sensitivities.} and \ref{Appendix - fig: Installed Capacity 2040 market sensitivities.}, respectively, while the price and emissions differences of the sensitivity with respect to the base scenario are introduced in Figure \ref{Appendix - fig: Price and emissions market sensitivities.}. 

\begin{table}[hbt!]
\centering
\resizebox{\textwidth}{!}{%
\begin{tabular}{|c|c|c|c|c|c|}
\hline
\textbf{\begin{tabular}[c]{@{}c@{}}Market under \\ sensitivity\end{tabular}} &
  \textbf{Parameter} &
  \textbf{Base Value} &
  \textbf{Sensitivity} &
  \textbf{ID} &
  \textbf{Description} \\ \hline
\multirow{8}{*}{\textbf{\begin{tabular}[c]{@{}c@{}}Capacity \\ Market\\ (CM)\end{tabular}}} &
  \multirow{4}{*}{\begin{tabular}[c]{@{}c@{}}Auction price \\ cap\end{tabular}} &
  \multirow{4}{*}{\begin{tabular}[c]{@{}c@{}}40 \\ {[}\texteuro/MWh-firm{]}\end{tabular}} &
  25\% &
  CPL2 &
  \multirow{4}{*}{\begin{tabular}[c]{@{}c@{}}Sensitivity to the  price cap in the capacity market\\  auctions. The PL1 scenario corresponds \\ to a price cap equivalent to 25\% \\ of the value used in the base case.\end{tabular}} \\ \cline{4-5}
 &  &  & 50\%                                                                                            & CPL1 &  \\ \cline{4-5}
 &  &  & 150\%                                                                                           & CPH1 &  \\ \cline{4-5}
 &  &  & 200\%                                                                                           & CPH2 &  \\ \cline{2-6} 
 &
  \multirow{4}{*}{Demand margin} &
  \multirow{4}{*}{0} &
  (-) 66\% &
  CQL2 &
  \multirow{4}{*}{\begin{tabular}[c]{@{}c@{}}Sensitivity to the margin over the maximum demand, \\ which is used to balance adequacy in the capacity market.\\ The QL1 scenario corresponds to a 33\% reduction \\ in the maximum demand for adequacy estimations\end{tabular}} \\ \cline{4-5}
 &  &  & (-) 33\%                                                                                        & CQL1 &  \\ \cline{4-5}
 &  &  & (+) 33\%                                                                                        & CQH1 &  \\ \cline{4-5}
 &  &  & (+) 66\%                                                                                        & CQH2 &  \\ \hline
\multirow{10}{*}{\textbf{\begin{tabular}[c]{@{}c@{}} \\ \\ \\ \\ Contract for \\ Difference market\\ (CfD)\end{tabular}}} &
  \multirow{4}{*}{\begin{tabular}[c]{@{}c@{}}Auction price \\ cap\end{tabular}} &
  \multirow{4}{*}{\begin{tabular}[c]{@{}c@{}}200 \\ {[}\texteuro/MWh{]}\end{tabular}} &
  25\% &
  RPL2 &
  \multirow{4}{*}{\begin{tabular}[c]{@{}c@{}}Sensitivity to the price cap in the contract for \\ difference auctions. The PL1 scenario\\  corresponds to a price cap equivalent to 25\% \\ of the value used in the base case.\end{tabular}} \\ \cline{4-5}
 &  &  & 50\%                                                                                            & RPL1 &  \\ \cline{4-5}
 &  &  & 150\%                                                                                           & RPH1 &  \\ \cline{4-5}
 &  &  & 200\%                                                                                           & RPH2 &  \\ \cline{2-6} 
 &
  \multirow{6}{*}{\begin{tabular}[c]{@{}c@{}} \\ \\ \\ \\ RES\\ target\end{tabular}} &
  \multirow{6}{*}{\begin{tabular}[c]{@{}c@{}} \\ \\ \\ 2025 = 21\%\\ 2030 = 40\%\\ 2035 = 65\%\\ 2040 = 90\%\end{tabular}} &
  \begin{tabular}[c]{@{}c@{}}2025 = 6\%, 2030 = 6\%, \\ 2035 = 42\%, 2040 = 90\%\end{tabular} &
  RQL3 &
  \multirow{6}{*}{\begin{tabular}[c]{@{}c@{}} \\ \\ \\ Sensitivity to the Renewable Energy Target. \\ The QL1 scenario indicates a slight reduction \\ in the RES target used during the base case, \\ showcased in Figure \ref{fig: Existing conditions.}.\end{tabular}} \\ \cline{4-5}
 &  &  & \begin{tabular}[c]{@{}c@{}}2025 = 13\%, 2030 = 23\%, \\ 2035 = 35\%, 2040 = 48\%\end{tabular}   & RQL2 &  \\ \cline{4-5}
 &  &  & \begin{tabular}[c]{@{}c@{}}2025 = 17\%, 2030 = 28\%, \\ 2035 = 47\%, 2040 = 66\%\end{tabular}  & RQL1 &  \\ \cline{4-5}
 &  &  & \begin{tabular}[c]{@{}c@{}}2025 = 28\%, 2030 = 51\%, \\ 2035 = 88\%, 2040 = 100\%\end{tabular} & RQH1 &  \\ \cline{4-5}
 &  &  & \begin{tabular}[c]{@{}c@{}}2025 = 36\%, 2030 = 74\%, \\ 2035 = 100\%, 2040 = 100\%\end{tabular} & RQH2 &  \\ \cline{4-5}
 &  &  & \begin{tabular}[c]{@{}c@{}}2025 = 42\%, 2030 = 90\%, \\ 2035 = 90\%, 2040 = 90\%\end{tabular} & RQH3 &  \\ \hline
\multirow{6}{*}{\textbf{Discount Rate}} &
  \multirow{6}{*}{\begin{tabular}[c]{@{}c@{}}Annual\\ Discount Rate\end{tabular}} &
  \multirow{6}{*}{8\%} &
  4\% &
  DL2 &
  \multirow{6}{*}{\begin{tabular}[c]{@{}c@{}}Sensitivity to the agents' discount rate. \\ The DL1 scenario corresponds \\ to a slight reduction \\ in the discount rate used during the base case.\end{tabular}} \\ \cline{4-5}
 &  &  & 6\%                                                                                             & DL1 &  \\ \cline{4-5}
 &  &  & 10\%                                                                                            & DH1 &  \\ \cline{4-5}
 &  &  & 12\%                                                                                            & DH2 &  \\ \cline{4-5}
 &  &  & 14\%                                                                                            & DH3 &  \\ \cline{4-5}
 &  &  & 16\%                                                                                            & DH4 &  \\ \hline
\end{tabular}%
}
\caption{Configurations used for sensitivities on relevant parameters on long-term electricity markets. Rows indicate the type of sensitivity applied, including the market under analysis, the parameter subject to the sensitivity, the base value utilized during the main simulations, the applied parameter during the sensitivity, the identifier code, and a short description for the corresponding scenario. All sensitivities are applied in the CM + CfD environment, using a 16-agent configuration. The first letter in the scenario ID corresponds to the type of sensitivity carried out. For CM and CfD sensitivities, the second letter indicates if target quantities (Q) or price-caps in the auctions (P) are subject to variations during the tests.}
\label{Appendix - Table: Market sensitivities.}
\end{table}

For the RES target for the CfD market: Six additional trajectories for the RES target were tested. These trajectories varied between 6 and 90\% RES penetration over average demand for 2030 (40\% in the base scenario), and between 48\% and 100\% for 2040 (90\% in the base scenario). The trajectories tested the required target of RES and the dynamics the target for the 2020-2040 period. In terms of installed capacity, increasing the RES target leads to increased onshore and offshore wind penetration, while the solar PV penetration decreased slightly. Wind penetration, with a higher capacity factor and a higher contribution to the adequacy targets than solar PV, also reduces Open-Cycle Gas Turbines (OCGT) investments. Interestingly, reducing the RES target over the base scenario resulted in the market achieving the penetration goal without resorting to CfD auctions. Instead, RES enter the system via naturally occurring merchant investments and the CM auctions. Large $CO_2$ emissions reductions were observed in the scenarios with increments in the RES targets. These emission reductions were more prominent around the year 2030, highlighting the importance of a dynamic understanding of decarbonization trends compared to a narrow focus on targets for particular years. As expected, total electricity prices increased when the RES target was augmented. This effect reinforces the need for a comprehensive analysis of energy policies, considering them as complementary options that balance system requirements and objectives with the decarbonization needs. Moreover, faster uptakes in the RES target yield higher emission reduction potential, especially around the year 2030, but also increase the pressure on system prices and reduce incentives for merchant investments.

For the adequacy target in the CM market, four variations were applied to the system maximum demand, the value used to estimate the adequacy target within the Capacity market. These variations were -66\%, -33\%, +33\%, and +66\%, while no adjustment to the maximum demand was applied in the base case. In terms of installed capacity, increasing the adequacy target incentivized large deployments of OCGT, Combined Cycle Gas Turbines (CCGT), short-term storage, and, to a lesser extent, offshore wind. Interestingly, this change also harmed Solar PV, both merchant and CfD investments, which show a substantial reduction. Interestingly, only the smallest adequacy target tested led to Capacity Market auctions being avoided altogether. For $CO_2$ emissions, variations in the adequacy target had relatively neutral net effects. Two factors explain this result. The first, most emission reductions come via the natural RES investments that, as mentioned earlier, were resilient to variations in these parameters (apart from the substitution between solar PV and offshore wind). Moreover, in the model, the additional capacity promoted through the CM operates with a very low capacity factor, given the over-installation of the system. This reduces the potential impact of the added fossil-fuel fleet in terms of $CO_2$ emissions. Nonetheless, the over-installation derived from the increased adequacy targets came with a substantial increment in system prices, considerably more pronounced than in the case of the CfD auctions. These results demonstrate the need to adequately set the targets of the capacity market to avoid large-scale inefficiencies and emphasize the need for integrated policy assessment. Moreover, the results also showcase the capabilities of the MARL framework to respond to very stringent system conditions.

Concerning the price cap sensitivities, it is worth noting that for the base scenario, a selected relatively high price caps (40 {[}\texteuro/MWh-firm{]} for the capacity market and 200 {[}\texteuro/MWh{]} for the CfD auctions), allowing market dynamics to be the main drivers behind price-setting. For the sensitivities, equivalent price caps of 25\%, 50\%, 150\%, and 200\% of the base price are applied. In general, the results show smaller impacts in the price cap variations compared to the changes for the targets in the corresponding markets. This is consistent with the expected model behavior, highlighting that market competition and demand levels remain the most significant drivers behind price formation in long-term auctions. Nevertheless, the model is not invariant to changes in price caps. Particularly, increasing the price caps in the CM resulted in CCGT plants substituting some shares of OCGT investments. For the CfD case, higher auction caps led to higher onshore wind investments, substituting some OCGT in the market. Two main reasons can explain this effect. First, during the simulation, price caps may be reached. If this situation emerges on a consistent basis, prices and profits in the market become easier to predict, facilitating learning. Second, bidding for long-term auctions uses the discretization of a continuous variable. Although we follow the guidelines for discretization provided in \cite{tang_discretizing_2020}, this is still a simplification over the direct modeling of continuous variables for bidding. Nonetheless, this design option was selected for the environment, as the benefits of implementing multi-discrete actions and action masking largely outweigh the use of continuous ones. 

\begin{figure}[hbt!]
    \centering
    \includegraphics[width=0.7\linewidth]{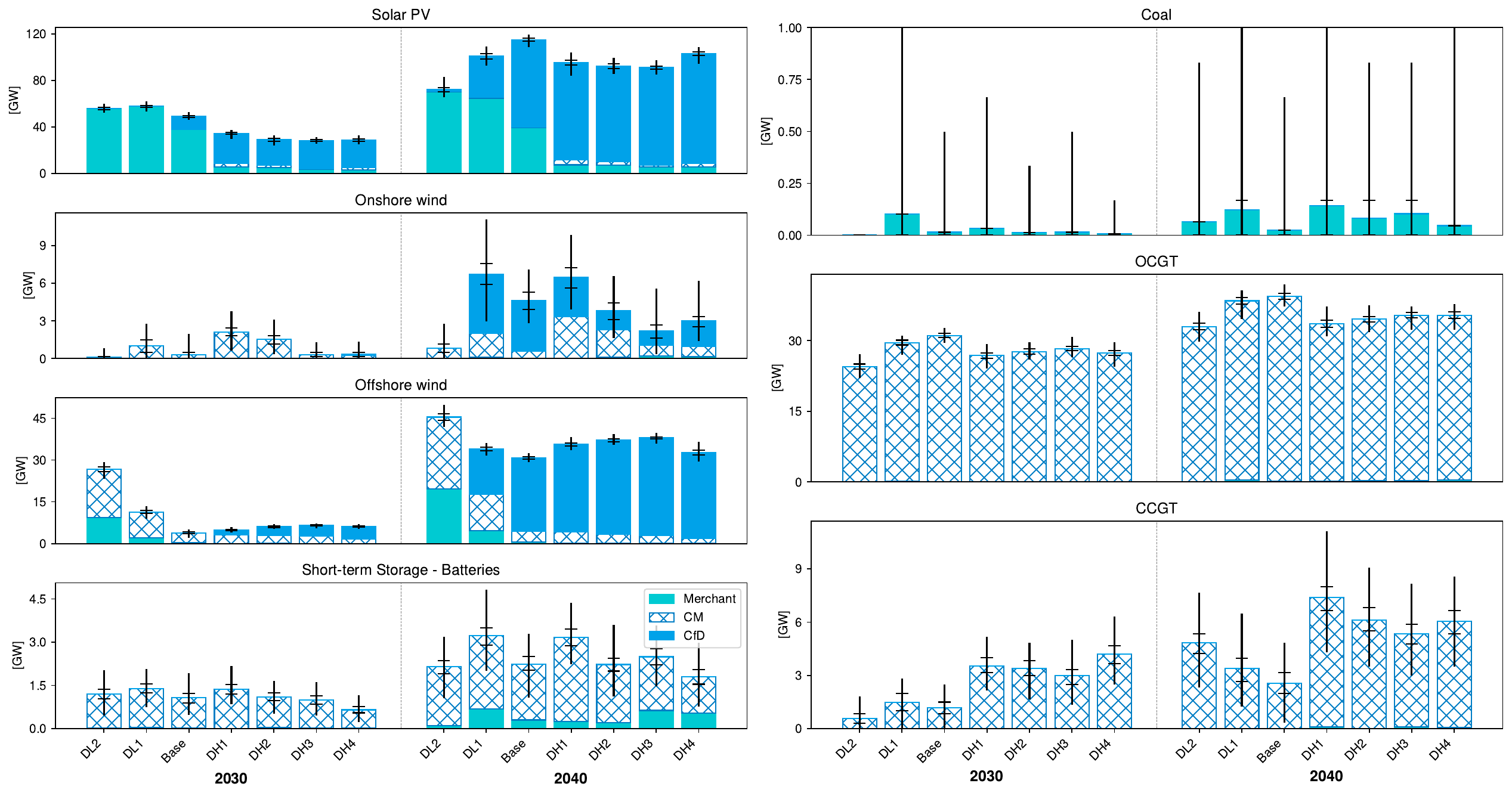} 
    \caption{ Installed Capacity of generation and storage technologies across the discount rate sensitivities in the year 2030 and 2040. Stacked bars indicate average installed capacities across runs. In each stacked bar, the horizontal markers display the 25th and 75th percentiles, and the vertical markers the minimum and maximum values. Results are grouped in column categories and divided by vertical dotted gray lines according to the investment year. Moreover, hatching highlights the mechanism used by agents to enter the market. Results from the MARL model are obtained across 200 market simulations using the trained agents.}
\label{Appendix - fig: Installed Capacity discount rate.}
\end{figure}

\begin{figure}[hbt!]
    \centering
    \includegraphics[width=0.7\linewidth]{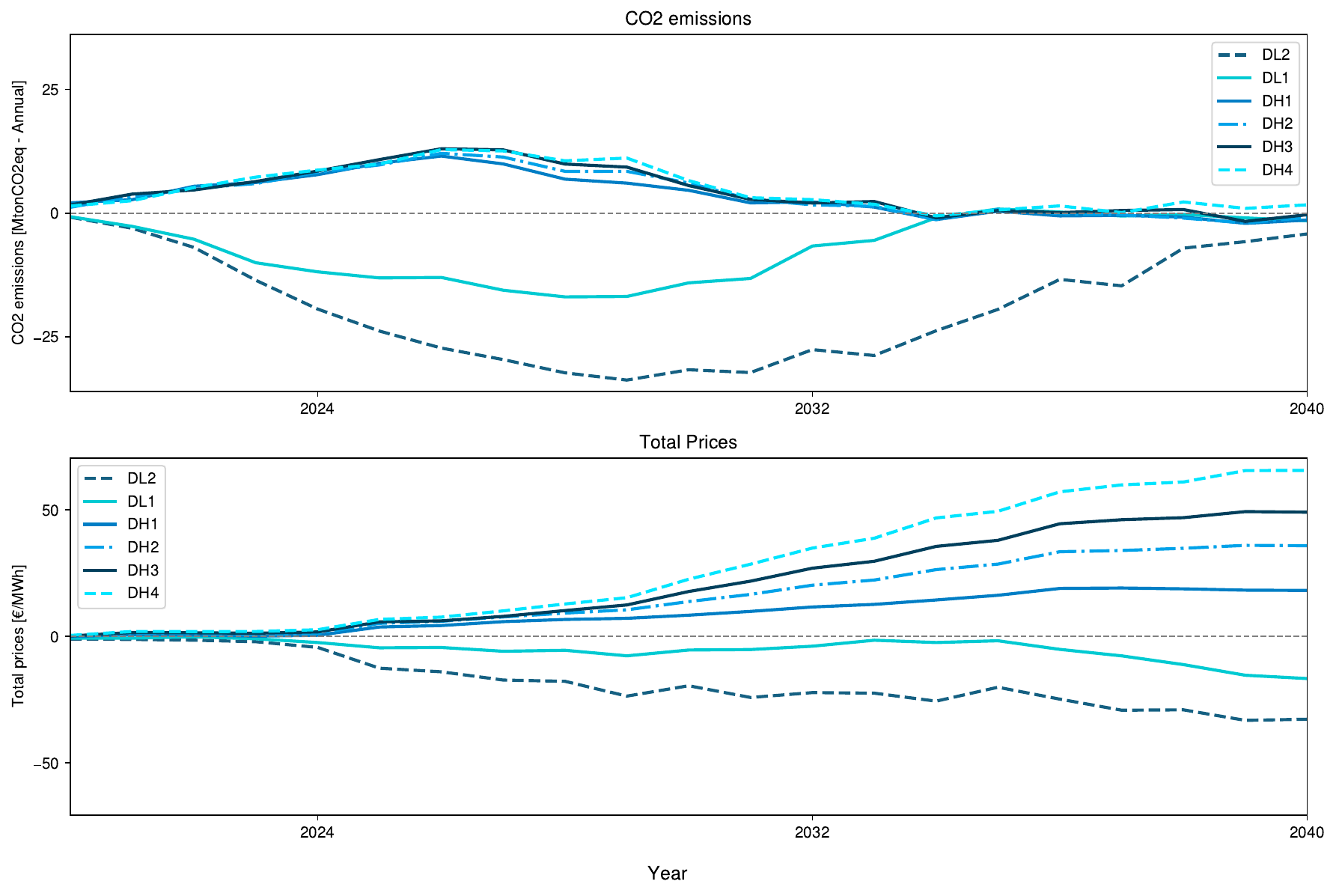} 
    \caption{Differences between discount rate sensitivities and base case of annual greenhouse gas and total electricity prices. Emissions are obtained using the electricity generated per technology, and its corresponding emission factor. Prices are calculated as the net price incurred by demand, including the short-term prices, the premiums from the capacity and contract for difference markets, and any financial settlement derived from these mechanisms. Columns are organized depending on the sensitivity type and the market under consideration. Solid lines display average values. Results are obtained across 200 market simulations using the trained agents.}
\label{Appendix - fig: Price and emissions discount rate.}
\end{figure}

For the discount rate sensitivities, results are presented in Figures \ref{Appendix - fig: Installed Capacity discount rate.} and \ref{Appendix - fig: Price and emissions discount rate.}.In this case, higher discount rates resulted in a rapid shift away from merchant investments, with the subsequent utilization of long-term auctions to meet demand requirements. However, the higher auction utilization and increased prices obtained through these mechanisms resulted in substantially higher system costs. The discount rate results are also indirectly linked to the price-cap sensitivities. In short, increasing the discount rate rapidly shifts investment away from merchant investments towards long-term auctions. Moreover, the prices in these auctions increased dramatically, compensating for the higher discount rate in the scenarios. This behavior is enabled by the relatively high price-caps in the base scenario, for both CM and CfD auctions. As a result, it is expected that in combined exercises that examine discount rate sensitivities and lower price caps, or in contrast, in exercises that actively seek to limit the increased prices emerging from high discount rate scenarios, price caps would have a greater effect on system results.

\begin{figure}[hbt!]
    \centering
    \includegraphics[width=0.7\linewidth]{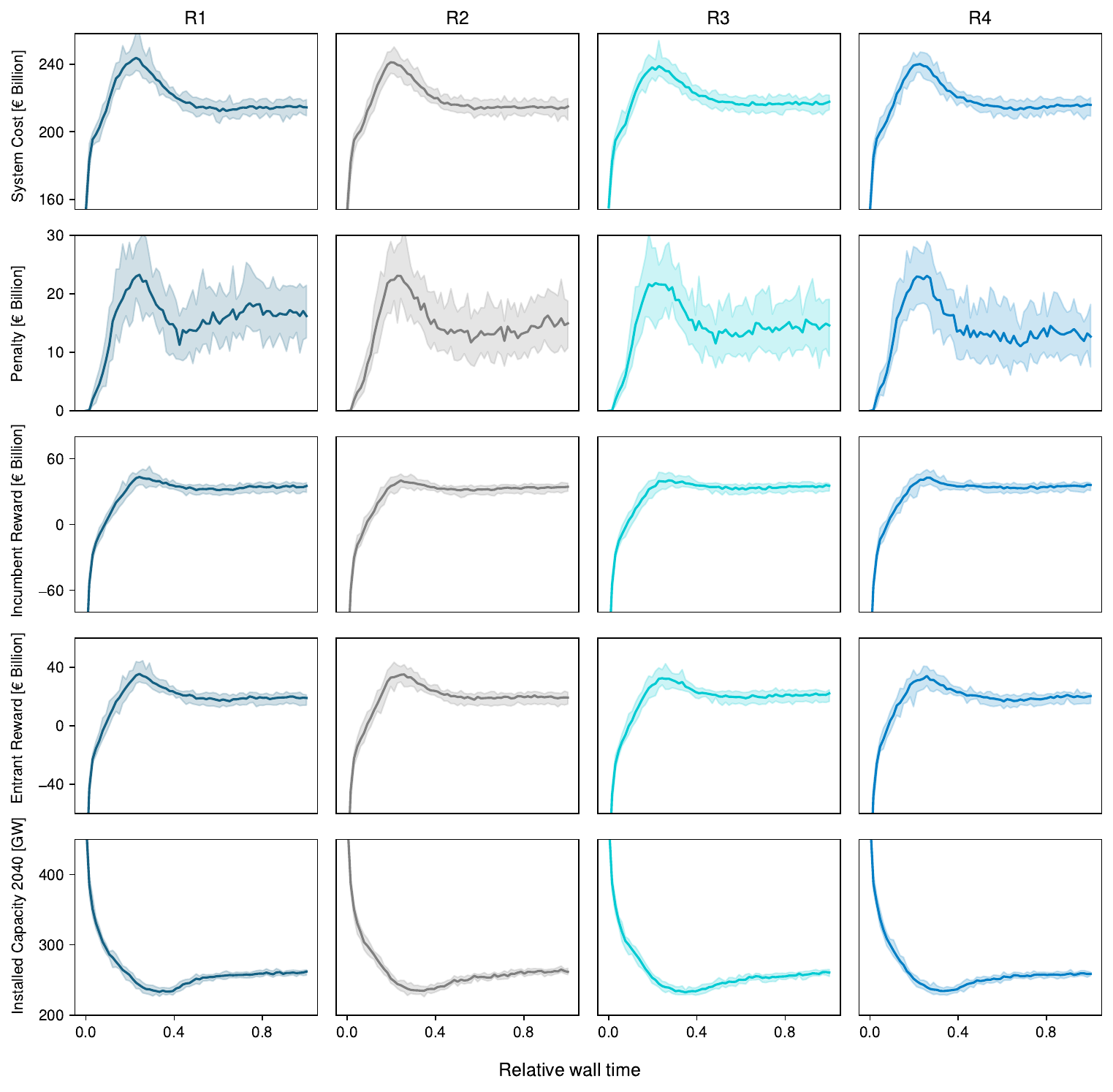} %
    \caption{Relevant Training Metrics for Different Market Configurations across four (R1-R4) independent training sessions. From top to bottom, the graph showcases the total electricity system cost, the agents' Penalty, calculated using the procedure described in Section \ref{SubSection - Hyperparameter Selection}, the aggregated reward of agents grouped in incumbent and entrants categories, and the total installed capacity in the year 2040. Solid lines display average values, while shaded areas are the minimum and maximum values during sampling. Results are obtained using intermediate checkpoints during training, and sampled from 25 market simulations.}
\label{Appendix - fig: Training - reruns.}
\end{figure}

\begin{figure}[hbt!]
    \centering
    \includegraphics[width=0.7\linewidth]{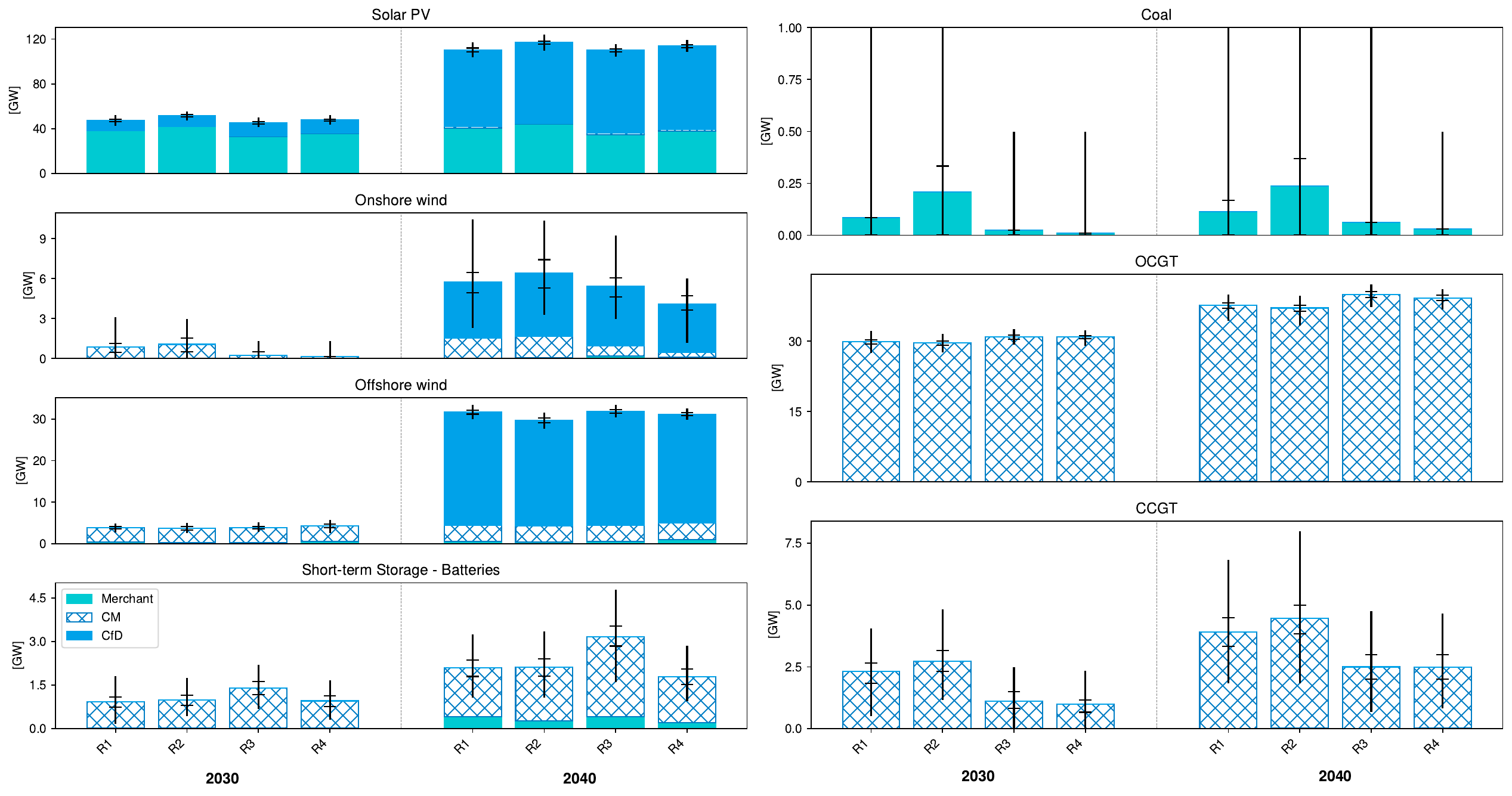} 
    \caption{Installed Capacity of generation and storage technologies in the year 2030 and 2040 across four (R1-R4) independent training sessions. Stacked bars indicate average installed capacities across runs. In each stacked bar, the horizontal markers display the 25th and 75th percentiles, and the vertical markers the minimum and maximum values. Results are grouped in column categories and divided by vertical dotted gray lines according to the investment year. Moreover, hatching highlights the mechanism used by agents to enter the market. Results from the MARL model are obtained across 200 market simulations using the trained agents.}
\label{Appendix - fig: Installed Capacity reruns.}
\end{figure}

Last, Figures \ref{Appendix - fig: Training - reruns.}-\ref{Appendix - fig: Installed Capacity reruns.} introduced the results of the robustness analysis. Along with the metrics presented, both from training and the market outcomes with the trained agents, most trends remain largely unaltered across the independent training sessions tested. Yet, noticeable variations are presented in the final installed capacities in the system, highlighting a degree of substitutability in the agents' and system portfolios.

\begin{figure}[hbt!]
    \centering
    \includegraphics[width=0.7\linewidth]{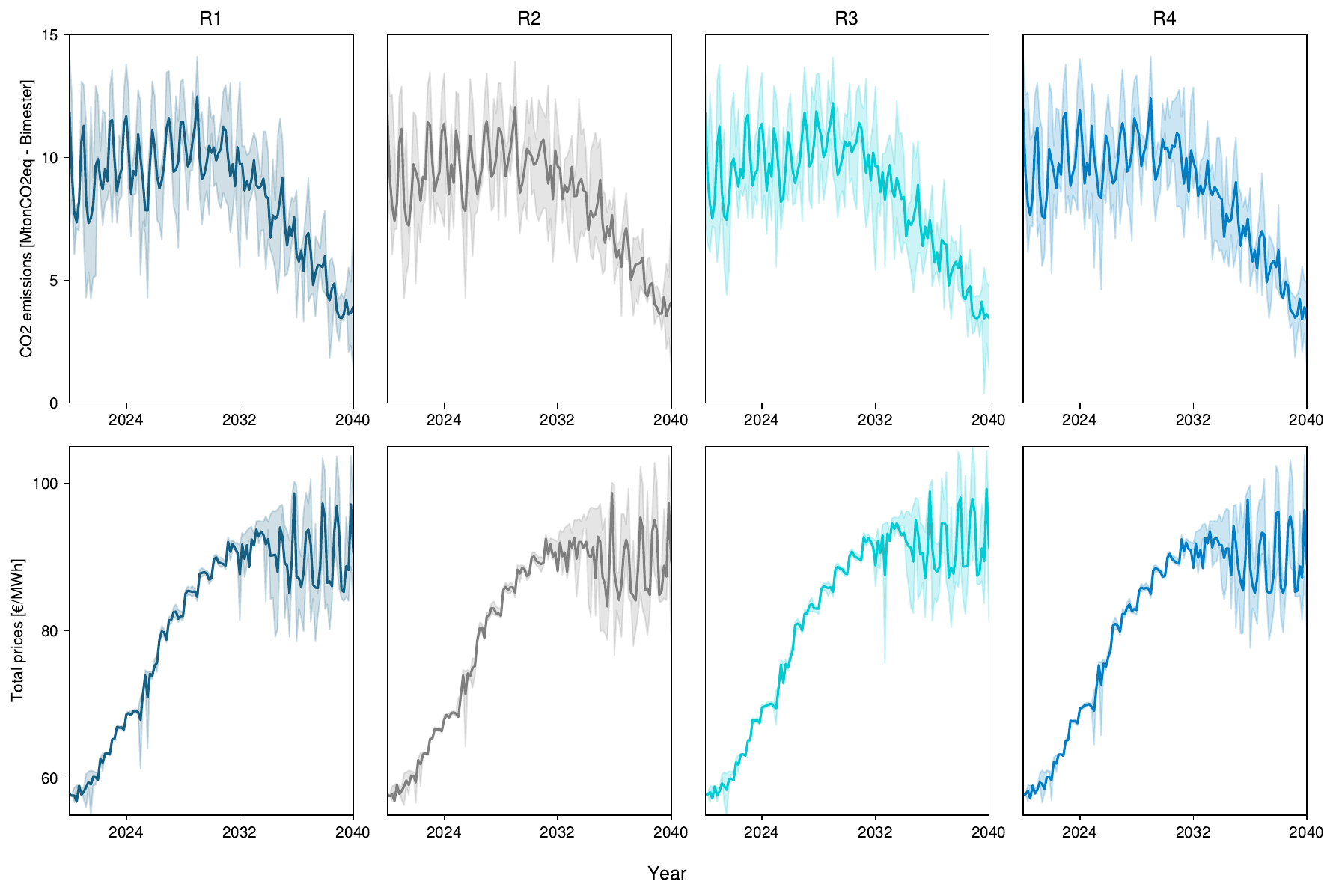} 
    \caption{Bi-monthly greenhouse gas and total electricity prices across four (R1-R4) independent training sessions. Emissions are obtained using the electricity generated per technology, and its corresponding emission factor. Prices are calculated as the net price incurred by demand, including the short-term prices, the premiums from the capacity and contract for difference markets, and any financial settlement derived from these mechanisms. Columns are organized depending on the training sessions. Solid lines display average values, while shaded areas are the 25th and 75th percentiles. Results are obtained across 200 market simulations using the trained agents.}
\label{Appendix - fig: Price and emissions market reruns.}
\end{figure}

\clearpage

\section{PyPSA description}
\label{Appendix - PyPSA description}

PyPSA \textit{(Python for Power System Analysis)} is an open-source toolbox for the simulation and optimization of energy systems, capable of representing generation, storage, transmission, and demand with high temporal resolution \cite{brown_pypsa_2018}. Building on this framework, PyPSA-Eur provides a pan-European extension that incorporates detailed datasets on the European transmission grid, generation capacities, storage technologies, and demand time series \cite{victoria_speed_2022,brown_pypsa-eur_2024}. Together, they enable transparent, reproducible, and policy-relevant analyses of energy transitions, particularly in assessing decarbonization pathways, cross-border electricity trade, and the large-scale integration of renewable resources. Unlike agent-based models like MARL, which simulate the strategic behavior and interactions of individual electricity firms, PyPSA and PyPSA-Eur rely on centralized optimization, determining system-wide dispatch and investment decisions to minimize overall cost or emissions rather than capturing firm-level decision-making dynamics. 

To ensure a consistent basis for comparison between PyPSA and the MARL framework, we established a harmonized modeling protocol covering generation, demand, policy, and technical constraints. PyPSA-Eur is applied to Italy using a two-node representation and a brownfield approach, with hourly temporal resolution and myopic foresight. The generation portfolio includes wind, solar, offshore wind, OCGT, CCGT, and coal, with resource profiles derived from \cite{antonini_weather-_2024,antonini_weather-_2024-1} under RCP 2.6. Existing assets are based on PyPSA-Eur data, while decommissioning is treated according to the asset lifetimes, update the decommissioning curve for the MARL model. Generation costs follow PyPSA assumptions without modification, and demand projections are drawn from Terna \cite{snam_documento_2024}, with potential adjustments to account for import shares. A stylized carbon tax is introduced exogenously, while adequacy constraints are kept unchanged across both models. No further policy constraints are imposed in the model. Storage is represented through battery investments with a three-hour duration and hydro pumped storage operation, excluding hydrogen. The simulations cover a time horizon of approximately 20 years, with PyPSA-Eur optimizing investments for 2030 and 2040 snapshot years. Finally, no unit commitment constraints are applied in the both models.

The differences in results between the two modeling approaches can be attributed to several structural features. First, in the Multi-Agent Reinforcement Learning framework, decision-making is decentralized, with individual agents pursuing revenue-maximizing strategies rather than optimizing system-wide outcomes, in contrast to the central planner approach of PyPSA. Second, market dynamics play a critical role, as prices, bidding strategies, competition, and the possibility of market inefficiencies directly influence both investment and dispatch decisions. Furthermore, the temporal dimension of decision-making diverges: agents focus on short-term profitability, whereas the central planner model optimizes system development over a long-term horizon. Together, these differences highlight how assumptions about behavior, information, and temporal scope shape the resulting energy system configurations.

Importantly, in the Article's core, we have maintained the comparison between the two modeling approaches and the results of installed capacities. Given that the simulations in PyPSA were carried out using two brownfield optimizations, one for 2030 and another for 2040, comparisons made using metrics associated with system costs might be misleading. The MARL model does not provide a similar metric, considering that demand pays system prices instead of facing an annuity corresponding to system costs, as in PyPSA. Moreover, MARL simulations resemble, in specific ways, a perfect foresight scenario, in which agents are trained on a stochastic simulation, but one in which the main policy characteristics remain deterministic and known to agents. Instead, brownfield PyPSA optimization is myopic; the mix for 2030 is optimized for the year conditions, without optimizing for the whole 2020-2040 pathway. Nonetheless, for completeness, we report a comparison between system costs and emissions for the years 2030 and 2040 in \ref{Appendix - Table: PyPSA and MARL comparison.}.

\begin{table}[hbt!]
\centering
\resizebox{\textwidth}{!}{%
\begin{tabular}{|ccc|cccc|cccc|}
\hline
\multicolumn{3}{|c|}{\multirow{2}{*}{\textbf{Model-Year}}} &
  \multicolumn{4}{c|}{\textbf{\begin{tabular}[c]{@{}c@{}}Yearly System Costs\\ {[}Billions \texteuro{]}\end{tabular}}} &
  \multicolumn{4}{c|}{\textbf{\begin{tabular}[c]{@{}c@{}}Yearly Emissions\\ {[}Mton $CO_2${]}\end{tabular}}} \\ \cline{4-11} 
\multicolumn{3}{|c|}{} &
  \multicolumn{1}{c|}{\textbf{CM + CfD}} &
  \multicolumn{1}{c|}{\textbf{EoM}} &
  \multicolumn{1}{c|}{\textbf{CfD}} &
  \textbf{CM} &
  \multicolumn{1}{c|}{\textbf{CM + CfD}} &
  \multicolumn{1}{c|}{\textbf{EoM}} &
  \multicolumn{1}{c|}{\textbf{CfD}} &
  \textbf{CM} \\ \hline
\multicolumn{1}{|c|}{\multirow{6}{*}{\textbf{MARL}}} &
  \multicolumn{1}{c|}{\multirow{3}{*}{\textbf{2030}}} &
  Mean &
  \multicolumn{1}{c|}{\textbf{29.75}} &
  \multicolumn{1}{c|}{\textbf{23.99}} &
  \multicolumn{1}{c|}{\textbf{26.89}} &
  \textbf{27.25} &
  \multicolumn{1}{c|}{\textbf{61.61}} &
  \multicolumn{1}{c|}{\textbf{45.79}} &
  \multicolumn{1}{c|}{\textbf{64.55}} &
  \textbf{41.67} \\ \cline{3-11} 
\multicolumn{1}{|c|}{} &
  \multicolumn{1}{c|}{} &
  Max &
  \multicolumn{1}{c|}{31.82} &
  \multicolumn{1}{c|}{27.93} &
  \multicolumn{1}{c|}{39.44} &
  30.49 &
  \multicolumn{1}{c|}{71.74} &
  \multicolumn{1}{c|}{53.86} &
  \multicolumn{1}{c|}{76.36} &
  52.09 \\ \cline{3-11} 
\multicolumn{1}{|c|}{} &
  \multicolumn{1}{c|}{} &
  Min &
  \multicolumn{1}{c|}{27.79} &
  \multicolumn{1}{c|}{18.91} &
  \multicolumn{1}{c|}{25.12} &
  22.26 &
  \multicolumn{1}{c|}{50.54} &
  \multicolumn{1}{c|}{35.60} &
  \multicolumn{1}{c|}{51.18} &
  31.93 \\ \cline{2-11} 
\multicolumn{1}{|c|}{} &
  \multicolumn{1}{c|}{\multirow{3}{*}{\textbf{2040}}} &
  Mean &
  \multicolumn{1}{c|}{\textbf{35.59}} &
  \multicolumn{1}{c|}{\textbf{36.17}} &
  \multicolumn{1}{c|}{\textbf{33.78}} &
  \textbf{29.83} &
  \multicolumn{1}{c|}{\textbf{23.05}} &
  \multicolumn{1}{c|}{\textbf{45.12}} &
  \multicolumn{1}{c|}{\textbf{24.26}} &
  \textbf{28.83} \\ \cline{3-11} 
\multicolumn{1}{|c|}{} &
  \multicolumn{1}{c|}{} &
  Max &
  \multicolumn{1}{c|}{40.23} &
  \multicolumn{1}{c|}{56.66} &
  \multicolumn{1}{c|}{38.73} &
  37.45 &
  \multicolumn{1}{c|}{35.73} &
  \multicolumn{1}{c|}{55.94} &
  \multicolumn{1}{c|}{35.01} &
  42.53 \\ \cline{3-11} 
\multicolumn{1}{|c|}{} &
  \multicolumn{1}{c|}{} &
  Min &
  \multicolumn{1}{c|}{29.33} &
  \multicolumn{1}{c|}{20.51} &
  \multicolumn{1}{c|}{28.47} &
  22.71 &
  \multicolumn{1}{c|}{12.16} &
  \multicolumn{1}{c|}{35.51} &
  \multicolumn{1}{c|}{13.10} &
  17.87 \\ \hline
\multicolumn{1}{|c|}{\multirow{2}{*}{\textbf{PyPSA}}} &
  \multicolumn{2}{c|}{\textbf{2030}} &
  \multicolumn{4}{c|}{31.28} &
  \multicolumn{4}{c|}{57.91} \\ \cline{2-11} 
\multicolumn{1}{|c|}{} &
  \multicolumn{2}{c|}{\textbf{2040}} &
  \multicolumn{4}{c|}{32.33} &
  \multicolumn{4}{c|}{65.58} \\ \hline
\end{tabular}%
}
\caption{Comparison of system costs and $CO_2$ emissions for 2030 and 2040 obtained with the MARL and PyPSA models. For MARL, system costs correspond to the total costs assumed by demand, considering all mechanisms present in the market design. For PyPSA, the annualized system costs are reported, also equivalent to the objective function from the central planning optimization problem. Electricity system emissions are obtained using the emission factors from the corresponding technologies. Results from the MARL model are obtained across 200 market simulations using the trained agents, in a configuration with 16 agents, and are reported as average, minimum, and maximum values across tests.}
\label{Appendix - Table: PyPSA and MARL comparison.}
\end{table}

\end{document}